%% file: root.tex
\documentclass{ifacconf}

\usepackage{graphicx}      % include this line if your document contains figures
\usepackage{natbib}        % required for bibliography

\usepackage{amsmath}            % Included additionally to template
\usepackage{amsfonts}           % Included additionally to template
\usepackage[section]{algorithm} % Included additionally to template
\usepackage{algpseudocode}      % Included additionally to template
\usepackage{booktabs}       % Included additionally to template
\usepackage{tabularx}       % Included additionally to template
\usepackage{supertabular}   % Included additionally to template
\usepackage{multirow}       % Included additionally to template
\usepackage{makecell}       % Included additionally to template
\usepackage{subcaption}	    % Included additionally to template
\usepackage{pgfplots}       % Included additionally to template
\usepackage{tikz}           % Included additionally to template
\usetikzlibrary{shapes}     % Included additionally to template
\usepackage[binary-units]{siunitx} % Included additionally to template

\begin{document}
\begin{frontmatter}

\title{Trajectory Planning for Non-Communicating Mobile Robots using Inverse Optimal Control}

\thanks[footnoteinfo]{\copyright 2026 the authors. This work has been accepted to IFAC for publication under a Creative Commons Licence CC-BY-NC-ND}

\author[First]{Nina Majer}
\author[First]{Yannick Epple}
\author[First]{Xin Ye}
\author[First]{Stefan Schwab}
\author[Second]{Sören Hohmann}

\address[First]{FZI Research Center for Information Technology, Haid-und-Neu-Str.10-14, 76131 Karlsruhe, Germany \\ (e-mail: majer@fzi.de).}
\address[Second]{Institute of Control Systems, Karlsruhe
Institute of Technology, Kaiserstraße 12, 76131 Karlsruhe, Germany \\ (e-mail: soeren.hohmann@kit.edu)}

\input{content/0_abstract.tex}


\end{frontmatter}
%===============================================================================

\input{content/1_introduction.tex}
\input{content/2_related_work.tex}
\input{content/3_problem_definition.tex}
\input{content/4_inverse_optimal_control_based_trajectory_prediction.tex}
\input{content/5_simulative_evaluation.tex}
\input{content/6_conclusion.tex}

\bibliography{ifacconf}             % bib file to produce the bibliography
                                            % with bibtex (preferred)
\end{document}

%% file: content/0_abstract.tex
\begin{abstract}                % Abstract of at most 2250 characters

To enable an efficient interaction of non-communicating mobile robots in collision avoidance scenarios, we present a novel combined trajectory planning and prediction algorithm. Inverse optimal control is used to estimate unknown goal states of all robots based on observed past trajectories. Each robot also takes the perspective of other robots in considering self-prediction and solves a joint prediction problem using the estimated goal states. The resulting predictions are then considered for planning. Simulation results of scenarios with 2-8 robots show that the median of the durations until all vehicles reach their goals is \num{9.8} \% faster compared to planning with constant acceleration based estimated goal states. Moreover, the proposed approach never leads to the solver being unable to find a solution to the planning or prediction problem.

\end{abstract}

%% file: content/1_introduction.tex
\section{Introduction}
\label{sec:introduction}

Various fields of application benefit from the use of autonomous mobile robots (AMRs). Especially, there is a trend in intralogistics to replace automated guided vehicles by AMRs because of the ability to navigate in semi-structured and changing environments without physical or virtual reference paths, see \cite{tenhompel2020}, p.155. Freely navigating AMRs thus support the transformation from rigidly structured to flexibly configurable factories in the course of Industry 4.0.

In local intersection scenarios, where robots move close to each other and want to avoid collisions, coordinated motion planning is necessary to ensure an efficient traffic flow and prevent deadlocks. In present intralogistics, the robots are coordinated by a central control. If the robots are no longer able to avoid each other autonomously without collisions, the central control intervenes to resolve the deadlock, as specified within the standard VDA5050, see \cite{vda2025}. However, this leads to long waiting times in case of a large robot fleet or failure of the central control. Hence, this work focuses on decentralized coordination, where the robots resolve the conflict independently.

If the robots are equipped with a communication interface, data on the planned trajectory or goal states can be transmitted to or received from other robots, facilitating the resolution of local intersection scenarios. We consider the case where no data exchange via a communication interface is possible. In this case, the non-communicating robots need the ability to predict the motion of other robots. Further, the interaction in intersection scenarios requires that the robots consider the mutual influence of the motions on each other, both in planning and in prediction, to resolve conflicts efficiently.

Therefore, we propose a novel combined trajectory planning and prediction algorithm that enables mobile robots to resolve local intersection scenarios in a decentralized manner without communication. The prediction phase of the algorithm starts with using inverse optimal control (IOC) to estimate unknown parameters of the goal states of all robots based on observed trajectories. Then, trajectory predictions for each robot are computed by solving a joint prediction problem using the estimated goal states. The predictions are considered in the planning phase.

This paper is structured as follows: In Section II, we provide an overview of the state of the art in motion prediction of AMRs as well as online applicable IOC. In Section III, the trajectory planning problem is given. Section IV introduces the implemented IOC based prediction method, followed by its combination with planning. A simulative evaluation is given in Section V and the conclusion and future research is outlined in Section VI.

%% file: content/2_related_work.tex
\section{Related Work}
\label{sec:related_work}

The survey on motion prediction in autonomous driving by \cite{karle2022} categorizes the existing approaches into physics-, pattern- and planning-based methods. Physics-based methods incorporate simple prediction methods based on the assumption that the vehicle to be predicted is driving at a constant speed or acceleration, possibly in combination with a constant turn or curvature rate. Pattern-based methods process data to find the most probable prototypical trajectory or maneuver. Besides classification and clustering methods without predefined classes, encoder-decoder architectures based on deep neural networks exist. While physics-based methods are suitable for accurate predictions over a short horizon, pattern-based methods, especially those based on encoder-decoder architectures, deliver more precise results for longer horizons and more complex vehicle interactions.

The key feature of planning-based methods is the assumption that the vehicle motion is influenced by an implicit objective. The goal is to estimate an underlying cost function that led to the optimal motion strategy, or to directly estimate the strategy to predict the future motion. The methods from the field of machine learning are inverse reinforcement learning and imitation learning where the Markov Decision Process serves as the mathematical framework respectively. In control theory, IOC has been developed with the same goal of identifying the underlying cost function of another agent that led to his optimal strategy. The modeling in IOC is deterministic, using differential equations that reflect the system dynamics.

In this paper, we use IOC to predict other robots' future states by analyzing their current and past states. Compared to physics-based methods, the mutual influence of the motion of interacting robots through the intended collision avoidance is considered. In contrast to pattern-based methods, especially deep learning, this gives us physical interpretability, explainability, and eliminates the need for intensive offline pre-training with a large data set.

IOC-based prediction must have the ability to use incompletely observed trajectories to estimate the unknown cost function of the optimization problem of another agent. Thus, the method can be applied online. To simplify the IOC problem, a common approach is to assume a parameterized cost function structure and thereby reducing the problem to the estimation of the unknown cost function parameters. Online IOC approaches that can be found in the literature are shortly outlined in the following.

Online IOC based on bilevel optimization, see \cite{reiter2022} exists. The idea behind bilevel optimization for IOC is to iteratively compute the solution resulting from the currently determined parameters, and adapt the parameters such that the difference between the currently determined solution strategy and the observations is minimized. The first step is referred to as solving the forward problem. In the second step, the inverse problem is solved. Before starting the bilevel optimization loop, an initial guess for the unknown parameters of the forward problem is set.

Furthermore, IOC based on the Kalman filter, see \cite{le_cleach2021}, \cite{menner2021_2} and \cite{zhao2024}, has been proposed, where the observed current and past states as well as the inputs can be subject to noise. Another procedure is to formulate optimality conditions of the forward problem. The system of equations resulting from optimality conditions is then solved for the unknown parameters. \cite{menner2021_1}, \cite{peters2023}, \cite{liu2023} and \cite{khan2025} implement this method by formulating the Karush Kuhn Tucker (KKT) conditions. Further, \cite{molloy2020} uses the discrete-time Pontryagin's Maximum Principle and \cite{karg2024} considers the Hamilton-Jacobi-Bellman equation.

We focus on solving the system of equations resulting from optimality conditions once since it reduces the computational effort to one computational iteration in a prediction step. Bilevel optimization and the Kalman filter require multiple computational iterations in a prediction step. \cite{liu2023}, \cite{khan2025} and \cite{peters2023} have also presented the integration of their online IOC into a combined planning and prediction algorithm. However, in these approaches, the ego agent does not consider self-prediction, i.e. taking other agents' perspective in prediction and determining how other agents predict the ego agent's motion. \cite{gil2025} also state the necessity of self-prediction to resolve interactive scenarios without communication. Whereby \cite{gil2025} present a model to consider predictability by other agents in planning.

Thus, we present a novel combined trajectory planning and prediction algorithm using IOC, that accounts for self-prediction. Within the prediction phase, the parameters of all robots' goal states are estimated by using IOC based on observed trajectories. The IOC method is based on solving the KKT conditions and builds on the procedure of \cite{menner2021_1}. Then, trajectory predictions are computed by solving a joint prediction problem using the estimated goal state parameters and considered within the planning problem. The combined algorithm is repeated until each robot reaches its goal state without the need for communication or centralized coordination. We have simulatively evaluated the algorithm in intersection scenarios with 2-8 robots. The results show that each of the scenario could be resolved collision-free and the median of the total durations until all vehicles reach their goals is \num{9.8} \% shorter compared to planning with constant acceleration based estimated goal states. Moreover, the proposed approach never results in the solver finding no solution to a planning or prediction problem.

%% file: content/3_problem_definition.tex
\section{Problem Definition}
\label{sec:problem_definition}

We consider $N$ mobile robots with the nonlinear discrete-time system dynamics of each robot $i \in \{1, \cdots, N \}$ given by $\mathbf{x}^{(i)}_{t+1} = f(\mathbf{x}^{(i)}_t, \mathbf{u}^{(i)}_t) \quad \forall t \geq 0 $ where $\mathbf{x}^{(i)}_t \in \mathbb{R}^{n_{\mathbf{x}}}$ and $\mathbf{u}^{(i)}_t \in \mathbb{R}^{n_{\mathbf{u}}}$ correspond to the robot's state and input at time step $t \in \mathbb{N}$. The robots aim to find a solution to the optimal control problem (OCP) at $t$ given by
\begin{equation} \label{eq:planning_ocp_robot_i}
\begin{array}{ll}
&\{{\mathbf{x}^{*}}^{(i)}_{t:t+K_{\mathrm{P}}|t}, {\mathbf{u}^{*}}^{(i)}_{t:t+K_{\mathrm{P}}-1|t}\} \\
&= \arg \underset{\mathbf{x}^{(i)}_{k|t}, \mathbf{u}^{(i)}_{k|t}}{\mathrm{min}} \sum_{k=t}^{t+K_{\mathrm{P}}-1} l(\mathbf{x}_{k|t}^{(i)}, \mathbf{u}_{k|t}^{(i)}, \mathbf{L}^{(i)}) \\
& \hspace{2cm} + \phi(\mathbf{x}_{t+K_{\mathrm{P}}|t}^{(i)}, \mathbf{L}^{(i)}) \\
 &\mathrm{s.t.} \hspace{0.1cm} \mathbf{x}^{(i)}_{k+1|t} = f(\mathbf{x}^{(i)}_{k|t}, \mathbf{u}^{(i)}_{k|t}) \hspace{0.9cm} \forall k \in \, \left[t, \cdots, t+K_{\mathrm{P}}-1 \right], \\
& \hspace{0.5cm} C(\mathbf{x}_{k|t}^{(i)}, \tilde{\mathbf{x}}_{k|t}^{(j)}, \mathbf{u}_{k|t}^{(i)}, \mathbf{D}^{(i)}) \leq \mathbf{0} \hspace{0.1cm} \forall k \in \left[t, \cdots, t+K_{\mathrm{P}} \right], \\
& \hspace{4.7cm} \forall j \in \{1, \cdots, N \}, j \neq i, \\
& \hspace{0.6cm} \mathbf{x}^{(i)}_{t|t} = \bar{\mathbf{x}}^{(i)}_t 
\end{array}
\end{equation}
in order to plan collision-free trajectories in a model predictive control fashion with the prediction horizon $K_{\mathrm{P}}$. 

The sequences of states
$\mathbf{x}^{(i)}_{t:t+K_{\mathrm{P}}|t} := \begin{bmatrix} {\mathbf{x}^{(i)}_{t|t}}^{\mathrm{T}}, \cdots, {\mathbf{x}^{(i)}_{t+K_{\mathrm{P}}|t}}^{\mathrm{T}} \end{bmatrix}^{\mathrm{T}}$ and inputs $\mathbf{u}^{(i)}_{t:t+K_{\mathrm{P}}-1|t} := \begin{bmatrix} {\mathbf{u}^{(i)}_{t|t}}^{\mathrm{T}}, \cdots, {\mathbf{u}^{(i)}_{t+K_{\mathrm{P}}-1|t}}^{\mathrm{T}} \end{bmatrix}^{\mathrm{T}}$
comprise the states and inputs planned at $t$ from $t$ to $t+K_{\mathrm{P}}$. The annotation $\tilde{(\cdot)}$ indicates a predicted or estimated variable and $\bar{(\cdot)}$ denotes an observed variable. Further, the parametric convex cost functions $l(\cdot)$ and $\phi(\cdot)$, the potentially nonlinear constraints function $C(\cdot)$ as well as the model function $f(\cdot)$ are assumed to be continuous and time-invariant. Additionally, $l(\cdot)$, $\phi(\cdot)$ and $C(\cdot)$ are assumed to be linear in $\mathbf{L}^{(i)}$ and $\mathbf{D}^{(i)}$. 

We assume that each robot can measure or estimate all states $\mathbf{x}^{(i)}_{0:t} := \begin{bmatrix}  {\mathbf{x}^{(i)}_{0}}^{\mathrm{T}}, \cdots, {\mathbf{x}^{(i)}_{t}}^{\mathrm{T}} \end{bmatrix}^{\mathrm{T}}$
$\forall i$ and $t \geq 0$ and store them. We consider scenarios, in which a robot $i$ knows its own parameter vector $\mathbf{L}^{(i)}$ but does not know $\mathbf{L}^{(j)}$ of the other robots $j \in \{1, \cdots, N \}, j \neq i$.

%% file: content/4_inverse_optimal_control_based_trajectory_prediction.tex
\section{Inverse Optimal Control based Trajectory Prediction}
\label{sec:inverse_optimal_control_based_trajectory_prediction}

This section presents the IOC approach to estimate the unknown parameter vector $\tilde{\mathbf{L}}^{(i)}$ of the cost function of each robot $i$ based on the observed robot's past trajectories $\bar{\mathbf{x}}^{(i)}_{0:t}$. We follow the idea of \cite{menner2021_1} to derive a shortest path problem, see OCP (\ref{eq:shortest_path_planning_ocp_robot_i}), from the considered infinite-horizon OCP in order to deal with incompletely observed segments of a trajectory that is assumed to be optimal w.r.t. the infinite-horizon OCP. We form the KKT optimality conditions of the shortest path OCP. Then, we solve the resulting system of equations and inequalities for the unknown parameters, given an observed trajectory segment $\bar{\mathbf{x}}^{(i)}_{t-K_{\mathrm{E}}:t}$. Using the estimated parameters $\tilde{\mathbf{L}}^{(i)} \, \forall i$, the future trajectories $\tilde{\mathbf{x}}_{t:t+K_{\mathrm{P}}}^{(i)}$ of each robot $i$ are predicted by solving a joint prediction optimization problem.

In the following, the individual steps of the IOC approach are elaborated. The shortest path OCP on the time interval $t-K_{\mathrm{E}}$ to $t$ that is inferred from the infinite-horizon version of the trajectory planning optimization problem (\ref{eq:planning_ocp_robot_i}) of a robot $i$ is given by
\begin{equation} \label{eq:shortest_path_planning_ocp_robot_i}
\begin{array}{ll}
&\{{\mathbf{x}^{*}}^{(i)}_{t-K_{\mathrm{E}}:t|t}, {\mathbf{u}^{*}}^{(i)}_{t-K_{\mathrm{E}}:t-1|t}\} \\
& \hspace{1cm} = \arg \underset{\mathbf{x}^{(i)}_{k|t}, \mathbf{u}^{(i)}_{k|t}}{\mathrm{min}} \sum_{k=t-K_{\mathrm{E}}}^{t-1} l(\mathbf{x}_{k|t}^{(i)}, \mathbf{u}_{k|t}^{(i)}, \tilde{\mathbf{L}}^{(i)}) \\
 &\mathrm{s.t.} \hspace{0.1cm} \mathbf{x}^{(i)}_{k+1|t} = f(\mathbf{x}^{(i)}_{k|t}, \mathbf{u}^{(i)}_{k|t}) \hspace{0.9cm} \forall k \in \, \left[t-K_{\mathrm{E}}, \cdots, t-1 \right], \\
& \hspace{0.5cm} C(\mathbf{x}_{k|t}^{(i)}, \mathbf{x}_{k|t}^{(j)}, \mathbf{u}_{k|t}^{(i)}, \mathbf{D}^{(i)}) \leq \mathbf{0} \hspace{0.1cm} \forall k \in \left[t-K_{\mathrm{E}}, \cdots, t \right], \\
& \hspace{4.7cm} \forall j \in \{1, \cdots, N \}, j \neq i, \\
& \hspace{0.6cm} \mathbf{x}^{(i)}_{t-K_{\mathrm{E}}|t} = \bar{\mathbf{x}}^{(i)}_{t-K_{\mathrm{E}}}, \\
& \hspace{0.6cm} \mathbf{x}^{(i)}_{t|t} = \bar{\mathbf{x}}^{(i)}_t,
\end{array}
\end{equation}
where $K_{\mathrm{E}}$ corresponds to the estimation horizon. \cite{menner2021_1} show that ${\mathbf{x}^{*}}^{(i)}_{t-K_{\mathrm{E}}:t|t}$ and ${\mathbf{u}^{*}}^{(i)}_{t-K_{\mathrm{E}}:t-1|t}$ also correspond to the minimizers of the infinite-horizon version of the OCP (\ref{eq:shortest_path_planning_ocp_robot_i}). In order to reformulate the OCP (\ref{eq:shortest_path_planning_ocp_robot_i}) such that it is only dependent on the inputs ${\mathbf{u}^{*}}^{(i)}_{t:t+k-1|t}$, the function $F_{t+k}(\cdot)$ is defined $\, \forall k \geq 0$ as
\begin{equation} \label{eq:definition_F_t_plus_k}
\begin{array}{ll}
\mathbf{x}^{(i)}_{t+k} = F_{t+k}(\mathbf{u}^{(i)}_{t:t+k-1},\bar{\mathbf{x}}^{(i)}_t), \\
F_{t+k}(\mathbf{u}^{(i)}_{t:t+k-1},\bar{\mathbf{x}}^{(i)}_t) \\
\hspace{1cm} := \left\{\begin{array}{ll} \bar{\mathbf{x}}^{(i)}_t \hspace{4.85cm} \mathrm{if} \, k = 0 \\ f(F_{t+k-1}(\mathbf{u}^{(i)}_{t:t+k-2},\bar{\mathbf{x}}^{(i)}_t),\mathbf{u}^{(i)}_{t+k-1}) \quad \mathrm{else}
\end{array} \right.
\end{array}
\end{equation}

The sequence of states and inputs during the estimation horizon $K_{\mathrm{E}}$ are substituted by $\mathbf{X}^{(i)} := \mathbf{x}^{(i)}_{t-K_{\mathrm{E}}:t}$ and \\ $\mathbf{U}^{(i)} := \mathbf{u}^{(i)}_{t-K_{\mathrm{E}}:t-1}$. Thus, we define the Lagrangian $\mathcal{L}(\mathbf{U}^{(i)},\lambda^{(i)},\nu^{(i)}, \tilde{\mathbf{L}}^{(i)})$ of the OCP (\ref{eq:shortest_path_planning_ocp_robot_i}) as
\begin{equation} \label{eq:Lagrange_function_shortest_path_planning_ocp_robot_i}
\begin{array}{ll}
\mathcal{L}(\mathbf{U}^{(i)},\mathbf{\lambda}^{(i)},\mathbf{\nu}^{(i)}, \tilde{\mathbf{L}}^{(i)}) = {\mathbf{\nu}^{(i)}}^{\mathrm{T}} (F_t(\mathbf{U}^{(i)},\bar{\mathbf{x}}^{(i)}_{t-K_{\mathrm{E}}}) - \mathbf{x}^{(i)}_t) \\ + \sum_{k=t-K_{\mathrm{E}}}^{t-1} l(F_k(\mathbf{U}^{(i)},\bar{\mathbf{x}}^{(i)}_{t-K_{\mathrm{E}}}),\mathbf{u}^{(i)}_k, \tilde{\mathbf{L}}^{(i)}) \\
+ \, {\mathbf{\lambda}_k^{(i)}}^{\mathrm{T}} \sum_{j=1, j \neq i}^{N} C(F_k(\mathbf{U}^{(i)},\bar{\mathbf{x}}^{(i)}_{t-K_{\mathrm{E}}}), \mathbf{x}_k^{(j)}, \mathbf{u}^{(i)}_k), \mathbf{D}^{(i)})
\end{array}
\end{equation}
with the Lagrange multipliers $\mathbf{\nu}^{(i)} \in \mathbb{R}^{n_{\mathbf{x}}}$ and \\ $\mathbf{\lambda}^{(i)} := \begin{bmatrix} {\mathbf{\lambda}^{(i)}_{t-K_{\mathrm{E}}}}^{\mathrm{T}}, \cdots, {\mathbf{\lambda}^{(i)}_{t-1}}^{\mathrm{T}} \end{bmatrix}^{\mathrm{T}}$. 

Accordingly, the KKT conditions of problem (\ref{eq:shortest_path_planning_ocp_robot_i}) representing the forward problem are given by
\begin{subequations}\label{eq:KKT_conditions_shortest_path_planning_ocp_robot_i}
\begin{align}
&\nabla_{\mathbf{U}^{(i)}} \mathcal{L}(\mathbf{U}^{(i)},\lambda^{(i)},\nu^{(i)}, \tilde{\mathbf{L}}^{(i)}) = \mathbf{0}, \label{eq:KKT_a} \hspace{1.5cm} \\[1ex]
&{\mathbf{\lambda}_k^{(i)}}^{\mathrm{T}} C(F_k(\mathbf{U}^{(i)},\bar{\mathbf{x}}^{(i)}_{t-K_{\mathrm{E}}}),\mathbf{u}^{(i)}_k, \mathbf{D}^{(i)}) = \mathbf{0} \nonumber \\ & \hspace{4.1cm} \forall k \in [t-K_{\mathrm{E}}, \dots, t-1], \label{eq:KKT_b} \\[1ex]
&\mathbf{\lambda}_k^{(i)} \geq \mathbf{0} \hspace{2.9cm} \forall k \in [t-K_{\mathrm{E}}, \dots, t-1], \label{eq:KKT_c} \\[1ex]
&C(F_k(\mathbf{U}^{(i)},\bar{\mathbf{x}}^{(i)}_{t-K_{\mathrm{E}}}),\mathbf{x}_k^{(j)},\mathbf{u}^{(i)}_k, \mathbf{D}^{(i)}) \leq \mathbf{0} \nonumber \\ \nonumber & \hspace{4.1cm} \forall k \in [t-K_{\mathrm{E}}, \dots, t] \\
&\hspace{4.15cm} \forall j \in \{1, \cdots, N \}, j \neq i, \label{eq:KKT_d} \\[1ex]
&F_t(\mathbf{U}^{(i)},\bar{\mathbf{x}}^{(i)}_{t-K_{\mathrm{E}}}) - \bar{\mathbf{x}}^{(i)}_t = \mathbf{0}. \label{eq:KKT_e}
\end{align}
\end{subequations}

To solve the system of equations and inequalities of the KKT conditions for $\tilde{\mathbf{L}}^{(i)}$, $\lambda^{(i)}$ and $\nu^{(i)}$, the inverse OCP
\begin{equation} \label{eq:inverse_ocp_robot_i}
\begin{array}{ll}
&\{{\mathbf{\nu}^{*}}^{(i)}, {\mathbf{\lambda}^{*}}^{(i)}, \tilde{\mathbf{L}}^{*(i)}\}
= \arg \underset{\mathbf{\nu}^{(i)}, \mathbf{\lambda}^{(i)}, \tilde{\mathbf{L}}^{(i)}}{\mathrm{min}} \sum_{k={t-K_{\mathrm{E}}}}^{t-1} {\mathbf{\lambda}_{k}^{(i)}}^2  \\ & \hspace{1cm} + {\mathbf{\nu}^{(i)}}^2 + \left\lVert {\mathbf{\lambda}_k^{(i)}}^{\mathrm{T}} C(F_k(\mathbf{U}^{(i)},\bar{\mathbf{x}}^{(i)}_{t-K_{\mathrm{E}}}),\mathbf{u}^{(i)}_k, \mathbf{D}^{(i)})\right\rVert_{2}^2 \\
&\hspace{1cm} \mathrm{s.t.} \hspace{1cm} (\ref{eq:KKT_a}), (\ref{eq:KKT_c})
\end{array}
\end{equation}
is solved. Hence, to predict the future trajectories $\tilde{\mathbf{x}}_{t:t+K_{\mathrm{P}}}^{(i)}$ of each robot $i$, the joint prediction problem 
\begin{equation} \label{eq:predicting_ocp_robot_i}
\begin{array}{ll}
&\{\tilde{\mathbf{x}}^{*}_{t:t+K_{\mathrm{P}}|t}, \tilde{\mathbf{u}}^{*}_{t:t+K_{\mathrm{P}}-1|t}\} \\
&= \arg \underset{\tilde{\mathbf{x}}_{k|t}, \tilde{\mathbf{u}}_{k|t}}{\mathrm{min}} \sum_{i=1}^{N} \bigg( \sum_{k=t}^{t+K_{\mathrm{P}}-1} l(\tilde{\mathbf{x}}_{k|t}^{(i)}, \tilde{\mathbf{u}}_{k|t}^{(i)}, \tilde{\mathbf{L}}^{*(i)}) \bigg) \\
& \hspace{2.5cm} + \phi(\tilde{\mathbf{x}}_{t+K_{\mathrm{P}}|t}^{(i)}, \tilde{\mathbf{L}}^{*(i)}) \\ \\
 &\mathrm{s.t.} \, \forall i \in \{1, \cdots, N \} \\
 &\hspace{0.5cm} \tilde{\mathbf{x}}^{(i)}_{k+1|t} = f(\tilde{\mathbf{x}}^{(i)}_{k|t}, \tilde{\mathbf{u}}^{(i)}_{k|t}) \hspace{1.0cm} \forall k \in \, \left[t, \cdots, t+K_{\mathrm{P}}-1 \right], \\
& \hspace{0.5cm} C(\tilde{\mathbf{x}}_{k|t}^{(i)}, \tilde{\mathbf{x}}_{k|t}^{(j)}, \tilde{\mathbf{u}}_{k|t}^{(i)}, \mathbf{D}^{(i)}) \leq \mathbf{0} \hspace{0.1cm} \forall k \in \left[t, \cdots, t+K_{\mathrm{P}} \right], \\
& \hspace{4.7cm} \forall j \in \{1, \cdots, N \}, j \neq i, \\
& \hspace{0.6cm} \tilde{\mathbf{x}}^{(i)}_{t|t} = \bar{\mathbf{x}}^{(i)}_t 
\end{array}
\end{equation}
is solved based on the estimated parameters $\tilde{\mathbf{L}}^{*(i)}$ where $\tilde{\mathbf{x}}_t \in \mathbb{R}^{n_{\mathbf{x}} \cdot N}$ and $\tilde{\mathbf{u}}_t \in \mathbb{R}^{n_{\mathbf{u}} \cdot (N-1)}$ correspond to the stacked states and inputs of the robots, i.e. \\ $\tilde{\mathbf{x}}_t :=
\begin{bmatrix}
{\tilde{\mathbf{x}}^{(1), \mathrm{T}}_t}, \cdots, {\tilde{\mathbf{x}}^{(N), \mathrm{T}}_t}
\end{bmatrix}^{\mathrm{T}}, \quad
\tilde{\mathbf{u}}_t :=
\begin{bmatrix}
{\tilde{\mathbf{u}}^{(1), \mathrm{T}}_t}, \cdots, {\tilde{\mathbf{u}}^{(N), \mathrm{T}}_t}
\end{bmatrix}^{\mathrm{T}}$.
We assume that each robot $i$ adopts the unified presented IOC-based prediction model resulting in a "soft social convention", similar as proposed in \cite{gil2025}.

Algorithm \ref{alg:combined_trajectory_planning_and_prediction} shows the steps of the combined trajectory planning and prediction method for each robot $i$ from the perspective of the robot $I$ to resolve intersection scenarios without communication but using IOC-based prediction. When enough observed states $\bar{\mathbf{X}}^{(i)} \, \forall i$ have been collected, i.e. $\, t \geq K_{\mathrm{E}}$, the unknown parameters $\tilde{\mathbf{L}}^{(i)}$ are estimated based on IOC to predict the future trajectories $\tilde{\mathbf{x}}_{t:t+K_{\mathrm{P}}}^{(i)}  \, \forall i$. If $\, t < K_{\mathrm{E}}$, the unknown parameters $\tilde{\mathbf{L}}^{(i)}$ are estimated based on an alternative simple prediction model.
\begin{algorithm}[ht!]
\floatname{algorithm}{Algorithm}
\caption{Combined planning and prediction}
\begin{algorithmic}[1]
    \State Set $t = 0$
    \Repeat $\, t \leftarrow t + 1$
        \State Measure $\bar{\mathbf{x}}_t^{(i)} \, \forall i$
        \State Update $\bar{\mathbf{X}}^{(i)}$ with $\bar{\mathbf{x}}_t^{(i)} \, \forall i$
        \State \textbf{if} $ \, t \geq K_{\mathrm{E}}$
            \State \hspace{2em} Estimate $\tilde{\mathbf{L}}^{(i)}$ by solving (\ref{eq:inverse_ocp_robot_i}) $\forall i$
        \State \textbf{else}
            \State \hspace{2em} Estimate $\tilde{\mathbf{L}}^{(i)}$ by simple prediction model
        \State Predict $\tilde{\mathbf{x}}_{t:t+K_{\mathrm{P}}}^{(i)}, \tilde{\mathbf{u}}_{t:t+K_{\mathrm{P}}-1}^{(i)}$ by solving (\ref{eq:predicting_ocp_robot_i})
        \State Plan $\mathbf{x}_{t:t+K_{\mathrm{P}}}^{(i)}, \mathbf{u}_{t:t+K_{\mathrm{P}}-1}^{(i)}$ by solving (\ref{eq:planning_ocp_robot_i})
        \State Apply $ {\mathbf{u}_{t|t}^{*}}^{(I)}$
    \Until Stopping criterion reached
\end{algorithmic} 
\label{alg:combined_trajectory_planning_and_prediction}
\end{algorithm}

%% file: content/5_simulative_evaluation.tex
\section{Simulative Evaluation}
\label{sec:simulative_evaluation}

\begin{figure}[!ht]
	\centering
 % t = 1.6s
    \begin{subfigure}{0.24\textwidth}
		\centering
		\input{img/mpc-ioc/scenario8a_mpc_ioc_planning30_safety_distance0.2_data_plot_of_paths_t16.tex}
        \vspace{-13pt}
		\caption{Estimated $\tilde{\mathbf{x}}_{\mathrm{F}}$ at $t=1.6$ \si{\second} }
	\end{subfigure}
	\begin{subfigure}{0.24\textwidth}
		\centering
		\input{img/mpc-real-goal-state/scenario8a_mpc_real-goal-state_planning30_safety_distance0.2_data_plot_of_paths_t16.tex}
        \vspace{-13pt}
		\caption{Exchanged $\mathbf{x}_{\mathrm{F}}$ at $t=1.6$ \si{\second} }
	\end{subfigure}
 % t = 3.0s
    \begin{subfigure}{0.24\textwidth}
		\centering
		\input{img/mpc-ioc/scenario8a_mpc_ioc_planning30_safety_distance0.2_data_plot_of_paths_t30.tex}
        \vspace{-13pt}
		\caption{Estimated $\tilde{\mathbf{x}}_{\mathrm{F}}$ at $t=3.0$ \si{\second} }
	\end{subfigure}
	\begin{subfigure}{0.24\textwidth}
		\centering
		\input{img/mpc-real-goal-state/scenario8a_mpc_real-goal-state_planning30_safety_distance0.2_data_plot_of_paths_t30.tex}
        \vspace{-13pt}
		\caption{Exchanged $\mathbf{x}_{\mathrm{F}}$ at $t=3.0$ \si{\second} }
	\end{subfigure}
 % t = 12.2s
    \begin{subfigure}{0.24\textwidth}
		\centering
		\input{img/mpc-ioc/scenario8a_mpc_ioc_planning30_safety_distance0.2_data_plot_of_paths.tex}
        \vspace{-13pt}
		\caption{Estimated $\tilde{\mathbf{x}}_{\mathrm{F}}$ at $t=12.2$ \si{\second} }
	\end{subfigure}
	\begin{subfigure}{0.24\textwidth}
		\centering
		\input{img/mpc-real-goal-state/scenario8a_mpc_real-goal-state_planning30_safety_distance0.2_data_plot_of_paths.tex}
        \vspace{-13pt}
		\caption{Exchanged $\mathbf{x}_{\mathrm{F}}$ at $t=12.2$ \si{\second} }
	\end{subfigure}
 % general caption / label
	\caption{Comparison of the combined planning and prediction algorithm in the 8-agent scenario with IOC-based estimated and exchanged known goal states $\mathbf{x}_{\mathrm{F}}$.}
	\label{fig:comparison_combined_planning_and_prediction_algorithm_8a}
\end{figure}
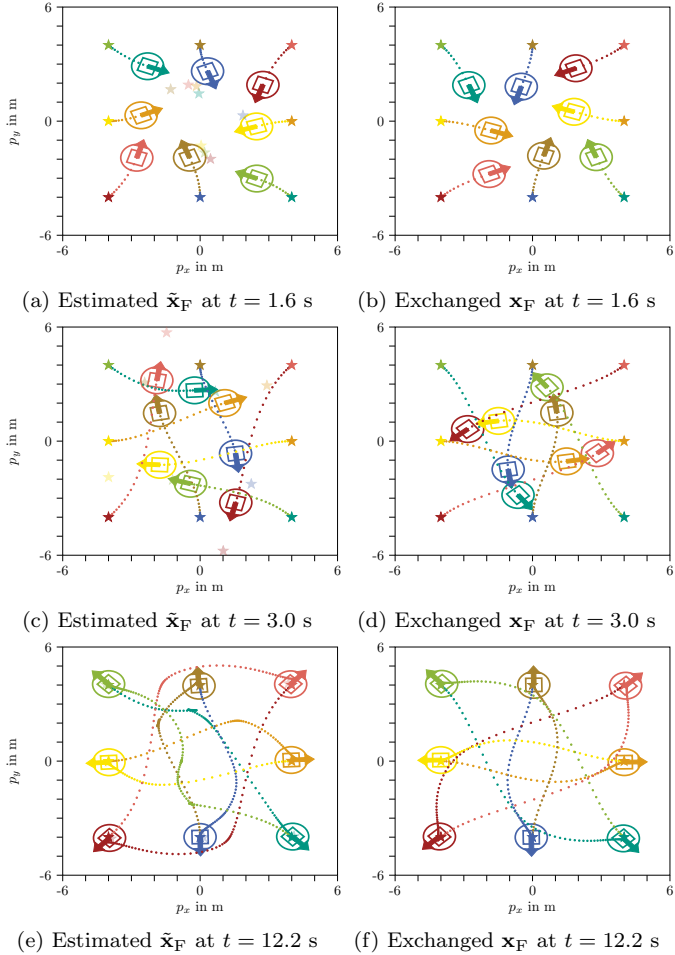

In this section, the combined planning and prediction algorithm is evaluated in simulations of collision avoidance scenarios with 2-8 mobile robots. We have simulated 40 scenarios with 2-8 mobile robots in a 2D simulation environment. As shown in Fig. \ref{fig:comparison_combined_planning_and_prediction_algorithm_8a}, the initial and goal positions are selected on a \num{4} \si{\meter} $\times$ \num{4} \si{\meter} area. 

We have considered differentially-driven vehicles whose nonlinear discrete-time system dynamics are given by
\begin{equation} 
\label{eq:vehicle_model_of_robot_i}
\begin{array}{ll}
    \mathbf{x}^{(i)}_{t+1} = \mathbf{x}^{(i)}_t + \Delta t \begin{bmatrix} x^{(i)}_{t,4} \cos(x^{(i)}_{t,3}) \\ x^{(i)}_{t,4} \sin(x^{(i)}_{t,3}) \\ u^{(i)}_{t,1} \\ u^{(i)}_{t,2} \end{bmatrix} = f^{(i)}(\mathbf{x}^{(i)}_t, \mathbf{u}^{(i)}_t)
\end{array}
\end{equation}
with $\mathbf{x}^{(i)}_t = \begin{bmatrix}  p^{(i)}_{x,t} & p^{(i)}_{y,t} & \psi^{(i)}_t & v^{(i)}_t \end{bmatrix}^{\mathrm{T}}$ and $\mathbf{u}^{(i)}_t = \begin{bmatrix}  \omega^{(i)}_t & a^{(i)}_t \end{bmatrix}^{\mathrm{T}}$.
The components $p^{(i)}_x$ and $p^{(i)}_y$ are the coordinates of the vehicle reference point $\mathbf{p}^{(i)}_r$ w.r.t. the inertial frame. Further, $\psi^{(i)}$ represents the heading, $v^{(i)}_t$ the velocity of $\mathbf{p}^{(i)}_r$, $\omega^{(i)}_t$ the angular velocity and $a^{(i)}_t$ the longitudinal acceleration.

Physical vehicle limitations and boundaries of the drivable area in terms of maximal and minimal values of the states, inputs and input rates are defined by
\begin{equation}  \label{eq:affine_state_input_constraints}
\begin{array}{ll}
    \mathbf{C}^{(i)}_{\mathbf{x}} \cdot \mathbf{x}^{(i)}_t - \mathbf{d}^{(i)}_{\mathbf{x}} &\leq \mathbf{0}, \\ 
    \mathbf{C}^{(i)}_{\mathbf{u}} \cdot \mathbf{u}^{(i)}_t - \mathbf{d}^{(i)}_{\mathbf{u}} &\leq \mathbf{0}, \\
    \mathbf{C}^{(i)}_{\mathbf{\triangle}\mathbf{u}} \cdot (\mathbf{u}^{(i)}_{t} - \mathbf{u}^{(i)}_{t-1}) - \mathbf{d}^{(i)}_{\mathbf{\triangle}\mathbf{u}} &\leq \mathbf{0},
\end{array}
\end{equation}
and circular collision avoidance constraints are defined by
\begin{equation}  \label{eq:coupled_state_constraints}
\begin{array}{ll}
    c^{(ij)}(\mathbf{x}^{(i)}_t, \tilde{\mathbf{x}}^{(j)}_t) = d^{(i)} - \left\lVert \begin{bmatrix} x^{(i)}_{t,1} \\ x^{(i)}_{t,2} \end{bmatrix} - \begin{bmatrix} \tilde{x}^{(j)}_{t,1} \\ \tilde{x}^{(j)}_{t,2} \end{bmatrix} \right\rVert_2 \leq 0
\end{array}
\end{equation}
where $d^{(i)} := r^{(i)} + r^{(j)}$. The distance $d^{(i)}$ represents the added radii of the virtual circles around $\mathbf{p}^{(i)}_r$ and $\mathbf{p}^{(j)}_r$. We summarize
$\mathbf{D}^{(i)} = \begin{bmatrix} \mathbf{d}^{(i)}_{\mathbf{x}} & \mathbf{d}^{(i)}_{\mathbf{u}} & d^{(i)} \end{bmatrix}^{\mathrm{T}}$. 

The intention of each robot $i$ to reach a desired goal state $\mathbf{x}^{(i)}_{\mathrm{F}}$ from an initial state $\mathbf{x}^{(i)}_{\mathrm{S}}$, is modelled by the weighted cost functions
\begin{equation} \label{eq:cost_function}
\begin{array}{ll}
l^{(i)}(\mathbf{x}_{t}^{(i)}, \mathbf{L}^{(i)}) &= \phi^{(i)}(\mathbf{x}_{t}^{(i)}, \mathbf{L}^{(i)}) \\
\nonumber &= (\mathbf{x}_{t}^{(i)} - \mathbf{L}^{(i)})^{\mathrm{T}} \mathbf{Q}_t (\mathbf{x}_{t}^{(i)} - \mathbf{L}^{(i)})
\end{array}
\end{equation}
$\forall i$ where $\mathbf{L}^{(i)} := \mathbf{x}^{(i)}_{\mathrm{F}}$. We set $\mathbf{x}^{(i)}_{\mathrm{F},4} := 0$ to force the vehicles to stop moving in their goal. Alg. \ref{alg:combined_trajectory_planning_and_prediction} is stopped at $t \geq 0$ by each robot $i$ when the goal state is reached w.r.t. a tolerance $\mu \geq 0$, i.e. $\left\lVert \mathbf{x}^{(I)}_t - \bar{\mathbf{x}}^{(I)}_{\mathrm{F}} \right\rVert_{2}^2 \leq \mu$. As a simple prediction model until the required estimation horizon for applying IOC is reached, we propagate the vehicle states by repeatedly applying the maximum jerk while respecting the physical vehicle limitations.

The parameterization of the simulated collision avoidance scenarios is given in Tab. \ref{tab:parameterization}. Our algorithm is implemented in Python using CasADi, see \cite{andersson2019}, and the Interior Point OPTimizer (IPOPT) as solver on a computer with an Intel i9-11900K CPU at \num{3.5} GHz $\times$ \num{16} with \num{32} GB RAM.

\begin{table}[!ht]
\caption{Parameterization.}
\label{tab:parameterization}
	\begin{center}
		\begin{tabular}{llrl}
			\toprule
			\textbf{Parameter Name and Variable} & \textbf{Value} & \textbf{Unit} \\
			\midrule
			Discretization time $\Delta t$ & \num{0.1} & \si{\second} \\
			Prediction horizon length $K_{\mathrm{P}}$ & \num{30} & \\
            Estimation horizon length $K_{\mathrm{E}}$ & \num{15} & \\
			Distance between actuated wheels $D^{(i)}$ & \num{0.7} & \si{\meter} \\
            Safety radius $r^{(i)}$ & \num{0.69} & \si{\meter} \\
            Max./min. position $p^{(i)}_{x/y,\mathrm{max/min}}$ & \num{10} &\si{\meter} \\
            Max./min. velocity $v^{(i)}_{\mathrm{max/min}}$ & \num{4.5} &\si{\meter\per\second} \\
            Max./min. acceleration $a^{(i)}_{\mathrm{max/min}}$ & \num{3} &\si{\meter\per\second\squared} \\
            Max./min. jerk $\Delta a^{(i)}_{\mathrm{max/min}}$ & \num{7} &\si{\meter\per\second\cubed} \\
            Max./min. orientation $\psi^{(i)}_{\mathrm{max/min}}$ & $\infty$ &\si{\radian} \\
            Max./min. angular velocity $\omega^{(i)}_{\mathrm{max/min}}$ & \num{1.57} &
           \si{\radian\per\second} \\
            Max./min. angular acceleration $\Delta \omega^{(i)}_{\mathrm{max/min}}$ & \num{0.5} &
           \si{\radian\per\second\squared} \\
            Goal reaching tolerance $\mu$ & \num{0.1} & \\
            \bottomrule
		\end{tabular}
	\end{center}
\end{table}

In Fig. \ref{fig:comparison_combined_planning_and_prediction_algorithm_8a}, the courses of the resulting trajectories of the combined algorithm in an 8-agent scenario are shown in the subfigures on the left. For comparison, the resulting trajectories under the assumption that all agents have knowledge of each other’s goals, for example due to communication at the beginning of the scenario, are given in the subfigures on the right. Our proposed approach results with \num{12.2} \si{\second} in a scenario duration that is only approximately \num{1.3} times longer than the planning approach with known goal states, which has a scenario duration of \num{9.3} \si{\second}.

\begin{figure}[!ht]
    \centering
     % Duration and driven distance
        \begin{subfigure}{0.22\textwidth}
    	\centering
        \input{img/box_plots/box_plot_duration.tex}
        \vspace{-10pt}
        \caption{Scenario duration of all vehicles (lightblue)/ one vehicle on average (lightgreen)}
    	\end{subfigure}
    	\begin{subfigure}{0.22\textwidth}
    		\centering
            \input{img/box_plots/box_plot_driven_distances.tex}
            \vspace{-10pt}
            \caption{Driven distance of all vehicles (lightblue)/ one vehicle on average (lightgreen)}
    	\end{subfigure}
    \centering
    % Minimum distance between vehicles and computational effort
    \begin{subfigure}{0.22\textwidth}
		\centering
        \input{img/box_plots/box_plot_minimum_distances_between_agents.tex}
        \vspace{-10pt}
        \caption{Minimum distances between vehicles with the red line as the collision boarder}
	\end{subfigure}
	\begin{subfigure}{0.22\textwidth}
		\centering
        \input{img/box_plots/box_plot_computation_times.tex}
        \vspace{-10pt}
        \caption{Solve times for planning (lightblue) and prediction (lightgreen)}
	\end{subfigure}
    \caption{Overall comparison of the proposed algorithm in all scenarios with exchanged (1), IOC-based estimated (2) and CA-based estimated goal states (3).}
    \label{fig:overall_comparison}
\end{figure}
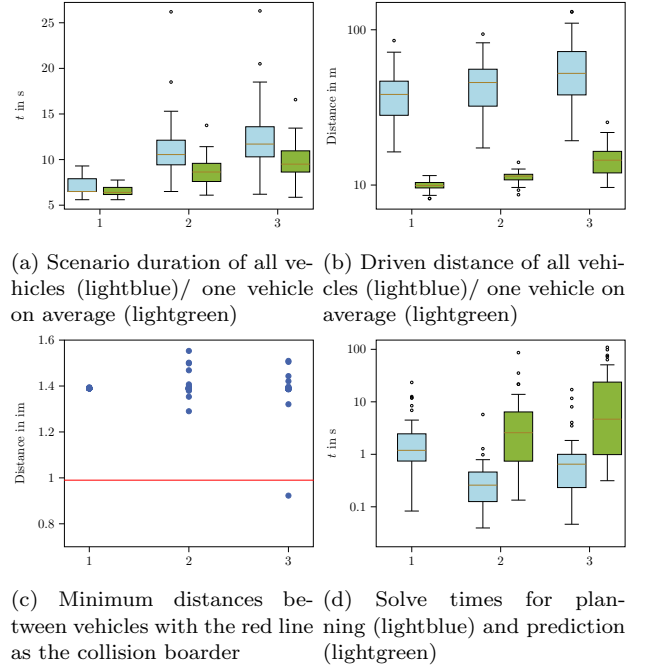

Fig. \ref{fig:overall_comparison} shows the overall comparison of all 40 simulated collision avoidance scenarios with 2-8 robots. The performances of the different methods are compared in terms of the total duration until all vehicles and one vehicle on average have reached their goal states, the driven distances of all vehicles and one vehicle on average and the minimum distances between vehicles within each scenario. Besides, the mean computation times for an agent to plan a trajectory and to predict all agents' trajectories for one time step within a scenario are given w.r.t. each approach.

Similar as in the 8-agent scenario, the scenario duration of the proposed approach with IOC-based estimated goal states increases by only approximately \num{62.3} \% considering the median value of \num{10.6} \si{\second} compared to this approach with known exchanged goal states that has a median scenario duration of \num{6.5} \si{\second}. In addition, the median duration of \num{8.6} \si{\second} until one vehicle on average has reached its goal in scenarios where the combined approach with IOC has been applied, increases by only approximately \num{35.0} \% w.r.t. the respective median duration of \num{6.4} \si{\second} of the combined approach with known exchanged goal states.

In previous work, we have simulatively evaluated a less computationally intensive MPC-based planner based on OCP (\ref{eq:planning_ocp_robot_i}) in combination with a CA-based prediction of other vehicles' trajectories, see \cite{majer2025}. There, we have found that numerous collisions occurred in the same investigated collision avoidance scenarios with 2–8 robots. In this publication, we thus decided to perform the comparison with a planner using CA-based prediction as a baseline only in combination with the joint prediction problem, see OCP (\ref{eq:predicting_ocp_robot_i}), which significantly has reduced the number of collisions. However, while our proposed IOC-based approach never results in a collision, the approach with CA-based estimations still leads to a collision in a 6-agent scenario.

If we further compare IOC-based and CA-based estimated goal states in the proposed approach, the approach with IOC outperforms the one with CA-based goal states with a median scenario duration of \num{11.7} \si{\second}. This represents a reduction of \num{9.8} \%. The improvement of the median scenario duration of a robot on average is \num{9.1} \%. Further, a reduction in the median driven distance of all vehicles of \num{12.7} \% using the proposed approach with IOC-based estimations compared to CA-based ones is achieved. The reduction in the median driven distance of one vehicle on average is \num{21.1} \%.

Another criterion is the incidence of which the solver has not been able to find a solution to the planning or prediction problem. While the proposed approach with IOC never leads to the solver finding no solution for an OCP, as in the case of known exchanged goal states, the proposed approach with CA-based estimated goal states results in the solver failing 104 times. In this case, if $v_t^{(i)} \geq 0$, the minimum acceleration and angular acceleration has been applied and if $v_t^{(i)} < 0$, the maximum values have been applied respectively in order to find a solution in the next time step.

Finally, we examine the computation times of the investigated methods, displayed in Fig. \ref{fig:overall_comparison} (d). The implementation focused on a simulation-based evaluation, without developing a real-time function. The median of the mean computation times for an agent to find a trajectory for one algorithmic time step of the combined approach with IOC within a scenario is approximately 2.4 times higher than the respective mean computation time of the baseline planning approach with initial exchange of the goal states. Thus, with the current implementation of the combined approach with IOC, predicting and replanning would not be possible in each MPC iteration.

The baseline planner operating with known communicated goal states involves solving problem (\ref{eq:predicting_ocp_robot_i}) with the real goal state parameters $\mathbf{L}^{(i)}$ with 6${\cdot}K_{\mathrm{P}}{\cdot}N$ optimization variables in each time step. The necessary algorithmic steps to cope with not knowing other robot's goal states when applying the proposed combined approach with IOC are solving problem (\ref{eq:inverse_ocp_robot_i}) for each agent with $N + 3{\cdot}N + n_c{\cdot}K_{\mathrm{E}}$ optimization variables, problem (\ref{eq:predicting_ocp_robot_i}) with 6${\cdot}K_{\mathrm{P}}{\cdot}N$ optimization variables and problem (\ref{eq:planning_ocp_robot_i}) with 6${\cdot}K_{\mathrm{P}}$ optimization variables in each time step, where $n_c$ denotes the number of constraints of problem (\ref{eq:planning_ocp_robot_i}).

%% file: img/mpc-ioc/scenario8a_mpc_ioc_planning30_safety_distance0.2_data_plot_of_paths_t16.tex
% This file was created with tikzplotlib v0.10.1.
\begin{tikzpicture}[scale=0.53]

\definecolor{brown1613334}{RGB}{161,33,34}
\definecolor{darkcyan0149129}{RGB}{0,149,129}
\definecolor{darkgoldenrod16612945}{RGB}{166,129,45}
\definecolor{darkgray176}{RGB}{176,176,176}
\definecolor{lightred}{RGB}{220,100,90}
\definecolor{gold2512280}{RGB}{251,228,0}
\definecolor{goldenrod22215426}{RGB}{222,154,26}
\definecolor{steelblue6999169}{RGB}{69,99,169}
\definecolor{yellowgreen13918159}{RGB}{139,181,59}

\pgfdeclareplotmark{myStar}{
	\node[star, draw, fill, star points=5, star point ratio=2.5, minimum size=0.25cm, inner sep = 0.3mm] {};
}

\begin{axis}[
tick align=outside,
tick pos=left,
x grid style={darkgray176},
xmin=-6, xmax=6,
xlabel={$p_x$ in m},
xtick style={color=black},
xtick={-6,-5,-4,-3,-2,-1,0,1,2,3,4,5,6},
xticklabels={-6, , , , , ,0, , , , , ,6},
y grid style={darkgray176},
ylabel={$p_y$ in m},
ymin=-6, ymax=6,
ytick style={color=black},
ytick={-6,-5,-4,-3,-2,-1,0,1,2,3,4,5,6},
yticklabels={-6, , , , , ,0, , , , , ,6}
]
\addplot [draw=darkcyan0149129, fill=darkcyan0149129, mark=*, mark size=0.5pt, only marks]
table{%
x  y
-4 4
-4 4
-3.99505063538986 3.99505063538986
-3.98013421410621 3.98028258292369
-3.95001311521805 3.95105160837882
-3.89918991269523 3.90318896196513
-3.825210102134 3.83625814153382
-3.72685029659685 3.75164230563561
-3.60273692920691 3.65118697767187
-3.45256549331041 3.53809443467605
-3.28067629548829 3.41852264264131
-3.09287385576742 3.29679212065544
-2.89407896014333 3.17832545708409
-2.68986881928642 3.06815598514374
-2.48642378846554 2.97054398204309
-2.29132536926902 2.886912636191
};
\addplot [draw=steelblue6999169, fill=steelblue6999169, mark=*, mark size=0.5pt, only marks]
table{%
x  y
0 4
0 4
-6.43586035573391e-13 3.99299999990604
0.000104977987195512 3.9720021290247
0.000672797031788106 3.93414265535512
0.00210426263901023 3.88643244765076
0.00481117768895697 3.83232871492768
0.00987518660670135 3.76492095843489
0.0176336169684737 3.69128682500988
0.0298378954382566 3.60466362937367
0.0482371598720926 3.50352810599938
0.0761442446992241 3.38156791107793
0.11657811954598 3.23254042338259
0.171060817893476 3.05637018329188
0.240479067566436 2.85365433617956
0.325002114753023 2.62559207519186
};
\addplot [draw=brown1613334, fill=brown1613334, mark=*, mark size=0.5pt, only marks]
table{%
x  y
4 4
4 4
3.99505063538986 3.99505063538986
3.98039488389662 3.98024767344021
3.9521719952083 3.95116547957166
3.90608770433228 3.90223062015554
3.84086483345737 3.83013803626454
3.75789973087718 3.73369509537553
3.65899114233327 3.61148844687786
3.54637374088752 3.46193227808604
3.42275316272125 3.28332587835744
3.29134479093359 3.0739298775615
3.15590776872559 2.83205291661706
3.01864558854431 2.55792019809446
2.88049513358545 2.25100581865348
2.74228907608935 1.91148355119531
};
\addplot [draw=goldenrod22215426, fill=goldenrod22215426, mark=*, mark size=0.5pt, only marks]
table{%
x  y
-4 0
-4 0
-3.99299999990604 -1.60683879907691e-13
-3.98599739947668 3.50129066615392e-05
-3.97199436449339 0.00014957133542229
-3.94400566093468 0.000607479664930101
-3.9027961258241 0.00182500525889212
-3.84147083963864 0.00475314321812134
-3.76707175843735 0.0100358530045755
-3.67415407988752 0.0192705729544038
-3.55603766878192 0.0349814169708317
-3.40967796477166 0.0601660905474878
-3.2352251726065 0.0980016840304019
-3.04057913917895 0.150145025290449
-2.81981339894016 0.219625636108827
-2.57713353739372 0.306312620026219
};
\addplot [draw=gold2512280, fill=gold2512280, mark=*, mark size=0.5pt, only marks]
table{%
x  y
4 0
4 0
3.99299999990604 -1.02471841117147e-13
3.9859444068464 -3.52698475969211e-05
3.97183697506185 -0.000151783743218222
3.94389179111575 -0.000613324499733479
3.89512888151205 -0.0020654080992629
3.8186241345296 -0.00574184118654287
3.71232603308451 -0.0133297849476324
3.58317934786546 -0.0262226490379824
3.43836926340391 -0.0455581462056396
3.28519620145196 -0.0720047430337676
3.11699012401659 -0.106839266420792
2.92715824011775 -0.151776665109929
2.70940323395735 -0.208697994888583
2.46386378745585 -0.277689514270475
};
\addplot [draw=lightred, fill=lightred, mark=*, mark size=0.5pt, only marks]
table{%
x  y
-4 -4
-4 -4
-3.99505063538986 -3.99505063538986
-3.98038385481785 -3.98023654171016
-3.95139092429317 -3.95036093888996
-3.90380070475112 -3.89982728124177
-3.83710364477754 -3.82610581407442
-3.75270534401379 -3.72799785580046
-3.65241397616133 -3.60408435562196
-3.5392461149417 -3.45379931435933
-3.41724851279675 -3.27754089464025
-3.28911706783702 -3.07337076782036
-3.15669122559838 -2.83687750269363
-3.0231308473636 -2.56708266015098
-2.89213412449133 -2.26306049027001
-2.76549683327796 -1.92495592717798
};
\addplot [draw=darkgoldenrod16612945, fill=darkgoldenrod16612945, mark=*, mark size=0.5pt, only marks]
table{%
x  y
0 -4
0 -4
-7.3052633138243e-13 -3.99299999990604
-0.000104889758315492 -3.97200776738102
-0.000732710932705323 -3.93013428379862
-0.00282307517687287 -3.86044114216658
-0.00780095495425121 -3.76090624587469
-0.0175071669651539 -3.63164881169682
-0.0342243885580823 -3.4729187367894
-0.0596890483983324 -3.2920990385539
-0.0952237641065639 -3.09669079184698
-0.141518499472559 -2.89428360054831
-0.200045732758659 -2.67846976430352
-0.272854933231161 -2.44292586422294
-0.359989470036326 -2.1883441939796
-0.459009382506438 -1.92102242561351
};
\addplot [draw=yellowgreen13918159, fill=yellowgreen13918159, mark=*, mark size=0.5pt, only marks]
table{%
x  y
4 -4
4 -4
3.99505063538986 -3.99505063538986
3.98013059953624 -3.98027903831048
3.95016919435965 -3.95120306687095
3.8996244179926 -3.90360268287301
3.82607017884055 -3.8370561474216
3.73350425719818 -3.75742300534324
3.61944684375507 -3.66510383118376
3.4880813052318 -3.56617354427016
3.34379370123537 -3.46628801717491
3.19112223031494 -3.37017272617875
3.03348588292465 -3.28077583747813
2.86537813129057 -3.19434454773783
2.68092133926757 -3.10775124135449
2.47425064452668 -3.01843567301109
};

% add circles
\draw[draw=darkcyan0149129,fill=none, very thick] (axis cs:-2.29132536926902,2.886912636191) circle (70.0);
\draw[draw=steelblue6999169,fill=none, very thick] (axis cs:0.325002114753023,2.62559207519186) circle (70.0);
\draw[draw=brown1613334,fill=none, very thick] (axis cs:2.74228907608935,1.91148355119531) circle (70.0);
\draw[draw=goldenrod22215426,fill=none, very thick] (axis cs:-2.57713353739372,0.306312620026219) circle (70.0);
\draw[draw=gold2512280,fill=none, very thick] (axis cs:2.46386378745585,-0.277689514270475) circle (70.0);
\draw[draw=lightred,fill=none, very thick] (axis cs:-2.76549683327796,-1.92495592717798) circle (70.0);
\draw[draw=darkgoldenrod16612945,fill=none, very thick] (axis cs:-0.459009382506438,-1.92102242561351) circle (70.0);
\draw[draw=yellowgreen13918159,fill=none, very thick] (axis cs:2.47425064452668,-3.01843567301109) circle (70.0);

% add solid stars as goal points
\addplot [draw=darkcyan0149129, fill=darkcyan0149129, mark=myStar]
table{%
	x  y
	4  -4
};
\addplot [draw=steelblue6999169, fill=steelblue6999169, mark=myStar]
table{%
	x  y
	0 -4
};
\addplot [draw=brown1613334, fill=brown1613334, mark=myStar]
table{%
	x  y
	-4  -4
};
\addplot [draw=goldenrod22215426, fill=goldenrod22215426, mark=myStar]
table{%
	x  y
	4 0
};
\addplot [draw=gold2512280, fill=gold2512280, mark=myStar]
table{%
	x  y
	-4  0
};
\addplot [draw=lightred, fill=lightred, mark=myStar]
table{%
	x  y
	4 4
};
\addplot [draw=darkgoldenrod16612945, fill=darkgoldenrod16612945, mark=myStar]
table{%
	x  y
	0  4
};
\addplot [draw=yellowgreen13918159, fill=yellowgreen13918159, mark=myStar]
table{%
	x  y
	-4 4
};

% add pale stars as goal points
\addplot [draw=darkcyan0149129, fill=darkcyan0149129, mark=myStar, opacity=0.3]
table{%
	x  y
	-0.05410642278020755  1.452140879060161
};
\addplot [draw=steelblue6999169, fill=steelblue6999169, mark=myStar, opacity=0.3]
table{%
	x  y
	1.870524254225004 0.2977961180203214
};
\addplot [draw=brown1613334, fill=brown1613334, mark=myStar, opacity=0.3]
table{%
	x  y
	0.4538196954322267  -1.987259965035476
};
\addplot [draw=goldenrod22215426, fill=goldenrod22215426, mark=myStar, opacity=0.3]
table{%
	x  y
	-0.16039424440793668 1.836396375065901
};
\addplot [draw=gold2512280, fill=gold2512280, mark=myStar, opacity=0.3]
table{%
	x  y
	0.04292315415371463  -1.2900130708173514
};
\addplot [draw=lightred, fill=lightred, mark=myStar, opacity=0.3]
table{%
	x  y
	-0.5114492626627999 1.9127957929772252
};
\addplot [draw=darkgoldenrod16612945, fill=darkgoldenrod16612945, mark=myStar, opacity=0.3]
table{%
	x  y
	-1.2848468150321328  1.67306086875278
};
\addplot [draw=yellowgreen13918159, fill=yellowgreen13918159, mark=myStar, opacity=0.3]
table{%
	x  y
	0.19551392248214816 -1.653301692500548
};

% add arrows for orientation
\path [draw=darkcyan0149129, fill=darkcyan0149129]
(axis cs:-2.2563240284887423, 2.9805871051971433)
--(axis cs:-2.3263267100492975, 2.7932381671848567)
--(axis cs:-1.7642798960124384, 2.5832301225031915)
--(axis cs:-1.8342825775729934, 2.3958811844909054)
--(axis cs:-1.354580679207588, 2.5368992283882243)
--(axis cs:-1.6242745328913282, 2.9579279985277642)
--(axis cs:-1.6942772144518832, 2.7705790605154776)
--cycle;
\path [draw=steelblue6999169, fill=steelblue6999169]
(axis cs:0.4180559270359898, 2.6622115784600147)
--(axis cs:0.23194830247005616, 2.5889725719237053)
--(axis cs:0.45166532207898297, 2.0306496982259046)
--(axis cs:0.26555769751304936, 1.9574106916895957)
--(axis cs:0.6911971474345677, 1.695053952362192)
--(axis cs:0.8238805712108502, 2.177127711298523)
--(axis cs:0.6377729466449167, 2.1038887047622135)
--cycle;
\path [draw=brown1613334, fill=brown1613334]
(axis cs:2.836313179430495, 1.8774325895217592)
--(axis cs:2.648264972748205, 1.945534512868861)
--(axis cs:2.4439592027068993, 1.3813898928219892)
--(axis cs:2.2559109960246087, 1.449491816169091)
--(axis cs:2.401779459353841, 0.9712425177838572)
--(axis cs:2.8200556160714805, 1.2451860461277857)
--(axis cs:2.63200740938919, 1.3132879694748874)
--cycle;
\path [draw=goldenrod22215426, fill=goldenrod22215426]
(axis cs:-2.613877170406551, 0.3993174878182782)
--(axis cs:-2.540389904380889, 0.21330775223415982)
--(axis cs:-1.9823606976285342, 0.4337695503111446)
--(axis cs:-1.9088734316028726, 0.2477598147270262)
--(axis cs:-1.6470848594731282, 0.673748950154527)
--(axis cs:-2.1293352296798576, 0.8057890214793813)
--(axis cs:-2.0558479636541955, 0.6197792858952629)
--cycle;
\path [draw=gold2512280, fill=gold2512280]
(axis cs:2.4921865984705365, -0.37359477156237936)
--(axis cs:2.435540976441164, -0.18178425697857065)
--(axis cs:1.8601094326897378, -0.3517211230666875)
--(axis cs:1.8034638106603658, -0.1599106084828788)
--(axis cs:1.5048112145368067, -0.5609176244173364)
--(axis cs:1.9734006767484824, -0.7353421522343049)
--(axis cs:1.9167550547191103, -0.5435316376504962)
--cycle;
\path [draw=lightred, fill=lightred]
(axis cs:-2.8607014569471922, -1.8943605086974313)
--(axis cs:-2.670292209608728, -1.9555513456585285)
--(axis cs:-2.4867196987254365, -1.3843236036431361)
--(axis cs:-2.2963104513869723, -1.4455144406042333)
--(axis cs:-2.4595426484724743, -0.972909690485659)
--(axis cs:-2.867538193402365, -1.2619419297209418)
--(axis cs:-2.6771289460639007, -1.323132766682039)
--cycle;
\path [draw=darkgoldenrod16612945, fill=darkgoldenrod16612945]
(axis cs:-0.5520703284370007, -1.9576237963871423)
--(axis cs:-0.3659484365758754, -1.8844210548398777)
--(axis cs:-0.585556661217669, -1.326055379256502)
--(axis cs:-0.3994347693565438, -1.2528526377092375)
--(axis cs:-0.8250230902427608, -0.9904129663078839)
--(axis cs:-0.9578004449399196, -1.4724608623510311)
--(axis cs:-0.7716785530787943, -1.3992581208037667)
--cycle;
\path [draw=yellowgreen13918159, fill=yellowgreen13918159]
(axis cs:2.436979039210268, -3.111230217231845)
--(axis cs:2.5115222498430922, -2.9256411287903354)
--(axis cs:1.9547549845185639, -2.7020114968918634)
--(axis cs:2.0292981951513878, -2.516422408450354)
--(axis cs:1.5463052023191328, -2.64571961984697)
--(axis cs:1.8056685632529157, -3.0731896737748823)
--(axis cs:1.8802117738857396, -2.887600585333373)
--cycle;

% add rectangles
\path [draw=darkcyan0149129, fill=none, very thick]
(axis cs:-1.8409600350165474, 3.0922685849815297)
--(axis cs:-2.08596942047849, 2.4365473019385275)
--(axis cs:-2.7416907035214924, 2.6815566874004704)
--(axis cs:-2.4966813180595495, 3.3372779704434725)
--cycle;
\path [draw=steelblue6999169, fill=none, very thick]
(axis cs:0.7788587191819474, 2.428071993640017)
--(axis cs:0.12748203320117982, 2.1717354707629357)
--(axis cs:-0.1288544896759014, 2.823112156743703)
--(axis cs:0.5225221963048662, 3.0794486796207843)
--cycle;
\path [draw=brown1613334, fill=none, very thick]
(axis cs:2.9521950719259302, 1.4632208236438735)
--(axis cs:2.2940263485379133, 1.7015775553587296)
--(axis cs:2.5323830802527696, 2.359746278746747)
--(axis cs:3.1905518036407865, 2.1213895470318906)
--cycle;
\path [draw=goldenrod22215426, fill=none, very thick]
(axis cs:-2.3802192156664206, 0.760432372843334)
--(axis cs:-2.1230137845766053, 0.10939829829891962)
--(axis cs:-2.7740478591210196, -0.1478071327908959)
--(axis cs:-3.031253290210835, 0.5032269417535185)
--cycle;
\path [draw=gold2512280, fill=none, very thick]
(axis cs:2.2273252254855866, -0.7124877533435418)
--(axis cs:2.0290655483827833, -0.04115095230021126)
--(axis cs:2.700402349426114, 0.15710872480259175)
--(axis cs:2.898662026528917, -0.5142280762407387)
--cycle;
\path [draw=lightred, fill=none, very thick]
(axis cs:-2.9916290514383523, -1.4846557796537476)
--(axis cs:-2.3251966857537276, -1.6988237090175877)
--(axis cs:-2.539364615117568, -2.3652560747022124)
--(axis cs:-3.2057969808021927, -2.151088145338372)
--cycle;
\path [draw=darkgoldenrod16612945, fill=none, very thick]
(axis cs:-0.91282749097112, -1.7234139125642538)
--(axis cs:-0.2614008694571819, -1.4672043171488278)
--(axis cs:-0.0051912740417559555, -2.118630938662766)
--(axis cs:-0.6566178955556942, -2.374840534078192)
--cycle;
\path [draw=yellowgreen13918159, fill=none, very thick]
(axis cs:2.0190191211465964, -3.2127659591762896)
--(axis cs:2.2799203583614807, -2.5632041496310065)
--(axis cs:2.9294821679067637, -2.8241053868458907)
--(axis cs:2.6685809306918795, -3.473667196391174)
--cycle;
\end{axis}

\end{tikzpicture}

%% file: img/mpc-real-goal-state/scenario8a_mpc_real-goal-state_planning30_safety_distance0.2_data_plot_of_paths_t16.tex
% This file was created with tikzplotlib v0.10.1.
\begin{tikzpicture}[scale=0.53]

\definecolor{brown1613334}{RGB}{161,33,34}
\definecolor{darkcyan0149129}{RGB}{0,149,129}
\definecolor{darkgoldenrod16612945}{RGB}{166,129,45}
\definecolor{darkgray176}{RGB}{176,176,176}
\definecolor{lightred}{RGB}{220,100,90}
\definecolor{gold2512280}{RGB}{251,228,0}
\definecolor{goldenrod22215426}{RGB}{222,154,26}
\definecolor{steelblue6999169}{RGB}{69,99,169}
\definecolor{yellowgreen13918159}{RGB}{139,181,59}

\pgfdeclareplotmark{myStar}{
	\node[star, draw, fill, star points=5, star point ratio=2.5, minimum size=0.25cm, inner sep = 0.3mm] {};
}

\begin{axis}[
tick align=outside,
tick pos=left,
x grid style={darkgray176},
xmin=-6, xmax=6,
xlabel={$p_x$ in m},
xtick style={color=black},
xtick={-6,-5,-4,-3,-2,-1,0,1,2,3,4,5,6},
xticklabels={-6, , , , , ,0, , , , , ,6},
y grid style={darkgray176},
ylabel={\, },
ymin=-6, ymax=6,
ytick style={color=black},
ytick={-6,-5,-4,-3,-2,-1,0,1,2,3,4,5,6},
yticklabels={-6, , , , , ,0, , , , , ,6}
]
\addplot [draw=darkcyan0149129, fill=darkcyan0149129, mark=*, mark size=0.5pt, only marks]
table{%
x  y
-4 4
-4 4
-3.99507514637513 3.99502565088066
-3.98045049975982 3.97995555601543
-3.9516568133705 3.94938031933761
-3.90469942323921 3.89747595507651
-3.83949114625506 3.82167274180027
-3.7577144462419 3.72063143382302
-3.66147711894839 3.5928255459744
-3.5533554900205 3.43660962051396
-3.43642581551124 3.2502858429641
-3.31591702759908 3.03928176649925
-3.18961731078021 2.79809071117563
-3.05859815100886 2.52745442111545
-2.92245778676761 2.22610916852397
-2.78107768464396 1.89430540187486
};
\addplot [draw=steelblue6999169, fill=steelblue6999169, mark=*, mark size=0.5pt, only marks]
table{%
x  y
0 4
0 4
-3.49036011776147e-05 3.9930087120137
-0.000349263620103627 3.97203023057243
-0.00160686474731761 3.93008510687223
-0.00509986883729701 3.86022962194586
-0.0125831383343907 3.76057226790399
-0.0261933101968415 3.63134993557009
-0.0484992846230012 3.47297673485171
-0.0823181015604637 3.28607544578791
-0.12882984556715 3.07111858248961
-0.187089665444767 2.83527621931824
-0.255239491721733 2.58541034247327
-0.330751495127744 2.32801117150947
-0.410688217340578 2.06909196709001
-0.491681653843179 1.81496842849176
};
\addplot [draw=brown1613334, fill=brown1613334, mark=*, mark size=0.5pt, only marks]
table{%
x  y
4 4
4 4
3.99502563031933 3.99507512627007
3.97995549343889 3.9804504403768
3.94938019166208 3.95165669711528
3.8974757370599 3.90469923494137
3.82167242803939 3.83949089253346
3.72063101535417 3.75771413348314
3.59282501638391 3.6614767584435
3.43660871815045 3.55335491863429
3.25028434695493 3.43642482828302
3.03922596786354 3.31588484382762
2.79813598730305 3.18963760645035
2.52741485540821 3.05857691231681
2.22600864836543 2.92240866762887
1.8941836284648 2.78101886261329
};
\addplot [draw=goldenrod22215426, fill=goldenrod22215426, mark=*, mark size=0.5pt, only marks]
table{%
x  y
-4 0
-4 0
-3.99300815753066 -3.49094358926463e-05
-3.97202856912092 -0.000349304435154328
-3.93008193107019 -0.00160700360826518
-3.86022452793345 -0.00510021786991318
-3.76056527112466 -0.0125838315762693
-3.63135001577469 -0.0261933039630261
-3.47298403665011 -0.0484980379612175
-3.28609002771794 -0.0823159486643129
-3.07113965908332 -0.128827556248344
-2.83530313544334 -0.187088214354805
-2.58542673318361 -0.255244073876006
-2.32776675277223 -0.330836473174353
-2.06870135371819 -0.410822873890371
-1.81453063143273 -0.491836428317428
};
\addplot [draw=gold2512280, fill=gold2512280, mark=*, mark size=0.5pt, only marks]
table{%
x  y
4 0
4 0
3.99300815753066 3.49094358926461e-05
3.97202856912092 0.000349304435154321
3.93008193107003 0.00160700360827252
3.86022452793295 0.005100217869945
3.76056527112384 0.0125838315763429
3.63135001577353 0.0261933039631648
3.47298403664862 0.0484980379614492
3.28609002771613 0.0823159486646541
3.07113965908126 0.12882755624879
2.83530313544113 0.187088214355339
2.58542673318011 0.255244073876934
2.32776675274971 0.330836473180889
2.06870135363827 0.410822873914642
1.81453062302818 0.491836430995057
};
\addplot [draw=lightred, fill=lightred, mark=*, mark size=0.5pt, only marks]
table{%
x  y
-4 -4
-4 -4
-3.99502563031933 -3.99507512627007
-3.97995549343889 -3.9804504403768
-3.94938019166208 -3.95165669711528
-3.8974757370599 -3.90469923494137
-3.82167242803939 -3.83949089253346
-3.72063101535417 -3.75771413348314
-3.59282501638391 -3.66147675844351
-3.43660871815045 -3.55335491863429
-3.25028434695492 -3.43642482828302
-3.03922596784534 -3.31588484381739
-2.798135987256 -3.18963760642537
-2.52741485505642 -3.05857691214564
-2.22600864771689 -2.92240866732418
-1.89418241328703 -2.78101834737147
};
\addplot [draw=darkgoldenrod16612945, fill=darkgoldenrod16612945, mark=*, mark size=0.5pt, only marks]
table{%
x  y
0 -4
0 -4
3.49036011777514e-05 -3.99300871201369
0.000349263620104572 -3.9720302305724
0.00160686474730431 -3.93008510687253
0.00509986883723357 -3.86022962194685
0.0125831383342413 -3.76057226790568
0.0261933101965593 -3.63134993557247
0.0484992846225287 -3.47297673485477
0.0823181015597676 -3.28607544579165
0.128829845566239 -3.07111858249385
0.187089665443717 -2.83527621932273
0.255239491807981 -2.58541034214778
0.330751495324288 -2.32801117078909
0.41068822743065 -2.06909193422801
0.491681673914423 -1.81496836414881
};
\addplot [draw=yellowgreen13918159, fill=yellowgreen13918159, mark=*, mark size=0.5pt, only marks]
table{%
x  y
4 -4
4 -4
3.99507514637513 -3.99502565088066
3.98045049975982 -3.97995555601543
3.9516568133705 -3.94938031933761
3.90469942323921 -3.89747595507651
3.83949114625506 -3.82167274180027
3.7577144462419 -3.72063143382302
3.66147711894839 -3.5928255459744
3.55335549002051 -3.43660962051396
3.43642581551124 -3.2502858429641
3.3159170275894 -3.03928176648432
3.18961731079408 -2.79809071121054
3.05859815037118 -2.52745441980584
2.9224577853236 -2.2261091654053
2.78107717194499 -1.89430418234666
};

% add circles
\draw[draw=darkcyan0149129,fill=none, very thick] (axis cs:-2.78107768464396,1.89430540187486) circle (70.0);
\draw[draw=steelblue6999169,fill=none, very thick] (axis cs:-0.491681653843179,1.81496842849176) circle (70.0);
\draw[draw=brown1613334,fill=none, very thick] (axis cs:1.8941836284648,2.78101886261329) circle (70.0);
\draw[draw=goldenrod22215426,fill=none, very thick] (axis cs:-1.81453063143273,-0.491836428317428) circle (70.0);
\draw[draw=gold2512280,fill=none, very thick] (axis cs:1.81453062302818,0.491836430995057) circle (70.0);
\draw[draw=lightred,fill=none, very thick] (axis cs:-1.89418241328703,-2.78101834737147) circle (70.0);
\draw[draw=darkgoldenrod16612945,fill=none, very thick] (axis cs:0.491681673914423,-1.81496836414881) circle (70.0);
\draw[draw=yellowgreen13918159,fill=none, very thick] (axis cs:2.78107717194499,-1.89430418234666) circle (70.0);

% add solid stars as goal points
\addplot [draw=darkcyan0149129, fill=darkcyan0149129, mark=myStar]
table{%
	x  y
	4  -4
};
\addplot [draw=steelblue6999169, fill=steelblue6999169, mark=myStar]
table{%
	x  y
	0 -4
};
\addplot [draw=brown1613334, fill=brown1613334, mark=myStar]
table{%
	x  y
	-4  -4
};
\addplot [draw=goldenrod22215426, fill=goldenrod22215426, mark=myStar]
table{%
	x  y
	4 0
};
\addplot [draw=gold2512280, fill=gold2512280, mark=myStar]
table{%
	x  y
	-4  0
};
\addplot [draw=lightred, fill=lightred, mark=myStar]
table{%
	x  y
	4 4
};
\addplot [draw=darkgoldenrod16612945, fill=darkgoldenrod16612945, mark=myStar]
table{%
	x  y
	0  4
};
\addplot [draw=yellowgreen13918159, fill=yellowgreen13918159, mark=myStar]
table{%
	x  y
	-4 4
};

% add arrows for orientation
\path [draw=darkcyan0149129, fill=darkcyan0149129]
(axis cs:-2.6884452032985866, 1.9319779834136799)
--(axis cs:-2.8737101659893334, 1.85663282033604)
--(axis cs:-2.6476746767564148, 1.300837932263799)
--(axis cs:-2.8329396394471615, 1.2254927691861595)
--(axis cs:-2.4043518692557617, 0.9679805884211248)
--(axis cs:-2.2771447513749203, 1.4515282584190783)
--(axis cs:-2.4624097140656676, 1.3761830953414387)
--cycle;
\path [draw=steelblue6999169, fill=steelblue6999169]
(axis cs:-0.3965288109127729, 1.7842123480805394)
--(axis cs:-0.5868344967735851, 1.8457245089029806)
--(axis cs:-0.7713709792409085, 1.274807451320544)
--(axis cs:-0.9616766651017207, 1.336319612142985)
--(axis cs:-0.7992424579553846, 0.8634399991876989)
--(axis cs:-0.39075960751928407, 1.1517831296756618)
--(axis cs:-0.5810652933800963, 1.213295290498103)
--cycle;
\path [draw=brown1613334, fill=brown1613334]
(axis cs:1.931856452936845, 2.6883864800664803)
--(axis cs:1.856510803992755, 2.8736512451601)
--(axis cs:1.300716508711897, 2.647614298327829)
--(axis cs:1.225370859767807, 2.8328790634214482)
--(axis cs:0.9678598029967037, 2.404290617892839)
--(axis cs:1.4514078066000775, 2.2770847681405906)
--(axis cs:1.3760621576559875, 2.46234953323421)
--cycle;
\path [draw=goldenrod22215426, fill=goldenrod22215426]
(axis cs:-1.783772621988977, -0.39668420892629663)
--(axis cs:-1.8452886408764833, -0.5869886477085594)
--(axis cs:-1.274375324529695, -0.7715367043710779)
--(axis cs:-1.335891343417201, -0.9618411431533406)
--(axis cs:-0.8630084375214164, -0.7994165227549588)
--(axis cs:-1.1513432867546827, -0.39092782680655236)
--(axis cs:-1.2128593056421888, -0.5812322655888151)
--cycle;
\path [draw=gold2512280, fill=gold2512280]
(axis cs:1.7837726135844532, 0.39668421160391704)
--(axis cs:1.8452886324719067, 0.5869886503861969)
--(axis cs:1.274375316125067, 0.771536707048557)
--(axis cs:1.3358913350125206, 0.9618411458308369)
--(axis cs:0.8630084291167808, 0.7994165254323239)
--(axis cs:1.1513432783501605, 0.3909278294839974)
--(axis cs:1.2128592972376138, 0.5812322682662772)
--cycle;
\path [draw=lightred, fill=lightred]
(axis cs:-1.9318552377612095, -2.688385964825528)
--(axis cs:-1.8565095888128504, -2.8736507299174114)
--(axis cs:-1.300715293537201, -2.6476137830723343)
--(axis cs:-1.225369644588842, -2.8328785481642176)
--(axis cs:-0.9678585878276142, -2.4042901026296746)
--(axis cs:-1.4514065914339191, -2.277084252888568)
--(axis cs:-1.37606094248556, -2.462349017980451)
--cycle;
\path [draw=darkgoldenrod16612945, fill=darkgoldenrod16612945]
(axis cs:0.39652883097586383, -1.7842122837628132)
--(axis cs:0.5868345168529822, -1.8457244445348067)
--(axis cs:0.7713709991689626, -1.2748073869034515)
--(axis cs:0.961676685046081, -1.336319547675445)
--(axis cs:0.7992424777743905, -0.8634399347632182)
--(axis cs:0.3907596274147259, -1.1517830653594645)
--(axis cs:0.5810653132918443, -1.213295226131458)
--cycle;
\path [draw=yellowgreen13918159, fill=yellowgreen13918159]
(axis cs:2.6884446896454914, -1.9319767615393977)
--(axis cs:2.8737096542444887, -1.8566316031539223)
--(axis cs:2.647674179088062, -1.3008367093569313)
--(axis cs:2.832939143687059, -1.225491550971456)
--(axis cs:2.404351380017612, -0.9679793593516751)
--(axis cs:2.277144249890068, -1.4515270261278825)
--(axis cs:2.462409214489065, -1.376181867742407)
--cycle;

% add rectangles
\path [draw=darkcyan0149129, fill=none, very thick]
(axis cs:-2.325009964549283, 1.701945752551922)
--(axis cs:-2.973437333966898, 1.4382376817801834)
--(axis cs:-3.2371454047386368, 2.086665051197798)
--(axis cs:-2.588718035321022, 2.3503731219695365)
--cycle;
\path [draw=steelblue6999169, fill=none, very thick]
(axis cs:-0.2662929850260296, 1.3742871967960666)
--(axis cs:-0.9323628855388724, 1.5895797596746106)
--(axis cs:-0.7170703226603284, 2.255649660187453)
--(axis cs:-0.05100042214748568, 2.0403570973089096)
--cycle;
\path [draw=brown1613334, fill=none, very thick]
(axis cs:1.7018251752031242, 2.324950638047299)
--(axis cs:1.4381154038988084, 2.973377315874966)
--(axis cs:2.086542081726476, 3.2370870871792814)
--(axis cs:2.3502518530307914, 2.5886604093516143)
--cycle;
\path [draw=goldenrod22215426, fill=none, very thick]
(axis cs:-1.3738448305106346, -0.26645669350160406)
--(axis cs:-1.5891508966169061, -0.9325222292395235)
--(axis cs:-2.2552164323548256, -0.717216163133252)
--(axis cs:-2.039910366248554, -0.05115062739533249)
--cycle;
\path [draw=gold2512280, fill=none, very thick]
(axis cs:1.3738448221061468, 0.2664566961791107)
--(axis cs:1.5891508882122336, 0.93252223191709)
--(axis cs:2.255216423950213, 0.7172161658110032)
--(axis cs:2.039910357844126, 0.051150630073023895)
--cycle;
\path [draw=lightred, fill=none, very thick]
(axis cs:-1.701823960035863, -2.324950122801046)
--(axis cs:-1.4381141887166062, -2.973376800622637)
--(axis cs:-2.086540866538197, -3.2370865719418935)
--(axis cs:-2.350250637857454, -2.5886598941203025)
--cycle;
\path [draw=darkgoldenrod16612945, fill=none, very thick]
(axis cs:0.26629300498045455, -1.3742871325128643)
--(axis cs:0.9323629055503686, -1.5895796952148413)
--(axis cs:0.7170703428483916, -2.2556495957847558)
--(axis cs:0.05100044227847739, -2.0403570330827785)
--cycle;
\path [draw=yellowgreen13918159, fill=none, very thick]
(axis cs:2.3250094567221633, -1.7019445214729976)
--(axis cs:2.9734368328186527, -1.438236467123833)
--(axis cs:3.237144887167817, -2.0866638432203226)
--(axis cs:2.5887175110713274, -2.3503718975694867)
--cycle;

\end{axis}

\end{tikzpicture}

%% file: img/mpc-ioc/scenario8a_mpc_ioc_planning30_safety_distance0.2_data_plot_of_paths_t30.tex
% This file was created with tikzplotlib v0.10.1.
\begin{tikzpicture}[scale=0.53]

\definecolor{brown1613334}{RGB}{161,33,34}
\definecolor{darkcyan0149129}{RGB}{0,149,129}
\definecolor{darkgoldenrod16612945}{RGB}{166,129,45}
\definecolor{darkgray176}{RGB}{176,176,176}
\definecolor{lightred}{RGB}{220,100,90}
\definecolor{gold2512280}{RGB}{251,228,0}
\definecolor{goldenrod22215426}{RGB}{222,154,26}
\definecolor{steelblue6999169}{RGB}{69,99,169}
\definecolor{yellowgreen13918159}{RGB}{139,181,59}

\pgfdeclareplotmark{myStar}{
	\node[star, draw, fill, star points=5, star point ratio=2.5, minimum size=0.25cm, inner sep = 0.3mm] {};
}

\begin{axis}[
tick align=outside,
tick pos=left,
x grid style={darkgray176},
xmin=-6, xmax=6,
xlabel={$p_x$ in m},
xtick style={color=black},
xtick={-6,-5,-4,-3,-2,-1,0,1,2,3,4,5,6},
xticklabels={-6, , , , , ,0, , , , , ,6},
y grid style={darkgray176},
ylabel={$p_y$ in m},
ymin=-6, ymax=6,
ytick style={color=black},
ytick={-6,-5,-4,-3,-2,-1,0,1,2,3,4,5,6},
yticklabels={-6, , , , , ,0, , , , , ,6}
]
\addplot [draw=darkcyan0149129, fill=darkcyan0149129, mark=*, mark size=0.5pt, only marks]
table{%
x  y
-4 4
-4 4
-3.99505063538986 3.99505063538986
-3.98013421410621 3.98028258292369
-3.95001311521805 3.95105160837882
-3.89918991269523 3.90318896196513
-3.825210102134 3.83625814153382
-3.72685029659685 3.75164230563561
-3.60273692920691 3.65118697767187
-3.45256549331041 3.53809443467605
-3.28067629548829 3.41852264264131
-3.09287385576742 3.29679212065544
-2.89407896014333 3.17832545708409
-2.68986881928642 3.06815598514374
-2.48642378846554 2.97054398204309
-2.29132536926902 2.886912636191
-2.11157112471229 2.81974770174105
-1.9480242256914 2.76822150141381
-1.79805527921942 2.73027277553149
-1.65589757456148 2.70212547572463
-1.51531937348518 2.68115197498445
-1.36978727260661 2.66570698350115
-1.21307081468324 2.65497685369064
-1.05082301183605 2.64913484867194
-0.884736594340887 2.64770306811173
-0.721916152114413 2.64994205204043
-0.569410088760657 2.65468976927326
-0.434213891013211 2.66057345834717
-0.323242535360687 2.66622278901729
-0.241932789370361 2.67055555296809
};
\addplot [draw=steelblue6999169, fill=steelblue6999169, mark=*, mark size=0.5pt, only marks]
table{%
x  y
0 4
0 4
-6.43586035573391e-13 3.99299999990604
0.000104977987195512 3.9720021290247
0.000672797031788106 3.93414265535512
0.00210426263901023 3.88643244765076
0.00481117768895697 3.83232871492768
0.00987518660670135 3.76492095843489
0.0176336169684737 3.69128682500988
0.0298378954382566 3.60466362937367
0.0482371598720926 3.50352810599938
0.0761442446992241 3.38156791107793
0.11657811954598 3.23254042338259
0.171060817893476 3.05637018329188
0.240479067566436 2.85365433617956
0.325002114753023 2.62559207519186
0.422203356104239 2.37859402062696
0.531329604837234 2.11305154415211
0.64882266793293 1.83509412501373
0.770737832712127 1.55065552908231
0.892956730840718 1.26551741976729
1.01133746537833 0.985484958385449
1.1219260850305 0.716424303983532
1.2211898743592 0.464235158259598
1.30656293312414 0.233940519433914
1.37688554845345 0.0286538013173183
1.43441516922336 -0.157284982978975
1.48111947451584 -0.329297690618955
1.51855463273549 -0.492706960487553
1.54771957384863 -0.652711324747837
};
\addplot [draw=brown1613334, fill=brown1613334, mark=*, mark size=0.5pt, only marks]
table{%
x  y
4 4
4 4
3.99505063538986 3.99505063538986
3.98039488389662 3.98024767344021
3.9521719952083 3.95116547957166
3.90608770433228 3.90223062015554
3.84086483345737 3.83013803626454
3.75789973087718 3.73369509537553
3.65899114233327 3.61148844687786
3.54637374088752 3.46193227808604
3.42275316272125 3.28332587835744
3.29134479093359 3.0739298775615
3.15590776872559 2.83205291661706
3.01864558854431 2.55792019809446
2.88049513358545 2.25100581865348
2.74228907608935 1.91148355119531
2.60725206182456 1.53860908421735
2.47984288722069 1.13884879542922
2.36172658273819 0.719596052778513
2.25359490816334 0.28837236247698
2.15536918882501 -0.147267906418214
2.06641111902218 -0.579794703168601
1.98573673893556 -1.00173566779105
1.91223102707671 -1.4056952734627
1.84486359249524 -1.78435262559984
1.78237362367522 -2.13341978711008
1.7232292551924 -2.45260944085925
1.66634576821968 -2.74169058221183
1.61108743921787 -3.00048277118962
1.55726190782548 -3.22888718612056
};
\addplot [draw=goldenrod22215426, fill=goldenrod22215426, mark=*, mark size=0.5pt, only marks]
table{%
x  y
-4 0
-4 0
-3.99299999990604 -1.60683879907691e-13
-3.98599739947668 3.50129066615392e-05
-3.97199436449339 0.00014957133542229
-3.94400566093468 0.000607479664930101
-3.9027961258241 0.00182500525889212
-3.84147083963864 0.00475314321812134
-3.76707175843735 0.0100358530045755
-3.67415407988752 0.0192705729544038
-3.55603766878192 0.0349814169708317
-3.40967796477166 0.0601660905474878
-3.2352251726065 0.0980016840304019
-3.04057913917895 0.150145025290449
-2.81981339894016 0.219625636108827
-2.57713353739372 0.306312620026219
-2.31954183319065 0.408079928244419
-2.04208584954373 0.526835078958769
-1.75119550442474 0.659396402588745
-1.45160052667812 0.802554343424475
-1.14909421512102 0.952028535029536
-0.849247293461227 1.1032506474467
-0.557444329093792 1.25158136610381
-0.277157156249936 1.39341750716071
-0.00338332466635427 1.52962277072152
0.258182986347758 1.65592443467456
0.501574593227612 1.76844596684831
0.720492605021245 1.8639162346097
0.914229320816495 1.94229765258637
1.08231287594434 2.0041613984071
};
\addplot [draw=gold2512280, fill=gold2512280, mark=*, mark size=0.5pt, only marks]
table{%
x  y
4 0
4 0
3.99299999990604 -1.02471841117147e-13
3.9859444068464 -3.52698475969211e-05
3.97183697506185 -0.000151783743218222
3.94389179111575 -0.000613324499733479
3.89512888151205 -0.0020654080992629
3.8186241345296 -0.00574184118654287
3.71232603308451 -0.0133297849476324
3.58317934786546 -0.0262226490379824
3.43836926340391 -0.0455581462056396
3.28519620145196 -0.0720047430337676
3.11699012401659 -0.106839266420792
2.92715824011775 -0.151776665109929
2.70940323395735 -0.208697994888583
2.46386378745585 -0.277689514270475
2.19221597597673 -0.357912745450165
1.89254617459046 -0.449104120295064
1.56984458309918 -0.54845271997297
1.23031808000466 -0.652333080224351
0.880108953995962 -0.756906107756368
0.525400655633006 -0.858303262161209
0.172467918881511 -0.952836361236203
-0.172205571213374 -1.03717728899596
-0.501894706710734 -1.10857319703858
-0.809600296626566 -1.16506424982475
-1.08964872868543 -1.20592708592423
-1.34149651877385 -1.23202093535201
-1.56437467208189 -1.2446586277522
-1.7603790769399 -1.24566422155393
};
\addplot [draw=lightred, fill=lightred, mark=*, mark size=0.5pt, only marks]
table{%
x  y
-4 -4
-4 -4
-3.99505063538986 -3.99505063538986
-3.98038385481785 -3.98023654171016
-3.95139092429317 -3.95036093888996
-3.90380070475112 -3.89982728124177
-3.83710364477754 -3.82610581407442
-3.75270534401379 -3.72799785580046
-3.65241397616133 -3.60408435562196
-3.5392461149417 -3.45379931435933
-3.41724851279675 -3.27754089464025
-3.28911706783702 -3.07337076782036
-3.15669122559838 -2.83687750269363
-3.0231308473636 -2.56708266015098
-2.89213412449133 -2.26306049027001
-2.76549683327796 -1.92495592717798
-2.64585589964039 -1.55266587108212
-2.53601644459534 -1.1534594066841
-2.4375326623897 -0.734846739563112
-2.35090229214507 -0.304437131508846
-2.27577254573342 0.130161449283094
-2.21115396538689 0.561396310525591
-2.15563631962877 0.981802251420252
-2.10760539128253 1.38400709199659
-2.06545846738678 1.76072940894226
-2.02749416073179 2.10773169547669
-1.99198095299298 2.42482505893324
-1.95763488842401 2.71186843601579
-1.92362295279058 2.96873641333537
-1.88956358754578 3.19532153024933
};
\addplot [draw=darkgoldenrod16612945, fill=darkgoldenrod16612945, mark=*, mark size=0.5pt, only marks]
table{%
x  y
0 -4
0 -4
-7.3052633138243e-13 -3.99299999990604
-0.000104889758315492 -3.97200776738102
-0.000732710932705323 -3.93013428379862
-0.00282307517687287 -3.86044114216658
-0.00780095495425121 -3.76090624587469
-0.0175071669651539 -3.63164881169682
-0.0342243885580823 -3.4729187367894
-0.0596890483983324 -3.2920990385539
-0.0952237641065639 -3.09669079184698
-0.141518499472559 -2.89428360054831
-0.200045732758659 -2.67846976430352
-0.272854933231161 -2.44292586422294
-0.359989470036326 -2.1883441939796
-0.459009382506438 -1.92102242561351
-0.567221761486127 -1.64588665511114
-0.681794407676639 -1.36692469670353
-0.798859920170129 -1.08982144898658
-0.915101331716478 -0.820795573496976
-1.02653350421181 -0.565748815589488
-1.13432024488977 -0.318429687184043
-1.24003086475645 -0.0719105915141691
-1.34494840063491 0.180292112565463
-1.44508688473846 0.432205257901649
-1.53678091077788 0.677672226131453
-1.61688559930504 0.910286561386112
-1.68296597935964 1.12330185873741
-1.73355268146762 1.30973323932439
-1.76986640268662 1.46902103226509
};
\addplot [draw=yellowgreen13918159, fill=yellowgreen13918159, mark=*, mark size=0.5pt, only marks]
table{%
x  y
4 -4
4 -4
3.99505063538986 -3.99505063538986
3.98013059953624 -3.98027903831048
3.95016919435965 -3.95120306687095
3.8996244179926 -3.90360268287301
3.82607017884055 -3.8370561474216
3.73350425719818 -3.75742300534324
3.61944684375507 -3.66510383118376
3.4880813052318 -3.56617354427016
3.34379370123537 -3.46628801717491
3.19112223031494 -3.37017272617875
3.03348588292465 -3.28077583747813
2.86537813129057 -3.19434454773783
2.68092133926757 -3.10775124135449
2.47425064452668 -3.01843567301109
2.24405214702977 -2.92597476795265
1.99380957996102 -2.83436022916342
1.72754457418105 -2.7476091124405
1.45242516155868 -2.66731209837039
1.17573996538833 -2.59430399806183
0.904815209163453 -2.52886540267068
0.646964979596802 -2.47093153231871
0.409440610378777 -2.42029988845869
0.199443365531454 -2.37684932512138
0.0187684832851382 -2.33965236761734
-0.132421104501857 -2.30789342374091
-0.255600652470607 -2.28085750334447
-0.350994155870668 -2.25851518766625
-0.418915211648602 -2.24123922887126
};

% add circles
\draw[draw=darkcyan0149129,fill=none, very thick] (axis cs:-0.241932789370361,2.67055555296809) circle (70.0);
\draw[draw=steelblue6999169,fill=none, very thick] (axis cs:1.54771957384863,-0.652711324747837) circle (70.0);
\draw[draw=brown1613334,fill=none, very thick] (axis cs:1.55726190782548,-3.22888718612056) circle (70.0);
\draw[draw=goldenrod22215426,fill=none, very thick] (axis cs:1.08231287594434,2.0041613984071) circle (70.0);
\draw[draw=gold2512280,fill=none, very thick] (axis cs:-1.7603790769399,-1.24566422155393) circle (70.0);
\draw[draw=lightred,fill=none, very thick] (axis cs:-1.88956358754578,3.19532153024933) circle (70.0);
\draw[draw=darkgoldenrod16612945,fill=none, very thick] (axis cs:-1.76986640268662,1.46902103226509) circle (70.0);
\draw[draw=yellowgreen13918159,fill=none, very thick] (axis cs:-0.418915211648602,-2.24123922887126) circle (70.0);

% add solid stars as goal points
\addplot [draw=darkcyan0149129, fill=darkcyan0149129, mark=myStar]
table{%
	x  y
	4  -4
};
\addplot [draw=steelblue6999169, fill=steelblue6999169, mark=myStar]
table{%
	x  y
	0 -4
};
\addplot [draw=brown1613334, fill=brown1613334, mark=myStar]
table{%
	x  y
	-4  -4
};
\addplot [draw=goldenrod22215426, fill=goldenrod22215426, mark=myStar]
table{%
	x  y
	4 0
};
\addplot [draw=gold2512280, fill=gold2512280, mark=myStar]
table{%
	x  y
	-4  0
};
\addplot [draw=lightred, fill=lightred, mark=myStar]
table{%
	x  y
	4 4
};
\addplot [draw=darkgoldenrod16612945, fill=darkgoldenrod16612945, mark=myStar]
table{%
	x  y
	0  4
};
\addplot [draw=yellowgreen13918159, fill=yellowgreen13918159, mark=myStar]
table{%
	x  y
	-4 4
};

% add pale stars as goal points
\addplot [draw=darkcyan0149129, fill=darkcyan0149129, mark=myStar, opacity=0.3]
table{%
	x  y
	0.6671248592788114  2.60447851828168
};
\addplot [draw=steelblue6999169, fill=steelblue6999169, mark=myStar, opacity=0.3]
table{%
	x  y
	2.232990785644596 -2.2512784213421964
};
\addplot [draw=brown1613334, fill=brown1613334, mark=myStar, opacity=0.3]
table{%
	x  y
	1.0097449040725057  -5.769374063591407
};
\addplot [draw=goldenrod22215426, fill=goldenrod22215426, mark=myStar, opacity=0.3]
table{%
	x  y
	2.916001279323902 2.923572953995362
};
\addplot [draw=gold2512280, fill=gold2512280, mark=myStar, opacity=0.3]
table{%
	x  y
	-3.9975095193581733  -1.8944356074572115
};
\addplot [draw=lightred, fill=lightred, mark=myStar, opacity=0.3]
table{%
	x  y
	-1.463926723708986 5.712389907631509
};
\addplot [draw=darkgoldenrod16612945, fill=darkgoldenrod16612945, mark=myStar, opacity=0.3]
table{%
	x  y
	-2.42137655085253  3.074734092565938
};
\addplot [draw=yellowgreen13918159, fill=yellowgreen13918159, mark=myStar, opacity=0.3]
table{%
	x  y
	-1.2037746230535553 -2.039749795266362
};

% add arrows for orientation
\path [draw=darkcyan0149129, fill=darkcyan0149129]
(axis cs:-0.24699157860566073, 2.7704275142558954)
--(axis cs:-0.23687400013506127, 2.570683591680285)
--(axis cs:0.36235776759176863, 2.6010363270920833)
--(axis cs:0.3724753460623681, 2.4012924045164734)
--(axis cs:0.7567868235076889, 2.721143445321087)
--(axis cs:0.3421226106505698, 3.0005241722433036)
--(axis cs:0.3522401891211693, 2.8007802496676937)
--cycle;
\path [draw=steelblue6999169, fill=steelblue6999169]
(axis cs:1.6468706794246977, -0.6397090840546527)
--(axis cs:1.4485684682725624, -0.6657135654410213)
--(axis cs:1.5265819124316686, -1.2606201988974268)
--(axis cs:1.3282797012795333, -1.2866246802837955)
--(axis cs:1.6777419807804734, -1.6442223805085128)
--(axis cs:1.9231863347359388, -1.2086112361246895)
--(axis cs:1.7248841235838037, -1.2346157175110581)
--cycle;
\path [draw=brown1613334, fill=brown1613334]
(axis cs:1.6539646955764624, -3.254354240154909)
--(axis cs:1.4605591200744974, -3.2034201320862112)
--(axis cs:1.3077567958684042, -3.783636858592106)
--(axis cs:1.1143512203664394, -3.732702750523408)
--(axis cs:1.3025913674819911, -4.195915063630385)
--(axis cs:1.694567946872334, -3.8855050747295015)
--(axis cs:1.501162371370369, -3.8345709666608037)
--cycle;
\path [draw=goldenrod22215426, fill=goldenrod22215426]
(axis cs:1.0512454060959393, 2.0992130281523007)
--(axis cs:1.1133803457927405, 1.9091097686618994)
--(axis cs:1.6836901242639444, 2.0955145877523034)
--(axis cs:1.745825063960746, 1.9054113282619023)
--(axis cs:2.0328291733963466, 2.314836096891107)
--(axis cs:1.5594202448703416, 2.475721106733106)
--(axis cs:1.6215551845671432, 2.2856178472427047)
--cycle;
\path [draw=gold2512280, fill=gold2512280]
(axis cs:-1.7655083804946288, -1.3455325861397358)
--(axis cs:-1.755249773385171, -1.1457958569681241)
--(axis cs:-2.354459960900006, -1.1150200356397506)
--(axis cs:-2.3442013537905484, -0.9152833064681388)
--(axis cs:-2.759062722797959, -1.1943711860066408)
--(axis cs:-2.374977175118922, -1.5144934939829742)
--(axis cs:-2.3647185680094642, -1.3147567648113625)
--cycle;
\path [draw=lightred, fill=lightred]
(axis cs:-1.9880922541505208, 3.2124125178478224)
--(axis cs:-1.791034920941039, 3.1782305426508377)
--(axis cs:-1.6884889953500837, 3.7694025422792823)
--(axis cs:-1.4914316621406019, 3.7352205670822975)
--(axis cs:-1.718653711560854, 4.180608196296738)
--(axis cs:-2.082603661769047, 3.8377664926732526)
--(axis cs:-1.8855463285595653, 3.803584517476268)
--cycle;
\path [draw=darkgoldenrod16612945, fill=darkgoldenrod16612945]
(axis cs:-1.8682801182012834, 1.451280102227493)
--(axis cs:-1.6714526871719566, 1.486761962302687)
--(axis cs:-1.777898267397539, 2.077244255390667)
--(axis cs:-1.5810708363682122, 2.112726115465861)
--(axis cs:-1.9472757030625905, 2.4531581874117236)
--(axis cs:-2.1715531294561923, 2.006280535240279)
--(axis cs:-1.9747256984268655, 2.041762395315473)
--cycle;
\path [draw=yellowgreen13918159, fill=yellowgreen13918159]
(axis cs:-0.4458848982750683, -2.33753375625091)
--(axis cs:-0.39194552502213575, -2.14494470149161)
--(axis cs:-0.9697126893000372, -1.983126581732812)
--(axis cs:-0.9157733160471047, -1.7905375269735118)
--(axis cs:-1.3818604854451044, -1.9715423626065975)
--(axis cs:-1.0775914358059022, -2.368304691251413)
--(axis cs:-1.0236520625529697, -2.1757156364921126)
--cycle;

% add rectangles
\path [draw=darkcyan0149129, fill=none, very thick]
(axis cs:0.08991331281340742, 3.037813179798957)
--(axis cs:0.1253248374605055, 2.3387094507843216)
--(axis cs:-0.5737788915541294, 2.3032979261372235)
--(axis cs:-0.6091904162012275, 3.002401655151859)
--cycle;
\path [draw=steelblue6999169, fill=none, very thick]
(axis cs:1.9402562857910117, -0.9542323518379283)
--(axis cs:1.2461985467585388, -1.0452480366902188)
--(axis cs:1.1551828619062483, -0.35119029765774573)
--(axis cs:1.8492406009387212, -0.26017461280545523)
--cycle;
\path [draw=brown1613334, fill=none, very thick]
(axis cs:1.8065869758336972, -3.6564816323692195)
--(axis cs:1.1296674615768203, -3.4782122541287777)
--(axis cs:1.3079368398172626, -2.8012927398719007)
--(axis cs:1.9848563540741395, -2.9795621181123426)
--cycle;
\path [draw=goldenrod22215426, fill=none, very thick]
(axis cs:1.3062574355831398, 2.445578246984705)
--(axis cs:1.5237297245219446, 1.7802168387683002)
--(axis cs:0.85836831630554, 1.5627445498294954)
--(axis cs:0.6408960273667352, 2.2281059580458997)
--cycle;
\path [draw=gold2512280, fill=none, very thick]
(axis cs:-2.127870915431772, -1.5772509351626993)
--(axis cs:-2.0919657905486693, -0.8781723830620582)
--(axis cs:-1.3928872384480282, -0.9140775079451606)
--(axis cs:-1.4287923633311306, -1.6131560600458017)
--cycle;
\path [draw=lightred, fill=none, very thick]
(axis cs:-2.174595464067649, 3.599990319960647)
--(axis cs:-1.484894797834463, 3.480353406771199)
--(axis cs:-1.604531711023911, 2.7906527405380133)
--(axis cs:-2.2942323772570967, 2.910289653727461)
--cycle;
\path [draw=darkgoldenrod16612945, fill=none, very thick]
(axis cs:-2.1764076621195314, 1.751375781434822)
--(axis cs:-1.487511653516888, 1.8755622916980013)
--(axis cs:-1.3633251432537086, 1.186666283095358)
--(axis cs:-2.052221151856352, 1.0624797728321789)
--cycle;
\path [draw=yellowgreen13918159, fill=none, very thick]
(axis cs:-0.8503399606700097, -2.483876171507404)
--(axis cs:-0.661552154284746, -1.809814479849852)
--(axis cs:0.01250953737280569, -1.998602286235116)
--(axis cs:-0.17627826901245802, -2.6726639778926677)
--cycle;
\end{axis}

\end{tikzpicture}

%% file: img/mpc-real-goal-state/scenario8a_mpc_real-goal-state_planning30_safety_distance0.2_data_plot_of_paths_t30.tex
% This file was created with tikzplotlib v0.10.1.
\begin{tikzpicture}[scale=0.53]

\definecolor{brown1613334}{RGB}{161,33,34}
\definecolor{darkcyan0149129}{RGB}{0,149,129}
\definecolor{darkgoldenrod16612945}{RGB}{166,129,45}
\definecolor{darkgray176}{RGB}{176,176,176}
\definecolor{lightred}{RGB}{220,100,90}
\definecolor{gold2512280}{RGB}{251,228,0}
\definecolor{goldenrod22215426}{RGB}{222,154,26}
\definecolor{steelblue6999169}{RGB}{69,99,169}
\definecolor{yellowgreen13918159}{RGB}{139,181,59}

\pgfdeclareplotmark{myStar}{
	\node[star, draw, fill, star points=5, star point ratio=2.5, minimum size=0.25cm, inner sep = 0.3mm] {};
}

\begin{axis}[
tick align=outside,
tick pos=left,
x grid style={darkgray176},
xmin=-6, xmax=6,
xlabel={$p_x$ in m},
xtick style={color=black},
xtick={-6,-5,-4,-3,-2,-1,0,1,2,3,4,5,6},
xticklabels={-6, , , , , ,0, , , , , ,6},
y grid style={darkgray176},
ylabel={\, },
ymin=-6, ymax=6,
ytick style={color=black},
ytick={-6,-5,-4,-3,-2,-1,0,1,2,3,4,5,6},
yticklabels={-6, , , , , ,0, , , , , ,6}
]
\addplot [draw=darkcyan0149129, fill=darkcyan0149129, mark=*, mark size=0.5pt, only marks]
table{%
x  y
-4 4
-4 4
-3.99507514637513 3.99502565088066
-3.98045049975982 3.97995555601543
-3.9516568133705 3.94938031933761
-3.90469942323921 3.89747595507651
-3.83949114625506 3.82167274180027
-3.7577144462419 3.72063143382302
-3.66147711894839 3.5928255459744
-3.5533554900205 3.43660962051396
-3.43642581551124 3.2502858429641
-3.31591702759908 3.03928176649925
-3.18961731078021 2.79809071117563
-3.05859815100886 2.52745442111545
-2.92245778676761 2.22610916852397
-2.78107768464396 1.89430540187486
-2.63394893357615 1.53253297397226
-2.48263110614505 1.14779728416462
-2.32811777044959 0.747267070698717
-2.17103869410679 0.338212188347461
-2.01186547424514 -0.0720478515639643
-1.85111966796691 -0.476187754038253
-1.68956842648475 -0.866956022661097
-1.52844585959403 -1.23718271454549
-1.36961768108775 -1.57993573497764
-1.21447001675003 -1.89121180543403
-1.06325047748852 -2.17077568946349
-0.916585490024998 -2.41864862188677
-0.774015148861154 -2.63733920452232
-0.633842546467272 -2.83114932076735
};
\addplot [draw=steelblue6999169, fill=steelblue6999169, mark=*, mark size=0.5pt, only marks]
table{%
x  y
0 4
0 4
-3.49036011776147e-05 3.9930087120137
-0.000349263620103627 3.97203023057243
-0.00160686474731761 3.93008510687223
-0.00509986883729701 3.86022962194586
-0.0125831383343907 3.76057226790399
-0.0261933101968415 3.63134993557009
-0.0484992846230012 3.47297673485171
-0.0823181015604637 3.28607544578791
-0.12882984556715 3.07111858248961
-0.187089665444767 2.83527621931824
-0.255239491721733 2.58541034247327
-0.330751495127744 2.32801117150947
-0.410688217340578 2.06909196709001
-0.491681653843179 1.81496842849176
-0.570254846835691 1.5718794980626
-0.643011550471424 1.34608368771133
-0.709606092764417 1.13514934491239
-0.771270333819266 0.932143290534008
-0.829136981897014 0.730137417767263
-0.884033891314172 0.522168736011604
-0.936274682666398 0.301218090968666
-0.98542542109457 0.0603190665834885
-1.02836629333916 -0.196985402521463
-1.0617694120189 -0.463883700740084
-1.08301440814363 -0.733291581262609
-1.09044108308327 -0.998044074713679
-1.08350341161807 -1.2515372105474
-1.06278025144724 -1.48839240226316
};
\addplot [draw=brown1613334, fill=brown1613334, mark=*, mark size=0.5pt, only marks]
table{%
x  y
4 4
4 4
3.99502563031933 3.99507512627007
3.97995549343889 3.9804504403768
3.94938019166208 3.95165669711528
3.8974757370599 3.90469923494137
3.82167242803939 3.83949089253346
3.72063101535417 3.75771413348314
3.59282501638391 3.6614767584435
3.43660871815045 3.55335491863429
3.25028434695493 3.43642482828302
3.03922596786354 3.31588484382762
2.79813598730305 3.18963760645035
2.52741485540821 3.05857691231681
2.22600864836543 2.92240866762887
1.8941836284648 2.78101886261329
1.53239969729141 2.63388432757081
1.14766242266477 2.48256427891924
0.74714061838308 2.32805208384323
0.338103178465866 2.17097710126602
-0.0721303741967428 2.01181107593804
-0.476234857804333 1.85107584307614
-0.866964531388321 1.68953688240656
-1.23714747423566 1.5284295596703
-1.57985172477097 1.36962014541528
-1.89108024427833 1.21449237856168
-2.1705975146868 1.06329431023687
-2.41842569027449 0.916652147380051
-2.63709065010327 0.774094925137855
-2.83089437987092 0.633923375523412
};
\addplot [draw=goldenrod22215426, fill=goldenrod22215426, mark=*, mark size=0.5pt, only marks]
table{%
x  y
-4 0
-4 0
-3.99300815753066 -3.49094358926463e-05
-3.97202856912092 -0.000349304435154328
-3.93008193107019 -0.00160700360826518
-3.86022452793345 -0.00510021786991318
-3.76056527112466 -0.0125838315762693
-3.63135001577469 -0.0261933039630261
-3.47298403665011 -0.0484980379612175
-3.28609002771794 -0.0823159486643129
-3.07113965908332 -0.128827556248344
-2.83530313544334 -0.187088214354805
-2.58542673318361 -0.255244073876006
-2.32776675277223 -0.330836473174353
-2.06870135371819 -0.410822873890371
-1.81453063143273 -0.491836428317428
-1.57149269835872 -0.57039857945289
-1.34583527101887 -0.643116274440134
-1.13501202839374 -0.70968143320216
-0.932092076457901 -0.771325507865053
-0.730146956502016 -0.829181194300506
-0.522212733084126 -0.884076104871031
-0.301268438287986 -0.93632339341332
-0.0603564646023237 -0.98548601223325
0.196955489661008 -1.0284385317152
0.463855552157082 -1.06185325959393
0.733258647347439 -1.08310998935763
0.997998812492106 -1.09054881649954
1.25146950481004 -1.08362432370442
1.48829008105535 -1.0629166068913
};
\addplot [draw=gold2512280, fill=gold2512280, mark=*, mark size=0.5pt, only marks]
table{%
x  y
4 0
4 0
3.99300815753066 3.49094358926461e-05
3.97202856912092 0.000349304435154321
3.93008193107003 0.00160700360827252
3.86022452793295 0.005100217869945
3.76056527112384 0.0125838315763429
3.63135001577353 0.0261933039631648
3.47298403664862 0.0484980379614492
3.28609002771613 0.0823159486646541
3.07113965908126 0.12882755624879
2.83530313544113 0.187088214355339
2.58542673318011 0.255244073876934
2.32776675274971 0.330836473180889
2.06870135363827 0.410822873914642
1.81453062302818 0.491836430995057
1.57149268622269 0.570398583336649
1.34583580620736 0.643116101932443
1.13501288506022 0.709681159159617
0.932093036245594 0.771325202385443
0.730147801638118 0.82918092146992
0.522213260810121 0.884075915543229
0.301268474134293 0.936323319996835
0.0603558493690617 0.985486071166642
-0.196957011212529 1.02843874130116
-0.463858238448488 1.06185361526911
-0.73326275281769 1.08311045797029
-0.998004573250293 1.09054933131056
-1.2514770334103 1.08362478800575
-1.48829920540904 1.06291692704848
};
\addplot [draw=lightred, fill=lightred, mark=*, mark size=0.5pt, only marks]
table{%
x  y
-4 -4
-4 -4
-3.99502563031933 -3.99507512627007
-3.97995549343889 -3.9804504403768
-3.94938019166208 -3.95165669711528
-3.8974757370599 -3.90469923494137
-3.82167242803939 -3.83949089253346
-3.72063101535417 -3.75771413348314
-3.59282501638391 -3.66147675844351
-3.43660871815045 -3.55335491863429
-3.25028434695492 -3.43642482828302
-3.03922596784534 -3.31588484381739
-2.798135987256 -3.18963760642537
-2.52741485505642 -3.05857691214564
-2.22600864771689 -2.92240866732418
-1.89418241328703 -2.78101834737147
-1.53239761209365 -2.63388345848934
-1.1476598164865 -2.48256320205694
-0.747137843688546 -2.32805093602335
-0.338100569211753 -2.17097600788132
0.0721324853743392 -2.01181016365184
0.476236140602198 -1.85107524525247
0.866964644070761 -1.68953675046231
1.23714608123854 -1.52843006272135
1.57984849123858 -1.36962147909385
1.89107521298979 -1.21449458482796
2.17059072253365 -1.06329744471803
2.41841720124744 -0.916656261144974
2.63708111571972 -0.774099695200354
2.83088450349259 -0.633928367004264
};
\addplot [draw=darkgoldenrod16612945, fill=darkgoldenrod16612945, mark=*, mark size=0.5pt, only marks]
table{%
x  y
0 -4
0 -4
3.49036011777514e-05 -3.99300871201369
0.000349263620104572 -3.9720302305724
0.00160686474730431 -3.93008510687253
0.00509986883723357 -3.86022962194685
0.0125831383342413 -3.76057226790568
0.0261933101965593 -3.63134993557247
0.0484992846225287 -3.47297673485477
0.0823181015597676 -3.28607544579165
0.128829845566239 -3.07111858249385
0.187089665443717 -2.83527621932273
0.255239491807981 -2.58541034214778
0.330751495324288 -2.32801117078909
0.41068822743065 -2.06909193422801
0.491681673914423 -1.81496836414881
0.570254881262706 -1.57187938908572
0.643011645149782 -1.34608339149202
0.709606242283502 -1.13514887548709
0.77127052929165 -0.93214267301349
0.82913719331823 -0.730136750822773
0.88403408630681 -0.522168141410132
0.93627481397365 -0.301217781288159
0.985425457744819 -0.060319244821679
1.02836618600829 0.196984324682601
1.06176914576956 0.463881304593688
1.08301401119373 0.733287450139573
1.09044063474478 0.998037810947
1.0835030396753 1.25152856584784
1.06278009192375 1.48838149514434
};
\addplot [draw=yellowgreen13918159, fill=yellowgreen13918159, mark=*, mark size=0.5pt, only marks]
table{%
x  y
4 -4
4 -4
3.99507514637513 -3.99502565088066
3.98045049975982 -3.97995555601543
3.9516568133705 -3.94938031933761
3.90469942323921 -3.89747595507651
3.83949114625506 -3.82167274180027
3.7577144462419 -3.72063143382302
3.66147711894839 -3.5928255459744
3.55335549002051 -3.43660962051396
3.43642581551124 -3.2502858429641
3.3159170275894 -3.03928176648432
3.18961731079408 -2.79809071121054
3.05859815037118 -2.52745441980584
2.9224577853236 -2.2261091654053
2.78107717194499 -1.89430418234666
2.63394809373568 -1.53253092378536
2.48263010848265 -1.1477947958048
2.32811677499691 -0.747264540367985
2.1710378570391 -0.338210012619877
2.01186495536355 0.0720492731307811
1.85111963500391 0.476188025877365
1.68956904097562 0.866954802637512
1.52844729002826 1.23717969535643
1.36962012211556 1.57993060864858
1.21447351274682 1.89120463284125
1.06325509152904 2.17076651573925
0.916591277435867 2.41863752548072
0.77402171074943 2.6373269750745
0.633849349116898 2.83113680864751
};

% add circles
\draw[draw=darkcyan0149129,fill=none, very thick] (axis cs:-0.633842546467272,-2.83114932076735) circle (70.0);
\draw[draw=steelblue6999169,fill=none, very thick] (axis cs:-1.06278025144724,-1.48839240226316) circle (70.0);
\draw[draw=brown1613334,fill=none, very thick] (axis cs:-2.83089437987092,0.633923375523412) circle (70.0);
\draw[draw=goldenrod22215426,fill=none, very thick] (axis cs:1.48829008105535,-1.0629166068913) circle (70.0);
\draw[draw=gold2512280,fill=none, very thick] (axis cs:-1.48829920540904,1.06291692704848) circle (70.0);
\draw[draw=lightred,fill=none, very thick] (axis cs:2.83088450349259,-0.633928367004264) circle (70.0);
\draw[draw=darkgoldenrod16612945,fill=none, very thick] (axis cs:1.06278009192375,1.48838149514434) circle (70.0);
\draw[draw=yellowgreen13918159,fill=none, very thick] (axis cs:0.633849349116898,2.83113680864751) circle (70.0);

% add solid stars as goal points
\addplot [draw=darkcyan0149129, fill=darkcyan0149129, mark=myStar]
table{%
	x  y
	4  -4
};
\addplot [draw=steelblue6999169, fill=steelblue6999169, mark=myStar]
table{%
	x  y
	0 -4
};
\addplot [draw=brown1613334, fill=brown1613334, mark=myStar]
table{%
	x  y
	-4  -4
};
\addplot [draw=goldenrod22215426, fill=goldenrod22215426, mark=myStar]
table{%
	x  y
	4 0
};
\addplot [draw=gold2512280, fill=gold2512280, mark=myStar]
table{%
	x  y
	-4  0
};
\addplot [draw=lightred, fill=lightred, mark=myStar]
table{%
	x  y
	4 4
};
\addplot [draw=darkgoldenrod16612945, fill=darkgoldenrod16612945, mark=myStar]
table{%
	x  y
	0  4
};
\addplot [draw=yellowgreen13918159, fill=yellowgreen13918159, mark=myStar]
table{%
	x  y
	-4 4
};

% add arrows for orientation
\path [draw=darkcyan0149129, fill=darkcyan0149129]
(axis cs:-0.5560589164770882, -2.768302627094968)
--(axis cs:-0.7116261764574559, -2.8939960144397316)
--(axis cs:-0.3345460144231662, -3.360697794380835)
--(axis cs:-0.49011327440353397, -3.486391181725598)
--(axis cs:-0.005375609743455967, -3.6089856206691886)
--(axis cs:-0.023411494462430738, -3.109311019691308)
--(axis cs:-0.17897875444279854, -3.2350044070360715)
--cycle;
\path [draw=steelblue6999169, fill=steelblue6999169]
(axis cs:-0.9638811101842993, -1.4735951100293454)
--(axis cs:-1.1616793927101807, -1.5031896944969747)
--(axis cs:-1.0728956393072933, -2.0965845420746185)
--(axis cs:-1.2706939218331745, -2.1261791265422474)
--(axis cs:-0.9148073291090942, -2.4773838148925664)
--(axis cs:-0.6772990742555306, -2.0373953731393604)
--(axis cs:-0.8750973567814119, -2.0669899576069892)
--cycle;
\path [draw=brown1613334, fill=brown1613334]
(axis cs:-2.768046701691958, 0.5561405409934618)
--(axis cs:-2.893742058049882, 0.7117062100533623)
--(axis cs:-3.3604390652295835, 0.33462014097959014)
--(axis cs:-3.4861344215875074, 0.4901858100394906)
--(axis cs:-3.6087227251704226, 0.005446593733791749)
--(axis cs:-3.109048352513735, 0.02348880285978916)
--(axis cs:-3.234743708871659, 0.1790544719196896)
--cycle;
\path [draw=goldenrod22215426, fill=goldenrod22215426]
(axis cs:1.4734972959393484, -0.9640167913782679)
--(axis cs:1.5030828661713518, -1.1618164224043324)
--(axis cs:2.096481759249545, -1.0730597117083225)
--(axis cs:2.1260673294815478, -1.2708593427343868)
--(axis cs:2.4772882361856716, -0.9149887557312839)
--(axis cs:2.037310618785538, -0.6774604496561938)
--(axis cs:2.0668961890175415, -0.8752600806822581)
--cycle;
\path [draw=gold2512280, fill=gold2512280]
(axis cs:-1.4735064062791874, 0.9640171136315486)
--(axis cs:-1.5030920045388927, 1.1618167404654112)
--(axis cs:-2.0964908850404806, 1.0730599456862953)
--(axis cs:-2.126076483300186, 1.270859572520158)
--(axis cs:-2.477297339578353, 0.9149889357499532)
--(axis cs:-2.0373196885210696, 0.67746069201857)
--(axis cs:-2.0669052867807753, 0.8752603188524326)
--cycle;
\path [draw=lightred, fill=lightred]
(axis cs:2.7680368179398753, -0.5561455384322258)
--(axis cs:2.8937321890453047, -0.7117111955763022)
--(axis cs:3.3604291604775343, -0.3346250822600133)
--(axis cs:3.4861245315829636, -0.49019073940408964)
--(axis cs:3.608712789212972, -0.005451511477115756)
--(axis cs:3.1090384182666746, -0.023493767971860535)
--(axis cs:3.2347337893721044, -0.17905942511593687)
--cycle;
\path [draw=darkgoldenrod16612945, fill=darkgoldenrod16612945]
(axis cs:0.9638809478066178, 1.4735842219868045)
--(axis cs:1.1616792360408823, 1.5031787683018754)
--(axis cs:1.0728955970956702, 2.0965736330046685)
--(axis cs:1.2706938853299345, 2.1261681793197393)
--(axis cs:0.9148073603483965, 2.477372936315662)
--(axis cs:0.6772990206271413, 2.0373845403745268)
--(axis cs:0.8750973088614057, 2.0669790866895976)
--cycle;
\path [draw=yellowgreen13918159, fill=yellowgreen13918159]
(axis cs:0.5560657108846174, 2.768290125176147)
--(axis cs:0.7116329873491787, 2.893983492118873)
--(axis cs:0.3345528865210012, 3.360685321512557)
--(axis cs:0.49012016298556255, 3.486378688455283)
--(axis cs:0.005382514403268757, 3.608973190970317)
--(axis cs:0.023418333591878393, 3.1092985876271055)
--(axis cs:0.17898561005643981, 3.234991954569831)
--cycle;

% add rectangles
\path [draw=darkcyan0149129, fill=none, very thick]
(axis cs:-0.14163641364829294, -2.8834285978796577)
--(axis cs:-0.68612182357958, -3.323355453586329)
--(axis cs:-1.1260486792862512, -2.778870043655042)
--(axis cs:-0.581563269354964, -2.338943187948371)
--cycle;
\path [draw=steelblue6999169, fill=none, very thick]
(axis cs:-0.6648427342085967, -1.7827488738651014)
--(axis cs:-1.357136723049181, -1.8863299195018033)
--(axis cs:-1.4607177686858832, -1.1940359306612187)
--(axis cs:-0.7684237798452987, -1.0904548850245168)
--cycle;
\path [draw=brown1613334, fill=none, very thick]
(axis cs:-2.883167427099379, 0.14171658104221913)
--(axis cs:-3.3231011743521126, 0.6861964227518709)
--(axis cs:-2.778621332642461, 1.126130170004605)
--(axis cs:-2.338687585389727, 0.5816503282949532)
--cycle;
\path [draw=goldenrod22215426, fill=none, very thick]
(axis cs:1.782664687444957, -0.6649925046896819)
--(axis cs:1.8862141832569683, -1.357291213280907)
--(axis cs:1.1939154746657432, -1.4608407090929183)
--(axis cs:1.0903659788537319, -0.7685420005016932)
--cycle;
\path [draw=gold2512280, fill=none, very thick]
(axis cs:-1.7826737554138152, 0.664992783134736)
--(axis cs:-1.886223349322784, 1.3572914770532551)
--(axis cs:-1.1939246554042648, 1.460841070962224)
--(axis cs:-1.090375061495296, 0.7685423770437048)
--cycle;
\path [draw=lightred, fill=none, very thick]
(axis cs:2.8831575040602218, -0.1417215675676285)
--(axis cs:3.3230913029292255, -0.6862013675718959)
--(axis cs:2.7786115029249583, -1.1261351664408996)
--(axis cs:2.3386777040559545, -0.5816553664366322)
--cycle;
\path [draw=darkgoldenrod16612945, fill=none, very thick]
(axis cs:0.6648426314624136, 1.782738043502929)
--(axis cs:1.3571366402823388, 1.8863189556056763)
--(axis cs:1.4607175523850864, 1.1940249467857509)
--(axis cs:0.7684235435651611, 1.0904440346830035)
--cycle;
\path [draw=yellowgreen13918159, fill=none, very thick]
(axis cs:0.14164322315414546, 2.8834161503107225)
--(axis cs:0.6861286907801102, 3.3233429346102628)
--(axis cs:1.1260554750796508, 2.778857466984298)
--(axis cs:0.5815700074536859, 2.3389306826847576)
--cycle;
\end{axis}

\end{tikzpicture}

%% file: img/mpc-ioc/scenario8a_mpc_ioc_planning30_safety_distance0.2_data_plot_of_paths.tex
% This file was created with tikzplotlib v0.10.1.
\begin{tikzpicture}[scale=0.53]

\definecolor{brown1613334}{RGB}{161,33,34}
\definecolor{darkcyan0149129}{RGB}{0,149,129}
\definecolor{darkgoldenrod16612945}{RGB}{166,129,45}
\definecolor{darkgray176}{RGB}{176,176,176}
\definecolor{lightred}{RGB}{220,100,90}
\definecolor{gold2512280}{RGB}{251,228,0}
\definecolor{goldenrod22215426}{RGB}{222,154,26}
\definecolor{steelblue6999169}{RGB}{69,99,169}
\definecolor{yellowgreen13918159}{RGB}{139,181,59}

\pgfdeclareplotmark{myStar}{
	\node[star, draw, fill, star points=5, star point ratio=2.5, minimum size=0.25cm, inner sep = 0.3mm] {};
}

\begin{axis}[
tick align=outside,
tick pos=left,
x grid style={darkgray176},
xmin=-6, xmax=6,
xlabel={$p_x$ in m},
xtick style={color=black},
xtick={-6,-5,-4,-3,-2,-1,0,1,2,3,4,5,6},
xticklabels={-6, , , , , ,0, , , , , ,6},
y grid style={darkgray176},
ylabel={$p_y$ in m},
ymin=-6, ymax=6,
ytick style={color=black},
ytick={-6,-5,-4,-3,-2,-1,0,1,2,3,4,5,6},
yticklabels={-6, , , , , ,0, , , , , ,6}
]
\addplot [draw=darkcyan0149129, fill=darkcyan0149129, mark=*, mark size=0.5pt, only marks]
table{%
x  y
-4 4
-4 4
-3.99505063538986 3.99505063538986
-3.98013421410621 3.98028258292369
-3.95001311521805 3.95105160837882
-3.89918991269523 3.90318896196513
-3.825210102134 3.83625814153382
-3.72685029659685 3.75164230563561
-3.60273692920691 3.65118697767187
-3.45256549331041 3.53809443467605
-3.28067629548829 3.41852264264131
-3.09287385576742 3.29679212065544
-2.89407896014333 3.17832545708409
-2.68986881928642 3.06815598514374
-2.48642378846554 2.97054398204309
-2.29132536926902 2.886912636191
-2.11157112471229 2.81974770174105
-1.9480242256914 2.76822150141381
-1.79805527921942 2.73027277553149
-1.65589757456148 2.70212547572463
-1.51531937348518 2.68115197498445
-1.36978727260661 2.66570698350115
-1.21307081468324 2.65497685369064
-1.05082301183605 2.64913484867194
-0.884736594340887 2.64770306811173
-0.721916152114413 2.64994205204043
-0.569410088760657 2.65468976927326
-0.434213891013211 2.66057345834717
-0.323242535360687 2.66622278901729
-0.241932789370361 2.67055555296809
-0.189230008666881 2.67322509361048
-0.163805682828713 2.6743185614489
-0.164468958499894 2.67429842089802
-0.189109963369706 2.67398471915284
-0.233145603805467 2.67442048858304
-0.289592208238455 2.67653939675453
-0.351375230323309 2.68088033148452
-0.411410153913156 2.6873754917755
-0.463843519532413 2.69532116865841
-0.501766646338486 2.70293011088934
-0.518554709757542 2.70722723863865
-0.508057507414478 2.70388642707407
-0.469838731890011 2.68903980339602
-0.405239760089984 2.659575577777
-0.321877273769519 2.61613631137604
-0.226847012305373 2.56073198659592
-0.12667872664479 2.49648393210113
-0.02735190680125 2.42737664777111
0.0656632999179806 2.35805728570356
0.146666687612735 2.2928180793512
0.21055419894054 2.23671917520107
0.263177298066571 2.18607739366263
0.310076987457305 2.13678372649279
0.356201751480885 2.08406783237658
0.405936944106209 2.02262370716082
0.463184582604053 1.94675079963273
0.531401557527264 1.85045478766546
0.613717272677447 1.72760621539817
0.709403748649715 1.57776658255282
0.818038382263553 1.40067648045665
0.939412772953817 1.19631937289449
1.07372731828168 0.964817519778267
1.22138772338993 0.706383750200753
1.38109064898347 0.424392326739668
1.55001868950399 0.124815850975306
1.7251288427541 -0.186106964825065
1.90327074409945 -0.502178980577748
2.08120458187856 -0.817263638067319
2.25553880068984 -1.12527330271981
2.42289716838124 -1.42023532314463
2.57997446375833 -1.69695665167974
2.72396018640099 -1.95014869154164
2.854189956271 -2.17851278671615
2.97083900262205 -2.38179778906279
3.07498050322259 -2.56126621421297
3.16757078122614 -2.71849205633698
3.24926842463566 -2.8552443534102
3.32132035937981 -2.97381459183194
3.38549870318465 -3.07771534958426
3.44374052226956 -3.17023064385391
3.49704586175397 -3.25323995015883
3.5460034577865 -3.32797237133191
3.59126991187919 -3.39577794897289
3.63293737378841 -3.45707852448874
3.67122707932741 -3.51225131736331
3.70641536668493 -3.56173141494011
3.73874811840447 -3.60598253154124
3.76848572297522 -3.64560380766516
3.79583069064469 -3.6810822525528
3.82075051310875 -3.71270765060712
3.84331396143 -3.74082433134671
3.86365280273633 -3.76579368252614
3.88221433420069 -3.78827270859303
3.89916437625504 -3.80856020540511
3.91465681931988 -3.82691286949995
3.92879151279877 -3.84350336035212
3.94165671105756 -3.85847560904467
3.95336867527249 -3.87199627763035
3.9640143034957 -3.88419035457726
3.97364366582484 -3.89513568165535
3.98236599119399 -3.90497451217481
3.99024892870869 -3.91379947723061
3.99734376778623 -3.92168256734108
4.00373787008952 -3.92873394809724
4.00947141703672 -3.93501021424945
4.01458194395158 -3.94056454985746
4.01920184627572 -3.94554894445854
4.02328661274177 -3.94992368292848
4.02681343048537 -3.95367358506442
4.02983490628824 -3.95686361753505
4.03242661055283 -3.95958144031431
4.03465375786459 -3.96190199953373
4.03656288528645 -3.96387925422084
4.03818378963188 -3.96554866488353
4.0396019681358 -3.9670018200999
4.0408211688562 -3.96824525022156
4.04182782339602 -3.96926756679094
4.04263399316903 -3.97008317937689
};
\addplot [draw=steelblue6999169, fill=steelblue6999169, mark=*, mark size=0.5pt, only marks]
table{%
x  y
0 4
0 4
-6.43586035573391e-13 3.99299999990604
0.000104977987195512 3.9720021290247
0.000672797031788106 3.93414265535512
0.00210426263901023 3.88643244765076
0.00481117768895697 3.83232871492768
0.00987518660670135 3.76492095843489
0.0176336169684737 3.69128682500988
0.0298378954382566 3.60466362937367
0.0482371598720926 3.50352810599938
0.0761442446992241 3.38156791107793
0.11657811954598 3.23254042338259
0.171060817893476 3.05637018329188
0.240479067566436 2.85365433617956
0.325002114753023 2.62559207519186
0.422203356104239 2.37859402062696
0.531329604837234 2.11305154415211
0.64882266793293 1.83509412501373
0.770737832712127 1.55065552908231
0.892956730840718 1.26551741976729
1.01133746537833 0.985484958385449
1.1219260850305 0.716424303983532
1.2211898743592 0.464235158259598
1.30656293312414 0.233940519433914
1.37688554845345 0.0286538013173183
1.43441516922336 -0.157284982978975
1.48111947451584 -0.329297690618955
1.51855463273549 -0.492706960487553
1.54771957384863 -0.652711324747837
1.5689139242244 -0.814332949920927
1.58155877407286 -0.981522871052739
1.58425690260775 -1.15393341330218
1.57566505183918 -1.32855264346706
1.55530847568276 -1.49909967573217
1.52341286091202 -1.67061636636106
1.48132362243589 -1.83972987250852
1.43208747520753 -1.99946536616107
1.37860579077462 -2.1464032143233
1.32171162079187 -2.28319793402698
1.25994947764104 -2.41637889226882
1.19609574034543 -2.53924127667704
1.13391948988749 -2.64551312041774
1.07790533842741 -2.73023627284633
1.02637431968601 -2.80032480114989
0.976518941104764 -2.86212899456612
0.924955265424003 -2.92107980575216
0.867564827436742 -2.98224914219846
0.799522317970466 -3.05056109570448
0.724933984364577 -3.12180279968904
0.647795919934269 -3.19258666252436
0.572647378982192 -3.25949614248894
0.503331591795083 -3.31996969130953
0.441151791453123 -3.37363385363535
0.385895763612913 -3.42128103978515
0.336910777023181 -3.46389800026289
0.293410667315229 -3.50242252727244
0.254556944897181 -3.53779238943853
0.220129758432379 -3.57028450503743
0.189409267293585 -3.60061683912809
0.161806638880227 -3.62938569625729
0.137143600697507 -3.65671486593391
0.11472328053108 -3.68340079387506
0.0946450114451899 -3.7093448851927
0.0768314508153533 -3.73461462284224
0.0610737366862636 -3.759405043903
0.0473318254734529 -3.78353798793132
0.0356879978968878 -3.80658214504561
0.0260573795488475 -3.82825288089289
0.0184008808004224 -3.84795967857222
0.0124718182407634 -3.86557212437687
0.00798171452397125 -3.88112591575781
0.00466122517001047 -3.89468812527798
0.00225510763585908 -3.90643528816378
0.000474980880698712 -3.91698277385567
-0.000778091442151341 -3.92633531777173
-0.00163763596101569 -3.93466046699155
-0.00220679015402687 -3.94209565386629
-0.00254753019985422 -3.94851189653184
-0.00273952926449636 -3.95426192940253
-0.0028224503546714 -3.95920040982144
-0.00283747581064374 -3.96362585934912
-0.00281304988249084 -3.96679982087189
-0.00278130055560651 -3.96864727534231
-0.00274002490941588 -3.97034263636993
-0.00267754822221512 -3.97247870377902
-0.00259764539465068 -3.97495795532082
-0.00250886187849027 -3.9775892128793
-0.00241916811843203 -3.98021739424831
-0.00234047664391742 -3.98259191030041
-0.00224413095994357 -3.9856789107594
-0.00219101767416646 -3.98751309202381
-0.00214268913495824 -3.98932768444608
-0.00211297975191682 -3.99054717663071
};
\addplot [draw=brown1613334, fill=brown1613334, mark=*, mark size=0.5pt, only marks]
table{%
x  y
4 4
4 4
3.99505063538986 3.99505063538986
3.98039488389662 3.98024767344021
3.9521719952083 3.95116547957166
3.90608770433228 3.90223062015554
3.84086483345737 3.83013803626454
3.75789973087718 3.73369509537553
3.65899114233327 3.61148844687786
3.54637374088752 3.46193227808604
3.42275316272125 3.28332587835744
3.29134479093359 3.0739298775615
3.15590776872559 2.83205291661706
3.01864558854431 2.55792019809446
2.88049513358545 2.25100581865348
2.74228907608935 1.91148355119531
2.60725206182456 1.53860908421735
2.47984288722069 1.13884879542922
2.36172658273819 0.719596052778513
2.25359490816334 0.28837236247698
2.15536918882501 -0.147267906418214
2.06641111902218 -0.579794703168601
1.98573673893556 -1.00173566779105
1.91223102707671 -1.4056952734627
1.84486359249524 -1.78435262559984
1.78237362367522 -2.13341978711008
1.7232292551924 -2.45260944085925
1.66634576821968 -2.74169058221183
1.61108743921787 -3.00048277118962
1.55726190782548 -3.22888718612056
1.50513792800955 -3.42681090945651
1.45542580440747 -3.59426844558399
1.40923523385155 -3.73151530064014
1.36680795543346 -3.84231025595542
1.32783035378966 -3.93155639598086
1.29165590633406 -4.00405211336369
1.2569042507502 -4.06490265104265
1.22076799080637 -4.12006357359763
1.17941782752183 -4.17490963106113
1.12806514078544 -4.23380784890929
1.0609434146863 -4.29990191337845
0.971732419256261 -4.37453568959865
0.854970256060648 -4.45626565730672
0.707380450827991 -4.54095454622084
0.527026378075453 -4.62499292534118
0.318949588302738 -4.70236568736893
0.088942023286567 -4.76872823987398
-0.158291939631056 -4.82161386121189
-0.417236237299826 -4.85955531277727
-0.681184910636106 -4.88208077452134
-0.944603703186698 -4.88990975882224
-1.20269341366286 -4.88452716284921
-1.45258081198598 -4.86778097541078
-1.69129128317662 -4.84151661170795
-1.91703030537669 -4.80788696557183
-2.12693141579578 -4.76873657401623
-2.32148259989822 -4.72597429946081
-2.50145668999904 -4.68124171473569
-2.6684748472479 -4.63572426074305
-2.82328810665266 -4.59058455617243
-2.96669913839518 -4.54675373866349
-3.09945753638262 -4.50501859735454
-3.22187238717301 -4.46609101133625
-3.33468494898556 -4.43037865665666
-3.43875828162274 -4.39812807714947
-3.53488005775063 -4.36947495876658
-3.62379268004394 -4.34448174383027
-3.70591029942204 -4.32320626573724
-3.78134957850082 -4.30568275101016
-3.85067766238808 -4.29172955029401
-3.91348329618895 -4.28130024754093
-3.96969766816546 -4.2741633833082
-4.01912452649756 -4.27000115486947
-4.06150312857952 -4.26842280055869
-4.09626565110289 -4.26891854965712
-4.12310356610082 -4.27081511682456
-4.14309787073537 -4.27345433014303
-4.15679177405918 -4.27612651792906
-4.16465575152409 -4.27821130735414
-4.16689912179493 -4.27896407683741
-4.16518788513965 -4.27826571909913
-4.1607795948035 -4.27614260058823
-4.15594790284779 -4.2734602501295
-4.15170749705477 -4.27079261057121
-4.1464576203969 -4.26710273745798
-4.13948338662295 -4.26169275183059
-4.13001459805305 -4.25363563877888
-4.11918726338689 -4.24363199968452
-4.10675607528257 -4.23127033492816
-4.09324135170718 -4.21692411695554
-4.07929371594135 -4.20125204346305
-4.06533708363917 -4.18477911856465
-4.05196220570539 -4.16831061813212
-4.03971145769636 -4.15265619983274
-4.02811351048055 -4.13738699052784
-4.01703955825742 -4.12250563114945
-4.00568784955507 -4.10706365434949
-3.99519101299862 -4.09272365184269
-3.98668178333463 -4.08113293596616
-3.98017376886977 -4.07236519508197
-3.97558740493119 -4.06628881148702
-3.972467868737 -4.06224091985863
-3.96984640503564 -4.05891591834989
-3.96738043892063 -4.05586850048067
-3.96507475000652 -4.05309091949704
-3.96301612557525 -4.05067670493177
-3.96120079624456 -4.04859572148539
-3.95959187367786 -4.04679237854907
-3.95816652335405 -4.04522670011494
-3.95694679729155 -4.04391150155692
-3.95593966457513 -4.04284329744939
-3.95513543415569 -4.0420025152926
};
\addplot [draw=goldenrod22215426, fill=goldenrod22215426, mark=*, mark size=0.5pt, only marks]
table{%
x  y
-4 0
-4 0
-3.99299999990604 -1.60683879907691e-13
-3.98599739947668 3.50129066615392e-05
-3.97199436449339 0.00014957133542229
-3.94400566093468 0.000607479664930101
-3.9027961258241 0.00182500525889212
-3.84147083963864 0.00475314321812134
-3.76707175843735 0.0100358530045755
-3.67415407988752 0.0192705729544038
-3.55603766878192 0.0349814169708317
-3.40967796477166 0.0601660905474878
-3.2352251726065 0.0980016840304019
-3.04057913917895 0.150145025290449
-2.81981339894016 0.219625636108827
-2.57713353739372 0.306312620026219
-2.31954183319065 0.408079928244419
-2.04208584954373 0.526835078958769
-1.75119550442474 0.659396402588745
-1.45160052667812 0.802554343424475
-1.14909421512102 0.952028535029536
-0.849247293461227 1.1032506474467
-0.557444329093792 1.25158136610381
-0.277157156249936 1.39341750716071
-0.00338332466635427 1.52962277072152
0.258182986347758 1.65592443467456
0.501574593227612 1.76844596684831
0.720492605021245 1.8639162346097
0.914229320816495 1.94229765258637
1.08231287594434 2.0041613984071
1.22423190037763 2.05054739961838
1.33913639178265 2.08286116652848
1.42986260744937 2.10385472336994
1.50199443903274 2.11665670365064
1.56157627598535 2.12377553648906
1.61496666038284 2.12683196626077
1.66859985116608 2.12632031179794
1.72870949607342 2.12142660006535
1.80098526016993 2.10992155574904
1.8893507904534 2.08881213548573
1.99717659929903 2.05469234755619
2.12090780545945 2.0061478810345
2.25686147237879 1.94258425112421
2.39981529720244 1.86514968436055
2.54717826517414 1.77463746915934
2.69349541101488 1.67408614016212
2.83414939442206 1.56708616406172
2.96644683485437 1.45693898114266
3.08892037898474 1.34628370984214
3.19944954766638 1.23878106726304
3.2992039646882 1.1352974596755
3.39033627170567 1.0353679000101
3.47423314141587 0.939036463145736
3.55126866531268 0.847258856352175
3.62207271729249 0.760568149831029
3.68736653748837 0.679209351815425
3.74780542737987 0.603323089902173
3.80411624114557 0.532798593851967
3.85679012112744 0.467655625517125
3.90620089775686 0.407892269962705
3.95264809511266 0.353496648736863
3.99623080597356 0.304530875500139
4.0372704975081 0.260703137168644
4.07546527674692 0.222291337861476
4.11063812531683 0.189291927186065
4.14263385290641 0.16155345708023
4.17149891580804 0.138652517225593
4.19760375355765 0.119901053670183
4.22082543043221 0.104962586894988
4.24031649190701 0.0938655739093801
4.25402644813262 0.0870932327980961
4.26126338819858 0.0840638948879686
4.26397223574784 0.0831144159911394
4.26308231057489 0.0833708958332638
4.25909670435269 0.0842917124300712
4.25500769361871 0.0850242189568274
4.2503412872975 0.0856374171596171
4.24546094694742 0.0860671856976446
4.23972567829105 0.0863480089354483
4.23137751514594 0.0864619588595244
4.22013256235666 0.0862627844665705
4.20446065157976 0.0855416028095195
4.18636630553646 0.0842575570931031
4.16634962667596 0.0824180658447191
4.14562257008768 0.0801440001920648
4.12359017999537 0.0774068431272082
4.09970332155541 0.0741841423041403
4.07606132926611 0.0708300920163744
4.0548492105529 0.0677117352485099
4.03604738625614 0.0649110776013572
4.01920008321135 0.0624335491296404
4.00612453988586 0.0605607846418669
3.99744225705783 0.0593686501223995
3.99246387688631 0.0587260556639128
3.98928397808631 0.0583513590877125
3.98668145753427 0.0580796532277276
3.98439012016689 0.0578756554670826
3.98250569316369 0.0577369464453609
3.98098431344532 0.05764661390638
};
\addplot [draw=gold2512280, fill=gold2512280, mark=*, mark size=0.5pt, only marks]
table{%
x  y
4 0
4 0
3.99299999990604 -1.02471841117147e-13
3.9859444068464 -3.52698475969211e-05
3.97183697506185 -0.000151783743218222
3.94389179111575 -0.000613324499733479
3.89512888151205 -0.0020654080992629
3.8186241345296 -0.00574184118654287
3.71232603308451 -0.0133297849476324
3.58317934786546 -0.0262226490379824
3.43836926340391 -0.0455581462056396
3.28519620145196 -0.0720047430337676
3.11699012401659 -0.106839266420792
2.92715824011775 -0.151776665109929
2.70940323395735 -0.208697994888583
2.46386378745585 -0.277689514270475
2.19221597597673 -0.357912745450165
1.89254617459046 -0.449104120295064
1.56984458309918 -0.54845271997297
1.23031808000466 -0.652333080224351
0.880108953995962 -0.756906107756368
0.525400655633006 -0.858303262161209
0.172467918881511 -0.952836361236203
-0.172205571213374 -1.03717728899596
-0.501894706710734 -1.10857319703858
-0.809600296626566 -1.16506424982475
-1.08964872868543 -1.20592708592423
-1.34149651877385 -1.23202093535201
-1.56437467208189 -1.2446586277522
-1.7603790769399 -1.24566422155393
-1.93392028143331 -1.23675103347088
-2.08897745853857 -1.21919402882846
-2.23067755323365 -1.19353471709973
-2.36421529510542 -1.16044190218297
-2.49381126325539 -1.11997019927568
-2.62114190918017 -1.07232152110264
-2.74587293640623 -1.0182572185463
-2.86652784422326 -0.959064894751574
-2.98158349893059 -0.89634910329954
-3.09098329133213 -0.831144491057573
-3.19228664158758 -0.766009813505977
-3.285546677397 -0.702113346936164
-3.3707582923927 -0.640601405499577
-3.44850464269158 -0.582069921769848
-3.51970836286658 -0.526713577502672
-3.58508052095605 -0.474737886269967
-3.64575641819084 -0.42589720448028
-3.70186866372867 -0.380512605290992
-3.75399436616872 -0.338511643269287
-3.802576295458 -0.299820406198738
-3.84826801787691 -0.264217384291372
-3.89104585916419 -0.231919512119181
-3.93091297274053 -0.203031423265979
-3.96786554912863 -0.177585634318106
-4.00177776987455 -0.155617015109547
-4.03245716378481 -0.137128014467381
-4.06063250007898 -0.121534540148077
-4.08571636043348 -0.108977046301496
-4.10743222641077 -0.0993115438069579
-4.12516672696327 -0.0924291169431618
-4.13862851607442 -0.0879441607474085
-4.14767450832734 -0.0854594554937503
-4.15369745215959 -0.0841270241137059
-4.15744295586741 -0.0834816969202794
-4.16144800530998 -0.0829709348491421
-4.16514779232651 -0.0826508066303549
-4.16727152028275 -0.082548735831047
-4.16688698426997 -0.0825535806200326
-4.160607549918 -0.0824326228348407
-4.15213640701046 -0.0820327717207639
-4.14179367889443 -0.0813012762320576
-4.12686694811433 -0.0799327947390201
-4.11058618272759 -0.0781713359263875
-4.09444709303833 -0.0762153463324217
-4.07795464454345 -0.0740596589530328
-4.06129338061517 -0.0717673434484552
-4.04435143294538 -0.0693699736476152
-4.02846883666049 -0.0671196495602519
-4.01483470672012 -0.0652048890025899
-4.00362587259179 -0.0636566901562063
-3.99567360649916 -0.0625897751493983
-3.99072511102595 -0.0619555922228411
-3.98784044340345 -0.0616097839995008
-3.98561893969675 -0.0613678893875703
-3.98365083373302 -0.0611783877024794
-3.98200743206751 -0.0610421762191453
-3.98060443866692 -0.0609454836044534
};
\addplot [draw=lightred, fill=lightred, mark=*, mark size=0.5pt, only marks]
table{%
x  y
-4 -4
-4 -4
-3.99505063538986 -3.99505063538986
-3.98038385481785 -3.98023654171016
-3.95139092429317 -3.95036093888996
-3.90380070475112 -3.89982728124177
-3.83710364477754 -3.82610581407442
-3.75270534401379 -3.72799785580046
-3.65241397616133 -3.60408435562196
-3.5392461149417 -3.45379931435933
-3.41724851279675 -3.27754089464025
-3.28911706783702 -3.07337076782036
-3.15669122559838 -2.83687750269363
-3.0231308473636 -2.56708266015098
-2.89213412449133 -2.26306049027001
-2.76549683327796 -1.92495592717798
-2.64585589964039 -1.55266587108212
-2.53601644459534 -1.1534594066841
-2.4375326623897 -0.734846739563112
-2.35090229214507 -0.304437131508846
-2.27577254573342 0.130161449283094
-2.21115396538689 0.561396310525591
-2.15563631962877 0.981802251420252
-2.10760539128253 1.38400709199659
-2.06545846738678 1.76072940894226
-2.02749416073179 2.10773169547669
-1.99198095299298 2.42482505893324
-1.95763488842401 2.71186843601579
-1.92362295279058 2.96873641333537
-1.88956358754578 3.19532153024933
-1.85552894168216 3.39152951741109
-1.82204116456577 3.55731979363822
-1.79004904740651 3.69281900694444
-1.75963148294729 3.80297951503236
-1.7306471597238 3.89285795054736
-1.70255718365126 3.96759944832587
-1.67470406613123 4.03132882629648
-1.646203465637 4.08749570384783
-1.61439236235362 4.14152812723523
-1.57559725638181 4.19828050830348
-1.52502487852753 4.26183337557748
-1.45704407715418 4.33485943870551
-1.36520952616155 4.41849229351306
-1.24366641321736 4.51112537601467
-1.08949576021763 4.60751912695892
-0.902356589699409 4.70257791287578
-0.680155037523335 4.79293273483892
-0.428557632631481 4.87281358467083
-0.154972233581877 4.93786990335428
0.132769349217848 4.98559208947426
0.426599530359263 5.01514021317382
0.718674893508234 5.0271264184085
1.00313515058624 5.02331252831974
1.27474426954713 5.00618863336413
1.5303339636836 4.9785240468089
1.77079776675416 4.94264399059669
1.9940844002731 4.90058768786579
2.201811180837 4.85428312643765
2.39480005720366 4.80526060114947
2.5743994695883 4.75493009063288
2.74161244253749 4.70452618146783
2.89681851933318 4.6551562774969
3.04043238409364 4.60779436980799
3.17305397700006 4.56317646924088
3.29545957419425 4.52182775696964
3.40825416788605 4.48415133354801
3.51257980429859 4.45025600520157
3.60896638665561 4.42030748455198
3.69792777665324 4.39439530293944
3.7796352512393 4.37257876267676
3.85381033540706 4.35493779109939
3.92153276184591 4.34112436898157
3.98277003687683 4.3309702570622
4.03731058381003 4.32422295223774
4.08487395159983 4.3205129007835
4.12491233821576 4.31939346485801
4.15858686283719 4.32027596797977
4.18416684697072 4.32245429847041
4.20118627005216 4.32499510681447
4.21054462561862 4.32700676582789
4.21306898144324 4.32771732450521
4.21002736364437 4.3266346500983
4.20266280192924 4.32345520759325
4.19224825880412 4.31814918802472
4.18083995273086 4.31144122380692
4.17058898143594 4.30461289516568
4.1594854351745 4.29631858673248
4.14663678859609 4.28568439446259
4.13260101506885 4.27292913308247
4.1184303664721 4.25892384436727
4.10465502485465 4.24424659081142
4.09043784562567 4.228051759577
4.07644193081398 4.21113774172696
4.06240446140355 4.19328148530218
4.04857253419717 4.17492018629188
4.035309344755 4.15670553637811
4.02323040485903 4.13969891728296
4.01164021374287 4.12310940199188
4.00056744022104 4.10712174443002
3.9908729955756 4.09309941513175
3.98311447259981 4.08192622866621
3.97749851513223 4.07391373677668
3.97385112547555 4.0687840166213
3.97174151583381 4.0658753070624
3.97013970578041 4.06371563672304
3.96852451237418 4.06159242949484
3.96688935598377 4.05949336266274
3.96518000357901 4.05735265035708
3.96269202836119 4.05430681144948
3.96030666700136 4.05145453764766
3.9581796179822 4.04896599199271
3.95628195237612 4.04679241747896
3.95461998672592 4.04492666439497
3.95377454818182 4.04399613811789
};
\addplot [draw=darkgoldenrod16612945, fill=darkgoldenrod16612945, mark=*, mark size=0.5pt, only marks]
table{%
x  y
0 -4
0 -4
-7.3052633138243e-13 -3.99299999990604
-0.000104889758315492 -3.97200776738102
-0.000732710932705323 -3.93013428379862
-0.00282307517687287 -3.86044114216658
-0.00780095495425121 -3.76090624587469
-0.0175071669651539 -3.63164881169682
-0.0342243885580823 -3.4729187367894
-0.0596890483983324 -3.2920990385539
-0.0952237641065639 -3.09669079184698
-0.141518499472559 -2.89428360054831
-0.200045732758659 -2.67846976430352
-0.272854933231161 -2.44292586422294
-0.359989470036326 -2.1883441939796
-0.459009382506438 -1.92102242561351
-0.567221761486127 -1.64588665511114
-0.681794407676639 -1.36692469670353
-0.798859920170129 -1.08982144898658
-0.915101331716478 -0.820795573496976
-1.02653350421181 -0.565748815589488
-1.13432024488977 -0.318429687184043
-1.24003086475645 -0.0719105915141691
-1.34494840063491 0.180292112565463
-1.44508688473846 0.432205257901649
-1.53678091077788 0.677672226131453
-1.61688559930504 0.910286561386112
-1.68296597935964 1.12330185873741
-1.73355268146762 1.30973323932439
-1.76986640268662 1.46902103226509
-1.79353270156163 1.60030436407654
-1.80742629382084 1.70859194074053
-1.81396810797519 1.79944405599214
-1.81563883691529 1.87941210281651
-1.81374007854727 1.954852247047
-1.80897085193692 2.02677016737833
-1.80250360949694 2.08977717745374
-1.79591502037481 2.13888994819898
-1.79124055342795 2.16795787044762
-1.79066664298011 2.17110112386672
-1.79511666401181 2.14877872151286
-1.80375792214956 2.10794005178509
-1.81523415858365 2.05545522817079
-1.82795920957533 1.99813911609882
-1.84068095351749 1.94288300743708
-1.8518326012751 1.89651328535312
-1.85954807290336 1.8657816516939
-1.86494337061622 1.84473594607549
-1.86616179052846 1.84012124286919
-1.86109348039751 1.8584118767659
-1.84731271708275 1.90501902076969
-1.82401141089025 1.97785541541127
-1.79013116128538 2.07464429046377
-1.74449886859553 2.19271583925458
-1.68745545369432 2.32545798476777
-1.619728784913 2.46647447697273
-1.54251150417395 2.60968191662692
-1.45762209064074 2.74966964576171
-1.36823283519619 2.88199661366898
-1.27671107301454 3.0047790752903
-1.18516151952834 3.11738872952392
-1.09474430355895 3.22047015449001
-1.0052261436905 3.31603683440895
-0.917555962255035 3.40450706582448
-0.832010467687813 3.48689786135107
-0.750230451767921 3.56281333339899
-0.6734004722003 3.63221367790084
-0.601764206123518 3.69579775482915
-0.535216394716592 3.75440820771135
-0.473495637867318 3.8088773345665
-0.416681033779006 3.8595938366204
-0.364612269233194 3.90706528484899
-0.317299924350147 3.95155651239733
-0.27484887834722 3.99309884092518
-0.237115465616951 4.03183511138195
-0.203791350822485 4.0680467764805
-0.175444498159321 4.10097195894894
-0.151329312099588 4.13120927320317
-0.131227433093742 4.15870426216534
-0.115091047512534 4.18306006829741
-0.102539116013273 4.20424239171864
-0.0932007942376924 4.22202994710923
-0.087040271150844 4.23547467219724
-0.084083891709616 4.24307584991575
-0.0834118735725695 4.24517387560153
-0.0837830361923682 4.24370490127212
-0.0845753287593634 4.23956616640295
-0.0854579925517927 4.23305266126962
-0.0860422824067157 4.22614537801022
-0.0863791707596015 4.2172630506822
-0.0863331369038463 4.20689649641615
-0.0858370681632956 4.19510301948433
-0.0848589647611373 4.18212062438035
-0.0833228359886986 4.16740144872045
-0.0810958457464323 4.15010771830268
-0.0784211509744548 4.13211513217258
-0.0750345907015995 4.11150935076707
-0.0712354800880423 4.0900778906483
-0.0673332016689682 4.06910532418222
-0.0634223251125043 4.04861474132316
-0.0595239162641632 4.02836174147794
-0.0563819229454147 4.01183234197399
-0.0537122555948538 3.99726613280997
-0.0517824024656015 3.98623605354029
-0.0506332426374382 3.97924370509531
-0.0499645783658637 3.97478816635218
-0.0495448632808251 3.97167594678096
-0.0492472212778354 3.96920540595331
-0.0490315422172775 3.96719048968632
-0.0488858768880706 3.9656436402758
-0.0488002998452029 3.9646029463058
};
\addplot [draw=yellowgreen13918159, fill=yellowgreen13918159, mark=*, mark size=0.5pt, only marks]
table{%
x  y
4 -4
4 -4
3.99505063538986 -3.99505063538986
3.98013059953624 -3.98027903831048
3.95016919435965 -3.95120306687095
3.8996244179926 -3.90360268287301
3.82607017884055 -3.8370561474216
3.73350425719818 -3.75742300534324
3.61944684375507 -3.66510383118376
3.4880813052318 -3.56617354427016
3.34379370123537 -3.46628801717491
3.19112223031494 -3.37017272617875
3.03348588292465 -3.28077583747813
2.86537813129057 -3.19434454773783
2.68092133926757 -3.10775124135449
2.47425064452668 -3.01843567301109
2.24405214702977 -2.92597476795265
1.99380957996102 -2.83436022916342
1.72754457418105 -2.7476091124405
1.45242516155868 -2.66731209837039
1.17573996538833 -2.59430399806183
0.904815209163453 -2.52886540267068
0.646964979596802 -2.47093153231871
0.409440610378777 -2.42029988845869
0.199443365531454 -2.37684932512138
0.0187684832851382 -2.33965236761734
-0.132421104501857 -2.30789342374091
-0.255600652470607 -2.28085750334447
-0.350994155870668 -2.25851518766625
-0.418915211648602 -2.24123922887126
-0.460198322657515 -2.22967686289706
-0.476998095160169 -2.22444169149562
-0.471903367655625 -2.22622148325897
-0.448753041095011 -2.23533644216583
-0.41462231179283 -2.25054105555632
-0.376365168716425 -2.26988242685877
-0.340497008027412 -2.29051824000063
-0.31298303753256 -2.30854838399118
-0.299015664846128 -2.31894634774878
-0.302907407368964 -2.31570297282061
-0.32829115657774 -2.29230948980813
-0.373696279904154 -2.24653410878835
-0.437458488908454 -2.17691219712334
-0.514569514401026 -2.08473902228779
-0.598312897426822 -1.97384999306148
-0.682438347261923 -1.84874601658843
-0.761370875829836 -1.71469576436039
-0.830464781698732 -1.57777311226588
-0.886278104891461 -1.44484010005966
-0.926862624297615 -1.32346357063141
-0.953250543519632 -1.21695106636597
-0.969720295557342 -1.11985707019106
-0.978957139774975 -1.02609527656872
-0.982190432121281 -0.929221614165821
-0.979368480765656 -0.822637428826347
-0.969393355083663 -0.700220445140757
-0.952111153334591 -0.569194194491906
-0.928507977570297 -0.436456491062057
-0.900514726043898 -0.308768035880767
-0.870841236188007 -0.192873690647737
-0.842721221998112 -0.0952644550831403
-0.81968927921999 -0.0221905757963506
-0.80472660281822 0.022190946744527
-0.798032887006884 0.0411131963145681
-0.798343471557625 0.040261950935212
-0.804015001295452 0.0249481302851848
-0.813168120520779 0.000221493545980427
-0.823361411886928 -0.0277575413669639
-0.83227929491669 -0.053030867615633
-0.837950758350206 -0.0699147323105189
-0.838614658929407 -0.0720315151954622
-0.833028483307497 -0.0525148654939885
-0.821136907111849 -0.00567640819571139
-0.804846144374435 0.0694652624003599
-0.787020948363637 0.171244347942161
-0.770046578577153 0.302861318824043
-0.757775421498128 0.459529066878839
-0.75368270674971 0.636658585865242
-0.760865749491944 0.830990079573317
-0.782146179484861 1.03893870858672
-0.820131587517121 1.25760423217377
-0.877330500443904 1.48444099549347
-0.955134767336366 1.71384451708986
-1.05364991082363 1.93966307032278
-1.17173235884941 2.15771055569752
-1.30772089801917 2.36632674549541
-1.45753994778682 2.56168692811408
-1.6169400393961 2.74124094807241
-1.78137893755702 2.90258790182214
-1.94664789503828 3.04516967598582
-2.10904522535076 3.16928161250559
-2.2658955097827 3.27613053035107
-2.41479885278463 3.36705578635228
-2.55483763350067 3.44411223322403
-2.68500254808345 3.50902891355052
-2.80573074316697 3.56396723731065
-2.91761883974885 3.61077323511175
-3.02148063477522 3.65106482767508
-3.118176730582 3.68622358275516
-3.20696897214238 3.71682002352931
-3.28808385417114 3.74365125190914
-3.36114519369112 3.76716296410967
-3.42690933381039 3.78801914829523
-3.4858600401072 3.80671163372799
-3.54187887341169 3.8247502764939
-3.59396780420811 3.84199234560465
-3.64203831945977 3.8585935691904
-3.68557238561655 3.87443601547588
-3.72462033549647 3.88950066151921
-3.759505684709 3.90385882670229
-3.79063452409501 3.91762481855611
-3.81837870070538 3.93083869059311
-3.84308708229817 3.94349399327269
-3.86506396698934 3.95556488714254
-3.88452188348956 3.96697814391107
-3.90161037533784 3.97763699447007
-3.91651672288376 3.98747960531386
-3.92945359144042 3.99648433975022
-3.94073946753511 4.00470538822133
-3.95090708022649 4.01246466889989
-3.96019783401418 4.01982101934658
-3.9685122771715 4.02659887821643
-3.97533139052842 4.03231044748599
-3.98064941523773 4.03687805815411
-3.98467626052242 4.04041748166722
-3.98769973784656 4.04313126621792
-3.9902290546558 4.0454368439053
-3.99222536157213 4.04728286231155
-3.99375202586948 4.04871365153416
-3.99487417695338 4.04977851295134
-3.99564248544039 4.05051600532615
};

% add circles
\draw[draw=darkcyan0149129,fill=none, very thick] (axis cs:4.04263399316903,-3.97008317937689) circle (70.0);
\draw[draw=steelblue6999169,fill=none, very thick] (axis cs:-0.00211297975191682,-3.99054717663071) circle (70.0);
\draw[draw=brown1613334,fill=none, very thick] (axis cs:-3.95513543415569,-4.0420025152926) circle (70.0);
\draw[draw=goldenrod22215426,fill=none, very thick] (axis cs:3.98098431344532,0.05764661390638) circle (70.0);
\draw[draw=gold2512280,fill=none, very thick] (axis cs:-3.98060443866692,-0.0609454836044534) circle (70.0);
\draw[draw=lightred,fill=none, very thick] (axis cs:3.95377454818182,4.04399613811789) circle (70.0);
\draw[draw=darkgoldenrod16612945,fill=none, very thick] (axis cs:-0.0488002998452029,3.9646029463058) circle (70.0);
\draw[draw=yellowgreen13918159,fill=none, very thick] (axis cs:-3.99564248544039,4.05051600532615) circle (70.0);

% add solid stars as goal points
\addplot [draw=darkcyan0149129, fill=darkcyan0149129, mark=myStar]
table{%
	x  y
	4  -4
};
\addplot [draw=steelblue6999169, fill=steelblue6999169, mark=myStar]
table{%
	x  y
	0 -4
};
\addplot [draw=brown1613334, fill=brown1613334, mark=myStar]
table{%
	x  y
	-4  -4
};
\addplot [draw=goldenrod22215426, fill=goldenrod22215426, mark=myStar]
table{%
	x  y
	4 0
};
\addplot [draw=gold2512280, fill=gold2512280, mark=myStar]
table{%
	x  y
	-4  0
};
\addplot [draw=lightred, fill=lightred, mark=myStar]
table{%
	x  y
	4 4
};
\addplot [draw=darkgoldenrod16612945, fill=darkgoldenrod16612945, mark=myStar]
table{%
	x  y
	0  4
};
\addplot [draw=yellowgreen13918159, fill=yellowgreen13918159, mark=myStar]
table{%
	x  y
	-4 4
};

% add pale stars as goal points
\addplot [draw=darkcyan0149129, fill=darkcyan0149129, mark=myStar, opacity=0.3]
table{%
	x  y
	4.100952625695713 -3.9573136817180794
};
\addplot [draw=steelblue6999169, fill=steelblue6999169, mark=myStar, opacity=0.3]
table{%
	x  y
	0.14248165902625193 -3.9999529262871314
};
\addplot [draw=brown1613334, fill=brown1613334, mark=myStar, opacity=0.3]
table{%
	x  y
	-3.8827504246992612  -4.07848791444377
};
\addplot [draw=goldenrod22215426, fill=goldenrod22215426, mark=myStar, opacity=0.3]
table{%
	x  y
	3.9627848375305548 0.026258837801745694
};
\addplot [draw=gold2512280, fill=gold2512280, mark=myStar, opacity=0.3]
table{%
	x  y
	-3.964264798096442  -0.03384296581820321
};
\addplot [draw=lightred, fill=lightred, mark=myStar, opacity=0.3]
table{%
	x  y
	3.8506553992588977 4.083745405262078
};
\addplot [draw=darkgoldenrod16612945, fill=darkgoldenrod16612945, mark=myStar, opacity=0.3]
table{%
	x  y
	-0.011766345324228147  3.948628639496802
};
\addplot [draw=yellowgreen13918159, fill=yellowgreen13918159, mark=myStar, opacity=0.3]
table{%
	x  y
	-4.1124968499667816 4.024730490485631
};

% add arrows for orientation
\path [draw=darkcyan0149129, fill=darkcyan0149129]
(axis cs:4.113636970080078, -3.899666013360575)
--(axis cs:3.9716310162579824, -4.0405003453932045)
--(axis cs:4.394134012355871, -4.466518206859491)
--(axis cs:4.252128058533776, -4.607352538892121)
--(axis cs:4.746805653332178, -4.680112948487366)
--(axis cs:4.678145920000062, -4.184849542794232)
--(axis cs:4.536139966177966, -4.325683874826861)
--cycle;
\path [draw=steelblue6999169, fill=steelblue6999169]
(axis cs:0.09786304249351219, -3.988357433270978)
--(axis cs:-0.10208900199734583, -3.992736919990442)
--(axis cs:-0.08895054183895432, -4.592593053463016)
--(axis cs:-0.28890258632981225, -4.596972540182479)
--(axis cs:0.019784453845402374, -4.990307399085)
--(axis cs:0.3109535471427617, -4.583834080024088)
--(axis cs:0.1110015026519037, -4.588213566743552)
--cycle;
\path [draw=brown1613334, fill=brown1613334]
(axis cs:-3.883301907983397, -4.111572225063839)
--(axis cs:-4.0269689603279835, -3.9724328055213602)
--(axis cs:-4.444387218955419, -4.40343396255512)
--(axis cs:-4.588054271300005, -4.264294543012641)
--(axis cs:-4.650832531868083, -4.760337777015532)
--(axis cs:-4.157053114266247, -4.681712801640077)
--(axis cs:-4.300720166610833, -4.542573382097598)
--cycle;
\path [draw=goldenrod22215426, fill=goldenrod22215426]
(axis cs:3.976350913127308, 0.15753921424040399)
--(axis cs:3.9856177137633324, -0.04224598642764399)
--(axis cs:4.584973315767476, -0.014445584519572105)
--(axis cs:4.5942401164035, -0.21423078518762007)
--(axis cs:4.97991031678556, 0.1039806170864998)
--(axis cs:4.566439714495428, 0.3851248168165238)
--(axis cs:4.575706515131452, 0.18533961614847588)
--cycle;
\path [draw=gold2512280, fill=gold2512280]
(axis cs:-3.975040078330348, -0.1607905530573051)
--(axis cs:-3.9861687990034924, 0.03889958584839831)
--(axis cs:-4.585239215720603, 0.005513423828964456)
--(axis cs:-4.596367936393747, 0.2052035627346679)
--(axis cs:-4.979055133195438, -0.1165890869701765)
--(axis cs:-4.562981774374314, -0.39386685398244237)
--(axis cs:-4.574110495047458, -0.19417671507673895)
--cycle;
\path [draw=lightred, fill=lightred]
(axis cs:3.880398863070067, 4.11193735413866)
--(axis cs:4.027150233293574, 3.976054922097121)
--(axis cs:4.434797529418189, 4.41630903276764)
--(axis cs:4.581548899641696, 4.280426600726102)
--(axis cs:4.6331867083895135, 4.777752989235422)
--(axis cs:4.141294788971177, 4.688073896850717)
--(axis cs:4.288046159194683, 4.552191464809179)
--cycle;
\path [draw=darkgoldenrod16612945, fill=darkgoldenrod16612945]
(axis cs:-0.14855367412719866, 3.957584091744259)
--(axis cs:0.05095307443679288, 3.971621800867341)
--(axis cs:0.008839947067547642, 4.570142046559315)
--(axis cs:0.20834669563153918, 4.584179755682397)
--(axis cs:-0.11898884546061163, 4.962136689125758)
--(axis cs:-0.3901735500604354, 4.542066628313152)
--(axis cs:-0.1906668014964439, 4.556104337436234)
--cycle;
\path [draw=yellowgreen13918159, fill=yellowgreen13918159]
(axis cs:-4.0652671921441, 3.9787357837019286)
--(axis cs:-3.9260177787366795, 4.122296226950372)
--(axis cs:-4.356699108482008, 4.540044467172632)
--(axis cs:-4.2174496950745874, 4.683604910421075)
--(axis cs:-4.7134447016826035, 4.746763072363252)
--(axis cs:-4.635197935296849, 4.252923580675747)
--(axis cs:-4.495948521889428, 4.39648402392419)
--cycle;

% add rectangles
\path [draw=darkcyan0149129, fill=none, very thick]
(axis cs:4.537604493414799, -3.972133517508455)
--(axis cs:4.040583655037465, -4.465053679622659)
--(axis cs:3.5476634929232618, -3.968032841245325)
--(axis cs:4.044684331300595, -3.4751126791311218)
--cycle;
\path [draw=steelblue6999169, fill=none, very thick]
(axis cs:0.3554671998661464, -4.33279915273065)
--(axis cs:-0.34436495585185656, -4.3481273562487734)
--(axis cs:-0.35969315936998003, -3.6482952005307703)
--(axis cs:0.3401389963480229, -3.632966997012647)
--cycle;
\path [draw=brown1613334, fill=none, very thick]
(axis cs:-3.9472120767520016, -4.536913841094964)
--(axis cs:-4.450046759958053, -4.049925872696289)
--(axis cs:-3.9630587915593782, -3.547091189490236)
--(axis cs:-3.460224108353326, -4.034079157888911)
--cycle;
\path [draw=goldenrod22215426, fill=none, very thick]
(axis cs:4.314391513501362, 0.42348761618850583)
--(axis cs:4.346825315727446, -0.27576058614966203)
--(axis cs:3.647577113389278, -0.30819438837574586)
--(axis cs:3.6151433111631945, 0.391053813962422)
--cycle;
\path [draw=gold2512280, fill=none, very thick]
(axis cs:-4.310586920573898, -0.42987848786743743)
--(axis cs:-4.349537442929904, 0.2690369983025245)
--(axis cs:-3.6506219567599425, 0.30798752065853063)
--(axis cs:-3.6116714344039362, -0.3909279655114313)
--cycle;
\path [draw=lightred, fill=none, very thick]
(axis cs:3.9347539063633765, 4.538605292081719)
--(axis cs:4.448383702145649, 4.0630167799363335)
--(axis cs:3.9727951900002636, 3.5493869841540615)
--(axis cs:3.459165394217991, 4.024975496299447)
--cycle;
\path [draw=darkgoldenrod16612945, fill=none, very thick]
(axis cs:-0.4225031007975812, 4.2891737653273925)
--(axis cs:0.27577051917638923, 4.338305747258178)
--(axis cs:0.3249025011071754, 3.6400321272842078)
--(axis cs:-0.373371118866795, 3.5909001453534217)
--cycle;
\path [draw=yellowgreen13918159, fill=none, very thick]
(axis cs:-4.4905597345881505, 4.042971703104361)
--(axis cs:-4.003186787662179, 4.5454332544739104)
--(axis cs:-3.5007252362926295, 4.058060307547939)
--(axis cs:-3.9880981832186007, 3.55559875617839)
--cycle;
\end{axis}

\end{tikzpicture}

%% file: img/mpc-real-goal-state/scenario8a_mpc_real-goal-state_planning30_safety_distance0.2_data_plot_of_paths.tex
% This file was created with tikzplotlib v0.10.1.
\begin{tikzpicture}[scale=0.53]

\definecolor{brown1613334}{RGB}{161,33,34}
\definecolor{darkcyan0149129}{RGB}{0,149,129}
\definecolor{darkgoldenrod16612945}{RGB}{166,129,45}
\definecolor{darkgray176}{RGB}{176,176,176}
\definecolor{lightred}{RGB}{220,100,90}
\definecolor{gold2512280}{RGB}{251,228,0}
\definecolor{goldenrod22215426}{RGB}{222,154,26}
\definecolor{steelblue6999169}{RGB}{69,99,169}
\definecolor{yellowgreen13918159}{RGB}{139,181,59}

\pgfdeclareplotmark{myStar}{
	\node[star, draw, fill, star points=5, star point ratio=2.5, minimum size=0.25cm, inner sep = 0.3mm] {};
}

\begin{axis}[
tick align=outside,
tick pos=left,
x grid style={darkgray176},
xmin=-6, xmax=6,
xlabel={$p_x$ in m},
xtick style={color=black},
xtick={-6,-5,-4,-3,-2,-1,0,1,2,3,4,5,6},
xticklabels={-6, , , , , ,0, , , , , ,6},
y grid style={darkgray176},
ylabel={\, },
ymin=-6, ymax=6,
ytick style={color=black},
ytick={-6,-5,-4,-3,-2,-1,0,1,2,3,4,5,6},
yticklabels={-6, , , , , ,0, , , , , ,6}
]
\addplot [draw=darkcyan0149129, fill=darkcyan0149129, mark=*, mark size=0.5pt, only marks]
table{%
x  y
-4 4
-4 4
-3.99507514637513 3.99502565088066
-3.98045049975982 3.97995555601543
-3.9516568133705 3.94938031933761
-3.90469942323921 3.89747595507651
-3.83949114625506 3.82167274180027
-3.7577144462419 3.72063143382302
-3.66147711894839 3.5928255459744
-3.5533554900205 3.43660962051396
-3.43642581551124 3.2502858429641
-3.31591702759908 3.03928176649925
-3.18961731078021 2.79809071117563
-3.05859815100886 2.52745442111545
-2.92245778676761 2.22610916852397
-2.78107768464396 1.89430540187486
-2.63394893357615 1.53253297397226
-2.48263110614505 1.14779728416462
-2.32811777044959 0.747267070698717
-2.17103869410679 0.338212188347461
-2.01186547424514 -0.0720478515639643
-1.85111966796691 -0.476187754038253
-1.68956842648475 -0.866956022661097
-1.52844585959403 -1.23718271454549
-1.36961768108775 -1.57993573497764
-1.21447001675003 -1.89121180543403
-1.06325047748852 -2.17077568946349
-0.916585490024998 -2.41864862188677
-0.774015148861154 -2.63733920452232
-0.633842546467272 -2.83114932076735
-0.494031750208415 -3.004189310771
-0.35139302473584 -3.16104178815991
-0.202328899439032 -3.30541990551548
-0.0424971443522278 -3.44032891699872
0.132711937674879 -3.56744929964814
0.326290918987128 -3.68637823042935
0.534842267415121 -3.79465008578256
0.753436582332993 -3.89001059709855
0.977026914384261 -3.97119790264228
1.20127739159233 -4.03804100887191
1.42256538621769 -4.09110170895947
1.63774056903907 -4.13148986956157
1.84426000159725 -4.16030378049917
2.03989411891968 -4.17885412074066
2.22366150272725 -4.18904100022802
2.39538923121362 -4.19255572952753
2.55518648752573 -4.19102608424169
2.70358406367102 -4.18583852582155
2.84096667200027 -4.17821334940244
2.96775772877336 -4.16915075232875
3.08497108622739 -4.15947076372531
3.19303829747552 -4.14978721284276
3.29281011277034 -4.14063707535083
3.38479310554613 -4.13240628234305
3.46952402497598 -4.1254325535496
3.54740216264858 -4.1199265046631
3.61891756698886 -4.11604991615037
3.68430467801452 -4.11384287385848
3.74374866423598 -4.11334183403351
3.79749687336941 -4.11445213181042
3.84566113746931 -4.11708491863988
3.88842152711316 -4.12103400269343
3.92590987152075 -4.12609728404821
3.95820496594631 -4.13197075733548
3.98538321520376 -4.1383233131833
4.00756290326601 -4.14476412460969
4.02521509595936 -4.1508932411598
4.03884544232991 -4.15639409782174
4.04896422061731 -4.16104013926006
4.05606481617927 -4.16468852201474
4.0606099789846 -4.16726794404593
4.06302360015713 -4.16876484648464
4.06368617001352 -4.16920994013082
4.06293328900005 -4.16866628631108
4.06105801567341 -4.16722053135979
4.05831450018106 -4.16497614021041
4.05491970273991 -4.16204597491824
4.05105423887477 -4.15854453948209
4.04686908804223 -4.15458611013137
4.0424837517024 -4.15027601384332
4.03682859720025 -4.14452650185808
4.03081750507905 -4.13823272516805
4.02498447256574 -4.13196984205851
4.0186897691147 -4.1250699359757
4.01263042175114 -4.11831484542698
4.00643625900449 -4.11132117913219
4.00071554847021 -4.10480062206206
3.99505306593488 -4.09830953812582
3.989969563854 -4.09246384301994
3.98501650568438 -4.08676845838778
3.98066467684772 -4.08177387091672
3.97646448629146 -4.07697452685103
3.9728552731041 -4.072873286825
3.9700005720043 -4.06965025224409
3.96788437917117 -4.06727794310195
3.96640456308003 -4.06563154519095
3.96543315054141 -4.06455923473043
};
\addplot [draw=steelblue6999169, fill=steelblue6999169, mark=*, mark size=0.5pt, only marks]
table{%
x  y
0 4
0 4
-3.49036011776147e-05 3.9930087120137
-0.000349263620103627 3.97203023057243
-0.00160686474731761 3.93008510687223
-0.00509986883729701 3.86022962194586
-0.0125831383343907 3.76057226790399
-0.0261933101968415 3.63134993557009
-0.0484992846230012 3.47297673485171
-0.0823181015604637 3.28607544578791
-0.12882984556715 3.07111858248961
-0.187089665444767 2.83527621931824
-0.255239491721733 2.58541034247327
-0.330751495127744 2.32801117150947
-0.410688217340578 2.06909196709001
-0.491681653843179 1.81496842849176
-0.570254846835691 1.5718794980626
-0.643011550471424 1.34608368771133
-0.709606092764417 1.13514934491239
-0.771270333819266 0.932143290534008
-0.829136981897014 0.730137417767263
-0.884033891314172 0.522168736011604
-0.936274682666398 0.301218090968666
-0.98542542109457 0.0603190665834885
-1.02836629333916 -0.196985402521463
-1.0617694120189 -0.463883700740084
-1.08301440814363 -0.733291581262609
-1.09044108308327 -0.998044074713679
-1.08350341161807 -1.2515372105474
-1.06278025144724 -1.48839240226316
-1.02993551763386 -1.70791337392646
-0.987142599511914 -1.91355008728291
-0.936417437287835 -2.10651768195053
-0.879795365790363 -2.2862615249154
-0.819483815956272 -2.45195461234771
-0.75744903763448 -2.60340368027085
-0.695331275116507 -2.74106394735665
-0.634409405648088 -2.8657531533045
-0.57564972334773 -2.97846504848422
-0.519772967878699 -3.0802107271722
-0.467263434321819 -3.17203620618835
-0.417956667620581 -3.25579040697076
-0.371963483682 -3.33252101903746
-0.329672739081967 -3.40247824454872
-0.290783499021626 -3.4669271092085
-0.255276832695904 -3.52649369865842
-0.223172841654284 -3.58157783932854
-0.194772477957269 -3.63182772155682
-0.169611568747547 -3.67826070390835
-0.147565113520831 -3.72115573606609
-0.128533949692173 -3.76063619718723
-0.112385877299718 -3.79676766588173
-0.0989793235345434 -3.82952380301633
-0.0881040074805193 -3.85890075796786
-0.0794862941883812 -3.8849425477891
-0.0728088378101142 -3.90777902235101
-0.067750896348081 -3.92755383373451
-0.0640096968823947 -3.94444572891573
-0.0613082838365643 -3.95868681239003
-0.0594096618844144 -3.97051764119061
-0.0580706285948806 -3.98051272738219
-0.0571887608907175 -3.9885306507033
-0.0566343483118447 -3.99479535984857
-0.0562805709467381 -3.99990012515307
-0.0560826729095768 -4.00367150890592
-0.0559836820452477 -4.00628344870489
-0.055941396618277 -4.0079438843883
};
\addplot [draw=brown1613334, fill=brown1613334, mark=*, mark size=0.5pt, only marks]
table{%
x  y
4 4
4 4
3.99502563031933 3.99507512627007
3.97995549343889 3.9804504403768
3.94938019166208 3.95165669711528
3.8974757370599 3.90469923494137
3.82167242803939 3.83949089253346
3.72063101535417 3.75771413348314
3.59282501638391 3.6614767584435
3.43660871815045 3.55335491863429
3.25028434695493 3.43642482828302
3.03922596786354 3.31588484382762
2.79813598730305 3.18963760645035
2.52741485540821 3.05857691231681
2.22600864836543 2.92240866762887
1.8941836284648 2.78101886261329
1.53239969729141 2.63388432757081
1.14766242266477 2.48256427891924
0.74714061838308 2.32805208384323
0.338103178465866 2.17097710126602
-0.0721303741967428 2.01181107593804
-0.476234857804333 1.85107584307614
-0.866964531388321 1.68953688240656
-1.23714747423566 1.5284295596703
-1.57985172477097 1.36962014541528
-1.89108024427833 1.21449237856168
-2.1705975146868 1.06329431023687
-2.41842569027449 0.916652147380051
-2.63709065010327 0.774094925137855
-2.83089437987092 0.633923375523412
-3.0039384856816 0.494105633749442
-3.16080450391049 0.351450797617055
-3.30520293001238 0.202361577003649
-3.44013640548931 0.0424961605568447
-3.567281950325 -0.132753136104832
-3.68622987028985 -0.326368025396323
-3.79451382242943 -0.534946843644961
-3.88988160178404 -0.753560839209726
-3.97107306129459 -0.977163544291077
-4.0379182522581 -1.20142092522946
-4.09098099455888 -1.42271019864929
-4.13137041553485 -1.637882770713
-4.16018566443033 -1.84439656763393
-4.17873863448058 -2.04002314097646
-4.18892944560955 -2.22378239329624
-4.1924493436461 -2.39550224757097
-4.1909259042333 -2.55529226486786
-4.1857453618076 -2.70368326677613
-4.17812784593008 -2.84106020153334
-4.16907333259783 -2.96784697426522
-4.15940173900199 -3.08505699454635
-4.14972673726738 -3.19312183025496
-4.1405852534786 -3.29289172224422
-4.13236310318187 -3.38487312439349
-4.1253979642432 -3.46960237638078
-4.11990036716709 -3.54747855096659
-4.11603202967146 -3.6189917127114
-4.1138329587632 -3.68437629104703
-4.11333952655787 -3.74381732240214
-4.1144570233496 -3.79756209603332
-4.11709649612446 -3.84572247554049
-4.12105165858463 -3.88847837158582
-4.12612026332422 -3.92596151184482
-4.1319981697813 -3.95825070054049
-4.13835410596106 -3.98542242582722
-4.14479717263729 -4.00759519163309
-4.15092741682623 -4.02524053154296
-4.15642840158819 -4.03886438604947
-4.1610737532277 -4.04897718744203
-4.16472081076517 -4.05607238542503
-4.16729843492532 -4.06061274076215
-4.16879320507421 -4.06302212490825
-4.16923594784601 -4.06368099312679
-4.16868982110407 -4.06292490306118
-4.16724155173714 -4.06104686727866
-4.16499467079559 -4.05830098854393
-4.16206209307805 -4.0549041802287
-4.15855836178613 -4.0510370120234
-4.15459778214604 -4.04685042127029
-4.15028569893233 -4.04246387023036
-4.14453396094431 -4.0368074797622
-4.13823800875376 -4.03079523330343
-4.13197317550906 -4.02496119556779
-4.12507141404564 -4.01866558351663
-4.11831472427592 -4.01260547169354
-4.1113196657174 -4.00641067613278
-4.10479801556202 -4.00068948668421
-4.09830607789281 -3.99502666396594
-4.09245980886364 -3.98994296681433
-4.08676407222453 -3.98498984456775
-4.08176936351198 -3.98063807918763
-4.07697008414192 -3.97643806993876
-4.07286908349381 -3.97282914480099
-4.06964643849647 -3.96997482901153
-4.06727464049733 -3.96785910826148
-4.0656288477095 -3.96637983896127
-4.06455721197709 -3.96540903480399
};
\addplot [draw=goldenrod22215426, fill=goldenrod22215426, mark=*, mark size=0.5pt, only marks]
table{%
x  y
-4 0
-4 0
-3.99300815753066 -3.49094358926463e-05
-3.97202856912092 -0.000349304435154328
-3.93008193107019 -0.00160700360826518
-3.86022452793345 -0.00510021786991318
-3.76056527112466 -0.0125838315762693
-3.63135001577469 -0.0261933039630261
-3.47298403665011 -0.0484980379612175
-3.28609002771794 -0.0823159486643129
-3.07113965908332 -0.128827556248344
-2.83530313544334 -0.187088214354805
-2.58542673318361 -0.255244073876006
-2.32776675277223 -0.330836473174353
-2.06870135371819 -0.410822873890371
-1.81453063143273 -0.491836428317428
-1.57149269835872 -0.57039857945289
-1.34583527101887 -0.643116274440134
-1.13501202839374 -0.70968143320216
-0.932092076457901 -0.771325507865053
-0.730146956502016 -0.829181194300506
-0.522212733084126 -0.884076104871031
-0.301268438287986 -0.93632339341332
-0.0603564646023237 -0.98548601223325
0.196955489661008 -1.0284385317152
0.463855552157082 -1.06185325959393
0.733258647347439 -1.08310998935763
0.997998812492106 -1.09054881649954
1.25146950481004 -1.08362432370442
1.48829008105535 -1.0629166068913
1.70778191692474 -1.0300864590706
1.91340223941217 -0.987307026920003
2.1064006387965 -0.9365850830147
2.28617663150284 -0.87996362247427
2.45189274474826 -0.819653495057585
2.60335638066991 -0.757621325883504
2.7410269483852 -0.695506238968248
2.86572478880797 -0.634586232066901
2.97844513591015 -0.575827020496593
3.08019957692874 -0.519949173659568
3.17203441475539 -0.467436926796114
3.25579967477391 -0.418125295716615
3.33254260726633 -0.372125446920929
3.40251303947215 -0.329826627259916
3.46697520479986 -0.290928541130204
3.52655468208488 -0.255412725345512
3.58165101721264 -0.223299604794445
3.63191241038491 -0.19489031265291
3.6783560923605 -0.16972084763179
3.72126098120451 -0.147666321720243
3.760750435868 -0.128627657570167
3.79689004122977 -0.112472718993764
3.82965338342524 -0.0990600337846242
3.8590365180237 -0.0881794230436237
3.8850833647585 -0.079557333699947
3.90792375740927 -0.0728764220106479
3.92770119668295 -0.0678159182225474
3.94459432804 -0.0640729541691568
3.95883523117521 -0.0613704397279854
3.97066451955896 -0.0594712312179784
3.98065677751363 -0.0581319624227222
3.98867078072156 -0.057250109140344
3.99493062616706 -0.0566958625655409
4.00002978714266 -0.0563423056297053
4.00379488181084 -0.056144648795837
4.00639996845668 -0.0560458802844062
4.00805311579316 -0.0560037715451251
};
\addplot [draw=gold2512280, fill=gold2512280, mark=*, mark size=0.5pt, only marks]
table{%
x  y
4 0
4 0
3.99300815753066 3.49094358926461e-05
3.97202856912092 0.000349304435154321
3.93008193107003 0.00160700360827252
3.86022452793295 0.005100217869945
3.76056527112384 0.0125838315763429
3.63135001577353 0.0261933039631648
3.47298403664862 0.0484980379614492
3.28609002771613 0.0823159486646541
3.07113965908126 0.12882755624879
2.83530313544113 0.187088214355339
2.58542673318011 0.255244073876934
2.32776675274971 0.330836473180889
2.06870135363827 0.410822873914642
1.81453062302818 0.491836430995057
1.57149268622269 0.570398583336649
1.34583580620736 0.643116101932443
1.13501288506022 0.709681159159617
0.932093036245594 0.771325202385443
0.730147801638118 0.82918092146992
0.522213260810121 0.884075915543229
0.301268474134293 0.936323319996835
0.0603558493690617 0.985486071166642
-0.196957011212529 1.02843874130116
-0.463858238448488 1.06185361526911
-0.73326275281769 1.08311045797029
-0.998004573250293 1.09054933131056
-1.2514770334103 1.08362478800575
-1.48829920540904 1.06291692704848
-1.70779190104821 1.03008661883169
-1.91341234568773 0.987307098235799
-2.10641027510027 0.936585186496891
-2.28618549086128 0.879963844073138
-2.45190074968199 0.81965387214516
-2.60336357973236 0.757621855553968
-2.741033417963 0.695506902708383
-2.86573060691009 0.634587005434677
-2.97845037642512 0.575827876587774
-3.08020430183821 0.519950088468384
-3.17203867193838 0.467437881956332
-3.25580352645961 0.418126261008018
-3.33254610780924 0.372126395092623
-3.40251622627599 0.329827542103567
-3.46697811248123 0.290929406887868
-3.52655734294237 0.255413528140448
-3.58165346104012 0.22330033341969
-3.63191465170624 0.194890968113162
-3.67835815825105 0.169721422968896
-3.72126289863692 0.147666811432951
-3.76075223100004 0.128628059024143
-3.79689173741272 0.112473033179123
-3.82965500059705 0.0990602660938429
-3.85903807213353 0.0881795829031332
-3.88508486705444 0.0795574334721387
-3.90792521498164 0.0728764749248393
-3.9277026115815 0.0678159369824708
-3.94459569925749 0.0640729497094222
-3.95883655534103 0.0613704209041732
-3.97066579260271 0.0594712046852861
-3.9806579948445 0.0581319325640197
-3.98867193801138 0.0572500791853566
-3.99493171959961 0.0566958344222091
-4.00003081466043 0.0563422799549616
-4.00379584097982 0.0561446258375891
-4.00640085759036 0.0560458597944084
-4.0080539341089 0.0560037529694131
};
\addplot [draw=lightred, fill=lightred, mark=*, mark size=0.5pt, only marks]
table{%
x  y
-4 -4
-4 -4
-3.99502563031933 -3.99507512627007
-3.97995549343889 -3.9804504403768
-3.94938019166208 -3.95165669711528
-3.8974757370599 -3.90469923494137
-3.82167242803939 -3.83949089253346
-3.72063101535417 -3.75771413348314
-3.59282501638391 -3.66147675844351
-3.43660871815045 -3.55335491863429
-3.25028434695492 -3.43642482828302
-3.03922596784534 -3.31588484381739
-2.798135987256 -3.18963760642537
-2.52741485505642 -3.05857691214564
-2.22600864771689 -2.92240866732418
-1.89418241328703 -2.78101834737147
-1.53239761209365 -2.63388345848934
-1.1476598164865 -2.48256320205694
-0.747137843688546 -2.32805093602335
-0.338100569211753 -2.17097600788132
0.0721324853743392 -2.01181016365184
0.476236140602198 -1.85107524525247
0.866964644070761 -1.68953675046231
1.23714608123854 -1.52843006272135
1.57984849123858 -1.36962147909385
1.89107521298979 -1.21449458482796
2.17059072253365 -1.06329744471803
2.41841720124744 -0.916656261144974
2.63708111571972 -0.774099695200354
2.83088450349259 -0.633928367004264
3.00392868854097 -0.494110534093055
3.16079517771239 -0.351455240444613
3.30519437270813 -0.202365193393438
3.44012881467335 -0.0424985941465172
3.56727538820841 0.132752165914618
3.68622411215003 0.326368377681388
3.79450860441725 0.534948207565038
3.88987674155091 0.753562941001735
3.97106844051609 0.977166133594695
4.0379137883928 1.2014238294059
4.09097668338889 1.42271323364546
4.13136619984257 1.63788582211257
4.16018153555903 1.84439954353786
4.17873472993773 2.04002608435456
4.1889258780633 2.22378532551967
4.1924461902578 2.3955052242273
4.19092322676486 2.55529533372547
4.1857432082384 2.7036864838647
4.17812625540996 2.84106363129861
4.16907232308931 2.96785074651181
4.15940132570959 3.08506117454792
4.14972692532513 3.19312648105086
4.14058604990629 3.29289684748235
4.13236450835032 3.38487871755113
4.1253999789923 3.46960839148151
4.11990298724752 3.54748492683681
4.1160352483722 3.61899838197114
4.11383676381304 3.68438318645159
4.11334389966208 3.74382436292775
4.11446194155461 3.79756919782813
4.11710192728743 3.8457295535779
4.12105756113495 3.88848533077979
4.1261265818849 3.92596824776958
4.13200483560223 3.95825711108577
4.13836103463639 3.98542841560251
4.14480427033582 4.00760068544031
4.15093458851412 4.02524549510898
4.15643555974884 4.03886881159601
4.16108082161588 4.04898108302326
4.16472772496146 4.05607576820273
4.16730514169956 4.06061563307787
4.1687996614425 4.06302455198971
4.16924212017554 4.06368298176191
4.1686956842927 4.06292648072962
4.16724708847944 4.06104806154174
4.16499987083467 4.05830182653253
4.16206695240373 4.05490468823886
4.15856288171156 4.05103721515035
4.15460196852814 4.04685034324462
4.15028956120835 4.04246353320915
4.14453747503626 4.0368068862258
4.13824117716473 4.03079439543524
4.13197601757717 4.02496013172153
4.12507393185194 4.01866430227707
4.11831693995254 4.01260398949404
4.11132159296591 4.00640900455751
4.10479968408135 4.00068764411201
4.0983075080009 3.99502466391315
4.0924610326186 3.98994082833011
4.08676511199211 3.98498758191313
4.0817702501197 3.98063571107606
4.07697083967623 3.9764356108594
4.07286973576087 3.9728266125705
4.06964701596456 3.96997224255408
4.06727516925665 3.96785648587873
4.06562935020415 3.96637719732907
4.06455770678637 3.96540638852123
};
\addplot [draw=darkgoldenrod16612945, fill=darkgoldenrod16612945, mark=*, mark size=0.5pt, only marks]
table{%
x  y
0 -4
0 -4
3.49036011777514e-05 -3.99300871201369
0.000349263620104572 -3.9720302305724
0.00160686474730431 -3.93008510687253
0.00509986883723357 -3.86022962194685
0.0125831383342413 -3.76057226790568
0.0261933101965593 -3.63134993557247
0.0484992846225287 -3.47297673485477
0.0823181015597676 -3.28607544579165
0.128829845566239 -3.07111858249385
0.187089665443717 -2.83527621932273
0.255239491807981 -2.58541034214778
0.330751495324288 -2.32801117078909
0.41068822743065 -2.06909193422801
0.491681673914423 -1.81496836414881
0.570254881262706 -1.57187938908572
0.643011645149782 -1.34608339149202
0.709606242283502 -1.13514887548709
0.77127052929165 -0.93214267301349
0.82913719331823 -0.730136750822773
0.88403408630681 -0.522168141410132
0.93627481397365 -0.301217781288159
0.985425457744819 -0.060319244821679
1.02836618600829 0.196984324682601
1.06176914576956 0.463881304593688
1.08301401119373 0.733287450139573
1.09044063474478 0.998037810947
1.0835030396753 1.25152856584784
1.06278009192375 1.48838149514434
1.02993559937795 1.70790114360655
0.987142918709089 1.91353755302012
0.936417923665848 2.10650553668671
0.879795898018483 2.28625021585685
0.81948432177874 2.45194430743289
0.757449489036165 2.60339438124182
0.695331664982634 2.7410556030005
0.634409736095781 2.86574569786508
0.575650001765485 2.97845841244871
0.519773200926776 3.080204848333
0.467263625587165 3.17203103277957
0.417956834620534 3.25578587302908
0.371963643787703 3.33251706370326
0.32967289601968 3.40247483001744
0.290783665439714 3.46692418829326
0.255277022380159 3.52649122371107
0.223173066417618 3.58157576593568
0.194772735064571 3.63182602545544
0.169611868849981 3.6782593412305
0.147565466712573 3.72115466233512
0.128534363451574 3.76063536794327
0.112386355121623 3.79676703984081
0.0989798640224691 3.82952334240976
0.0881046045441912 3.85890043000022
0.0794869385011231 3.88494232558654
0.0728095191094554 3.90777888455868
0.06775160507323 3.92755376422899
0.0640104248342301 3.94444571597242
0.0613090245998454 3.95868684685308
0.059410410653863 3.97051771579779
0.0580713825133642 3.98051283393954
0.0571895175165227 3.98853078379066
0.0566351061156868 3.99479551450718
0.0562813292224501 3.99990029572077
0.0560834311491452 4.00367169125627
0.0559844400451326 4.00628363935409
0.0559421543685441 4.00794408024541
};
\addplot [draw=yellowgreen13918159, fill=yellowgreen13918159, mark=*, mark size=0.5pt, only marks]
table{%
x  y
4 -4
4 -4
3.99507514637513 -3.99502565088066
3.98045049975982 -3.97995555601543
3.9516568133705 -3.94938031933761
3.90469942323921 -3.89747595507651
3.83949114625506 -3.82167274180027
3.7577144462419 -3.72063143382302
3.66147711894839 -3.5928255459744
3.55335549002051 -3.43660962051396
3.43642581551124 -3.2502858429641
3.3159170275894 -3.03928176648432
3.18961731079408 -2.79809071121054
3.05859815037118 -2.52745441980584
2.9224577853236 -2.2261091654053
2.78107717194499 -1.89430418234666
2.63394809373568 -1.53253092378536
2.48263010848265 -1.1477947958048
2.32811677499691 -0.747264540367985
2.1710378570391 -0.338210012619877
2.01186495536355 0.0720492731307811
1.85111963500391 0.476188025877365
1.68956904097562 0.866954802637512
1.52844729002826 1.23717969535643
1.36962012211556 1.57993060864858
1.21447351274682 1.89120463284125
1.06325509152904 2.17076651573925
0.916591277435867 2.41863752548072
0.77402171074943 2.6373269750745
0.633849349116898 2.83113680864751
0.494038435891655 3.00417698984008
0.351399172818558 3.1610301022048
0.20233408847022 3.30540919092072
0.0425009709923766 3.44031939474895
-0.132709788230241 3.56744103783573
-0.326290287676386 3.6863709683782
-0.534842805315289 3.79464351987068
-0.753437982652236 3.89000451524304
-0.977028899064621 3.97119216408878
-1.20127976549813 4.0380355157659
-1.42256795016669 4.09109644408415
-1.63774319041291 4.13148479225886
-1.84426257790718 4.16029887501807
-2.03989657495331 4.17884940503262
-2.22366381287972 4.18903649620554
-2.39539139536039 4.19255146167881
-2.55518851766344 4.19102207279355
-2.70358597081228 4.18583478571603
-2.84096847358998 4.17820989202602
-2.96775945722037 4.16914758376864
-3.08497276180317 4.15946788783318
-3.19303994171915 4.14978462950918
-3.29281173346258 4.14063478364053
-3.38479470772377 4.13240427825837
-3.46952560248675 4.12543083224687
-3.54740370310098 4.11992505912971
-3.61891906100641 4.11604873770292
-3.68430611536951 4.11384195168008
-3.74375003189206 4.11334115507099
-3.79749815629754 4.11445168189769
-3.84566232322945 4.11708468087526
-3.88842259861987 4.12103395768768
-3.92591080983937 4.12609740859028
-3.95820575279351 4.13197102471851
-3.98538383500295 4.13832369245134
-4.0075633464001 4.14476458311648
-4.02521536546198 4.15089374619333
-4.03884554917322 4.15639462052853
-4.0489641799619 4.16104065625632
-4.05606464487212 4.16468901539735
-4.06060969403894 4.16726840077326
-4.06302321787225 4.16876525760426
-4.06368570552375 4.16921030006723
-4.06293275604552 4.16866659224663
-4.06105742650974 4.16722078271949
-4.0583138655511 4.16497633820299
-4.05491903191146 4.16204612212209
-4.05105353969789 4.15854463943989
-4.04686836708862 4.15458616705424
-4.04248301437919 4.1502760322949
-4.0368278417084 4.1445264747404
-4.03081673263233 4.13823265356201
-4.02498368640324 4.1319697319375
-4.01868897132448 4.12506978996249
-4.01262961567311 4.11831467038316
-4.00643544783412 4.11132098072845
-4.00071473587768 4.10480040814806
-3.99505225531941 4.09830931504323
-3.98996875863524 4.09246361783466
-3.98501570901786 4.08676823673741
-3.98066389164328 4.08177365832172
-3.97646371529177 4.0769743277786
-3.97285451861865 4.07287310513581
-3.96999983602626 4.06965009100622
-3.9678836633975 4.06727780459671
-3.96640386889479 4.06563143100561
-3.96543247900134 4.06455914587277
};

% add circles
\draw[draw=darkcyan0149129,fill=none, very thick] (axis cs:3.96543315054141,-4.06455923473043) circle (70.0);
\draw[draw=steelblue6999169,fill=none, very thick] (axis cs:-0.055941396618277,-4.0079438843883) circle (70.0);
\draw[draw=brown1613334,fill=none, very thick] (axis cs:-4.06455721197709,-3.96540903480399) circle (70.0);
\draw[draw=goldenrod22215426,fill=none, very thick] (axis cs:4.00805311579316,-0.0560037715451251) circle (70.0);
\draw[draw=gold2512280,fill=none, very thick] (axis cs:-4.0080539341089,0.0560037529694131) circle (70.0);
\draw[draw=lightred,fill=none, very thick] (axis cs:4.06455770678637,3.96540638852123) circle (70.0);
\draw[draw=darkgoldenrod16612945,fill=none, very thick] (axis cs:0.0559421543685441,4.00794408024541) circle (70.0);
\draw[draw=yellowgreen13918159,fill=none, very thick] (axis cs:-3.96543247900134,4.06455914587277) circle (70.0);

% add solid stars as goal points
\addplot [draw=darkcyan0149129, fill=darkcyan0149129, mark=myStar]
table{%
	x  y
	4  -4
};
\addplot [draw=steelblue6999169, fill=steelblue6999169, mark=myStar]
table{%
	x  y
	0 -4
};
\addplot [draw=brown1613334, fill=brown1613334, mark=myStar]
table{%
	x  y
	-4  -4
};
\addplot [draw=goldenrod22215426, fill=goldenrod22215426, mark=myStar]
table{%
	x  y
	4 0
};
\addplot [draw=gold2512280, fill=gold2512280, mark=myStar]
table{%
	x  y
	-4  0
};
\addplot [draw=lightred, fill=lightred, mark=myStar]
table{%
	x  y
	4 4
};
\addplot [draw=darkgoldenrod16612945, fill=darkgoldenrod16612945, mark=myStar]
table{%
	x  y
	0  4
};
\addplot [draw=yellowgreen13918159, fill=yellowgreen13918159, mark=myStar]
table{%
	x  y
	-4 4
};

% add arrows for orientation
\path [draw=darkcyan0149129, fill=darkcyan0149129]
(axis cs:4.03927751958576, -3.997127720136423)
--(axis cs:3.8915887814970596, -4.131990749324437)
--(axis cs:4.296177869061101, -4.575056963590539)
--(axis cs:4.148489130972401, -4.709919992778554)
--(axis cs:4.63974829648148, -4.803002925173934)
--(axis cs:4.591555345238503, -4.305330905214511)
--(axis cs:4.443866607149802, -4.440193934402525)
--cycle;
\path [draw=steelblue6999169, fill=steelblue6999169]
(axis cs:0.044047288594882744, -4.006439614019101)
--(axis cs:-0.15593008183143675, -4.0094481547575)
--(axis cs:-0.14690445961623888, -4.6093802660364585)
--(axis cs:-0.34688183004255835, -4.612388806774858)
--(axis cs:-0.04089869292628053, -5.007830736519898)
--(axis cs:0.25305028123640005, -4.60336318455966)
--(axis cs:0.05307291081008063, -4.606371725298059)
--cycle;
\path [draw=brown1613334, fill=brown1613334]
(axis cs:-3.9971254349126424, -4.039253164170722)
--(axis cs:-4.131988989041537, -3.891564905437258)
--(axis cs:-4.57505376524193, -4.296155567823942)
--(axis cs:-4.709917319370825, -4.148467309090478)
--(axis cs:-4.802998505644411, -4.639726805448464)
--(axis cs:-4.305326656984141, -4.59153208529087)
--(axis cs:-4.440190211113035, -4.443843826557407)
--cycle;
\path [draw=goldenrod22215426, fill=goldenrod22215426]
(axis cs:4.0065491552628325, 0.04398491832889626)
--(axis cs:4.009557076323488, -0.15599246141914647)
--(axis cs:4.609489215567616, -0.1469686982371798)
--(axis cs:4.612497136628272, -0.34694607798522253)
--(axis cs:5.007940014533374, -0.04096416624184735)
--(axis cs:4.603473373446305, 0.2529860612589056)
--(axis cs:4.606481294506961, 0.053008681510862916)
--cycle;
\path [draw=gold2512280, fill=gold2512280]
(axis cs:-4.006549992002723, -0.04398493718172986)
--(axis cs:-4.0095578762150765, 0.15599244312055607)
--(axis cs:-4.609490017121934, 0.14696879048349595)
--(axis cs:-4.6124979013342875, 0.3469461707857819)
--(axis cs:-5.00794083562033, 0.04096433190764623)
--(axis cs:-4.6034742486972275, -0.2529859701210759)
--(axis cs:-4.606482132909581, -0.05300858981878998)
--cycle;
\path [draw=lightred, fill=lightred]
(axis cs:3.9971259982144094, 4.039250580432756)
--(axis cs:4.13198941535833, 3.8915621966097045)
--(axis cs:4.575054566827484, 4.296152448041465)
--(axis cs:4.709917983971405, 4.148464064218413)
--(axis cs:4.802999625901627, 4.6397234742408315)
--(axis cs:4.305327732539643, 4.591529215687568)
--(axis cs:4.440191149683564, 4.443840831864517)
--cycle;
\path [draw=darkgoldenrod16612945, fill=darkgoldenrod16612945]
(axis cs:-0.044046530626673026, 4.006439795389657)
--(axis cs:0.15593083936376123, 4.0094483651011625)
--(axis cs:0.14690513022924281, 4.609380475072466)
--(axis cs:0.346882500219677, 4.6123890447839715)
--(axis cs:0.0408993058110134, 5.0078309301975805)
--(axis cs:-0.25304960975162566, 4.603363335649453)
--(axis cs:-0.05307223976119144, 4.606371905360959)
--cycle;
\path [draw=yellowgreen13918159, fill=yellowgreen13918159]
(axis cs:-4.039276834199626, 3.9971276161159164)
--(axis cs:-3.8915881238030545, 4.131990675629623)
--(axis cs:-4.296177302344176, 4.575056806819337)
--(axis cs:-4.148488591947605, 4.709919866333045)
--(axis cs:-4.639747776569876, 4.803002697855626)
--(axis cs:-4.591554723137318, 4.305330687791923)
--(axis cs:-4.443866012740747, 4.44019374730563)
--cycle;

% add rectangles
\path [draw=darkcyan0149129, fill=none, very thick]
(axis cs:4.459898743275661, -4.087004225306631)
--(axis cs:3.9429881599652084, -4.559024827464681)
--(axis cs:3.470967557807159, -4.042114244154228)
--(axis cs:3.9878781411176116, -3.570093641996179)
--cycle;
\path [draw=steelblue6999169, fill=none, very thick]
(axis cs:0.2992839479199808, -4.352639336342161)
--(axis cs:-0.4006368485721373, -4.363169228926559)
--(axis cs:-0.41116674115653484, -3.66324843243444)
--(axis cs:0.2887540553355833, -3.6527185398500426)
--cycle;
\path [draw=brown1613334, fill=none, very thick]
(axis cs:-4.087000445035087, -4.459874707313118)
--(axis cs:-4.559022884486218, -3.9429658017459936)
--(axis cs:-4.042113978919093, -3.470943362294862)
--(axis cs:-3.5700915394679615, -3.9878522678619865)
--cycle;
\path [draw=goldenrod22215426, fill=none, very thick]
(axis cs:4.352749668496088, 0.2992205048700968)
--(axis cs:4.363277392208382, -0.40070032424805263)
--(axis cs:3.6633565630902325, -0.411228047960347)
--(axis cs:3.6528288393779382, 0.28869278115780245)
--cycle;
\path [draw=gold2512280, fill=none, very thick]
(axis cs:-4.352750552266282, -0.2992204599312056)
--(axis cs:-4.3632781470095185, 0.400700371126795)
--(axis cs:-3.663357315951518, 0.41122796587003185)
--(axis cs:-3.652829721208281, -0.2886928651879688)
--cycle;
\path [draw=lightred, fill=none, very thick]
(axis cs:4.0870013984748494, 4.459872040213431)
--(axis cs:4.55902335847857, 3.9429626968327502)
--(axis cs:4.04211401509789, 3.4709407368290295)
--(axis cs:3.570092055094169, 3.98785008020971)
--cycle;
\path [draw=darkgoldenrod16612945, fill=none, very thick]
(axis cs:-0.29928324010985147, 4.352639480733534)
--(axis cs:0.40063755485666824, 4.363169474723805)
--(axis cs:0.4111675488469397, 3.6632486797572854)
--(axis cs:-0.28875324611958003, 3.652718685767014)
--cycle;
\path [draw=yellowgreen13918159, fill=none, very thick]
(axis cs:-4.459898076344327, 4.087004034917782)
--(axis cs:-3.942987589956328, 4.559024743215757)
--(axis cs:-3.470966881658353, 4.042114256827758)
--(axis cs:-3.9878773680463526, 3.570093548529783)
--cycle;
\end{axis}

\end{tikzpicture}

%% file: img/box_plots/box_plot_duration.tex
% This file was created with tikzplotlib v0.10.1.
\begin{tikzpicture}[scale=0.48]

\definecolor{darkgray176}{RGB}{176,176,176}
\definecolor{darkgoldenrod16612945}{RGB}{166,129,45}
\definecolor{lightblue}{RGB}{173,216,230}
\definecolor{yellowgreen13918159}{RGB}{139,181,59}

\begin{axis}[
tick align=outside,
tick pos=left,
x grid style={darkgray176},
xmin=-0.5, xmax=6.5,
xtick style={color=black},
xtick={0.5,3,5.5},
xticklabels={1,2,3},
y grid style={darkgray176},
ylabel={$t$ in s},
ymin=4.57, ymax=27.23,
ytick style={color=black}
]
\path [draw=black, fill=lightblue]
(axis cs:-0.4,6.5)
--(axis cs:0.4,6.5)
--(axis cs:0.4,7.9)
--(axis cs:-0.4,7.9)
--(axis cs:-0.4,6.5)
--cycle;
\addplot [black]
table {%
0 6.5
0 5.6
};
\addplot [black]
table {%
0 7.9
0 9.3
};
\addplot [black]
table {%
-0.2 5.6
0.2 5.6
};
\addplot [black]
table {%
-0.2 9.3
0.2 9.3
};
\path [draw=black, fill=lightblue]
(axis cs:2.1,9.425)
--(axis cs:2.9,9.425)
--(axis cs:2.9,12.125)
--(axis cs:2.1,12.125)
--(axis cs:2.1,9.425)
--cycle;
\addplot [black]
table {%
2.5 9.425
2.5 6.5
};
\addplot [black]
table {%
2.5 12.125
2.5 15.3
};
\addplot [black]
table {%
2.3 6.5
2.7 6.5
};
\addplot [black]
table {%
2.3 15.3
2.7 15.3
};
\addplot [red, mark=o, mark size=1, mark options={solid,fill opacity=0,draw=black}, only marks]
table {%
2.5 26.2
2.5 18.5
};
\path [draw=black, fill=lightblue]
(axis cs:4.6,10.3)
--(axis cs:5.4,10.3)
--(axis cs:5.4,13.6)
--(axis cs:4.6,13.6)
--(axis cs:4.6,10.3)
--cycle;
\addplot [black]
table {%
5 10.3
5 6.2
};
\addplot [black]
table {%
5 13.6
5 18.5
};
\addplot [black]
table {%
4.8 6.2
5.2 6.2
};
\addplot [black]
table {%
4.8 18.5
5.2 18.5
};
\addplot [red, mark=o, mark size=1, mark options={solid,fill opacity=0,draw=black}, only marks]
table {%
5 26.3
5 20.5
};
\path [draw=black, fill=yellowgreen13918159]
(axis cs:0.6,6.16875)
--(axis cs:1.4,6.16875)
--(axis cs:1.4,6.94875)
--(axis cs:0.6,6.94875)
--(axis cs:0.6,6.16875)
--cycle;
\addplot [black]
table {%
1 6.16875
1 5.6
};
\addplot [black]
table {%
1 6.94875
1 7.75714285714286
};
\addplot [black]
table {%
0.8 5.6
1.2 5.6
};
\addplot [black]
table {%
0.8 7.75714285714286
1.2 7.75714285714286
};
\path [draw=black, fill=yellowgreen13918159]
(axis cs:3.1,7.59666666666667)
--(axis cs:3.9,7.59666666666667)
--(axis cs:3.9,9.5875)
--(axis cs:3.1,9.5875)
--(axis cs:3.1,7.59666666666667)
--cycle;
\addplot [black]
table {%
3.5 7.59666666666667
3.5 6.1
};
\addplot [black]
table {%
3.5 9.5875
3.5 11.4166666666667
};
\addplot [black]
table {%
3.3 6.1
3.7 6.1
};
\addplot [black]
table {%
3.3 11.4166666666667
3.7 11.4166666666667
};
\addplot [red, mark=o, mark size=1, mark options={solid,fill opacity=0,draw=black}, only marks]
table {%
3.5 13.75
};
\path [draw=black, fill=yellowgreen13918159]
(axis cs:5.6,8.63333333333333)
--(axis cs:6.4,8.63333333333333)
--(axis cs:6.4,10.9583333333333)
--(axis cs:5.6,10.9583333333333)
--(axis cs:5.6,8.63333333333333)
--cycle;
\addplot [black]
table {%
6 8.63333333333333
6 5.86666666666667
};
\addplot [black]
table {%
6 10.9583333333333
6 13.45
};
\addplot [black]
table {%
5.8 5.86666666666667
6.2 5.86666666666667
};
\addplot [black]
table {%
5.8 13.45
6.2 13.45
};
\addplot [red, mark=o, mark size=1, mark options={solid,fill opacity=0,draw=black}, only marks]
table {%
6 16.5714285714286
};
\addplot [darkgoldenrod16612945]
table {%
-0.4 6.5
0.4 6.5
};
\addplot [darkgoldenrod16612945]
table {%
2.1 10.55
2.9 10.55
};
\addplot [darkgoldenrod16612945]
table {%
4.6 11.7
5.4 11.7
};
\addplot [darkgoldenrod16612945]
table {%
0.6 6.4
1.4 6.4
};
\addplot [darkgoldenrod16612945]
table {%
3.1 8.6375
3.9 8.6375
};
\addplot [darkgoldenrod16612945]
table {%
5.6 9.5
6.4 9.5
};
\end{axis}

\end{tikzpicture}

%% file: img/box_plots/box_plot_driven_distances.tex
% This file was created with tikzplotlib v0.10.1.
\begin{tikzpicture}[scale=0.48]

\definecolor{darkgray176}{RGB}{176,176,176}
\definecolor{darkgoldenrod16612945}{RGB}{166,129,45}
\definecolor{lightblue}{RGB}{173,216,230}
\definecolor{yellowgreen13918159}{RGB}{139,181,59}

\begin{axis}[
log basis y={10},
tick align=outside,
tick pos=left,
x grid style={darkgray176},
xmin=-0.5, xmax=6.5,
xtick style={color=black},
xtick={0.5,3,5.5},
xticklabels={1,2,3},
ylabel={Distance in m},
y grid style={darkgray176},
ymin=7, ymax=150,
ymode=log,
ytick style={color=black},
ytick={0.1,1,10,100},
yticklabels={0.1,1,10,100}
]
\path [draw=black, fill=lightblue]
(axis cs:-0.4,28.13)
--(axis cs:0.4,28.13)
--(axis cs:0.4,46.666)
--(axis cs:-0.4,46.666)
--(axis cs:-0.4,28.13)
--cycle;
\addplot [black]
table {%
0 28.13
0 16.35
};
\addplot [black]
table {%
0 46.666
0 71.728
};
\addplot [black]
table {%
-0.2 16.35
0.2 16.35
};
\addplot [black]
table {%
-0.2 71.728
0.2 71.728
};
\addplot [red, mark=o, mark size=1, mark options={solid,fill opacity=0,draw=black}, only marks]
table {%
0 85.156
};
\path [draw=black, fill=lightblue]
(axis cs:2.1,32.2205)
--(axis cs:2.9,32.2205)
--(axis cs:2.9,55.70925)
--(axis cs:2.1,55.70925)
--(axis cs:2.1,32.2205)
--cycle;
\addplot [black]
table {%
2.5 32.2205
2.5 17.334
};
\addplot [black]
table {%
2.5 55.70925
2.5 82.436
};
\addplot [black]
table {%
2.3 17.334
2.7 17.334
};
\addplot [black]
table {%
2.3 82.436
2.7 82.436
};
\addplot [red, mark=o, mark size=1, mark options={solid,fill opacity=0,draw=black}, only marks]
table {%
2.5 93.766
};
\path [draw=black, fill=lightblue]
(axis cs:4.6,38.076)
--(axis cs:5.4,38.076)
--(axis cs:5.4,72.3905)
--(axis cs:4.6,72.3905)
--(axis cs:4.6,38.076)
--cycle;
\addplot [black]
table {%
5 38.076
5 19.299
};
\addplot [black]
table {%
5 72.3905
5 110.28
};
\addplot [black]
table {%
4.8 19.299
5.2 19.299
};
\addplot [black]
table {%
4.8 110.28
5.2 110.28
};
\addplot [red, mark=o, mark size=1, mark options={solid,fill opacity=0,draw=black}, only marks]
table {%
5 130.806
5 177.436
5 130.191
};
\path [draw=black, fill=yellowgreen13918159]
(axis cs:0.6,9.58258333333333)
--(axis cs:1.4,9.58258333333333)
--(axis cs:1.4,10.3835)
--(axis cs:0.6,10.3835)
--(axis cs:0.6,9.58258333333333)
--cycle;
\addplot [black]
table {%
1 9.58258333333333
1 8.58024999999999
};
\addplot [black]
table {%
1 10.3835
1 11.507
};
\addplot [black]
table {%
0.8 8.58024999999999
1.2 8.58024999999999
};
\addplot [black]
table {%
0.8 11.507
1.2 11.507
};
\addplot [red, mark=o, mark size=1, mark options={solid,fill opacity=0,draw=black}, only marks]
table {%
1 8.175
1 8.20133333333333
};
\path [draw=black, fill=yellowgreen13918159]
(axis cs:3.1,10.7994583333333)
--(axis cs:3.9,10.7994583333333)
--(axis cs:3.9,11.72925)
--(axis cs:3.1,11.72925)
--(axis cs:3.1,10.7994583333333)
--cycle;
\addplot [black]
table {%
3.5 10.7994583333333
3.5 9.65
};
\addplot [black]
table {%
3.5 11.72925
3.5 12.6795
};
\addplot [black]
table {%
3.3 9.65
3.7 9.65
};
\addplot [black]
table {%
3.3 12.6795
3.7 12.6795
};
\addplot [red, mark=o, mark size=1, mark options={solid,fill opacity=0,draw=black}, only marks]
table {%
3.5 8.667
3.5 9.274
3.5 14.00775
};
\path [draw=black, fill=yellowgreen13918159]
(axis cs:5.6,11.9728333333333)
--(axis cs:6.4,11.9728333333333)
--(axis cs:6.4,16.478125)
--(axis cs:5.6,16.478125)
--(axis cs:5.6,11.9728333333333)
--cycle;
\addplot [black]
table {%
6 11.9728333333333
6 9.6495
};
\addplot [black]
table {%
6 16.478125
6 21.801
};
\addplot [black]
table {%
5.8 9.6495
6.2 9.6495
};
\addplot [black]
table {%
5.8 21.801
6.2 21.801
};
\addplot [red, mark=o, mark size=1, mark options={solid,fill opacity=0,draw=black}, only marks]
table {%
6 25.348
};
\addplot [darkgoldenrod16612945]
table {%
-0.4 38.3455
0.4 38.3455
};
\addplot [darkgoldenrod16612945]
table {%
2.1 45.7435
2.9 45.7435
};
\addplot [darkgoldenrod16612945]
table {%
4.6 52.388
5.4 52.388
};
\addplot [darkgoldenrod16612945]
table {%
0.6 9.936125
1.4 9.936125
};
\addplot [darkgoldenrod16612945]
table {%
3.1 11.4109583333333
3.9 11.4109583333333
};
\addplot [darkgoldenrod16612945]
table {%
5.6 14.4565
6.4 14.4565
};
\end{axis}

\end{tikzpicture}

%% file: img/box_plots/box_plot_minimum_distances_between_agents.tex
% This file was created with tikzplotlib v0.10.1.
\begin{tikzpicture}[scale=0.48]

\definecolor{darkgray176}{RGB}{176,176,176}
\definecolor{steelblue6999169}{RGB}{69,99,169}

\begin{axis}[
tick align=outside,
tick pos=left,
x grid style={darkgray176},
xmin=-0.25, xmax=2.25,
xtick style={color=black},
xtick={0,1,2},
xticklabels={1,2,3},
y grid style={darkgray176},
ylabel={Distance in im},
ymin=0.7, ymax=1.6,
ytick style={color=black}
]
\draw[red, thick] (axis cs:-0.5,0.99) -- (axis cs:7,0.99);
\addplot [draw=steelblue6999169, fill=steelblue6999169, mark=*, only marks]
table{%
x  y
0 1.39020671759817
0 1.38999723314873
0 1.38995711086427
0 1.38978950113205
0 1.39000165159562
0 1.39020671875351
0 1.38991757401444
0 1.38989639805958
0 1.38994419627331
0 1.3899578383338
0 1.38995596292442
0 1.38997841587594
0 1.39022772964495
0 1.38997788076079
0 1.38996381728165
0 1.39011975038351
0 1.38994826555152
0 1.38984401970507
0 1.38992088658805
0 1.3899675685332
0 1.38998175630482
0 1.38994428354368
0 1.38993171808371
0 1.3898640246744
0 1.3894443398808
0 1.38985543727413
0 1.38990952476569
0 1.38989546083087
0 1.39182336337043
0 1.39444050008691
0 1.38994233986479
0 1.38988352550877
0 1.38960903484048
0 1.38972379407329
0 1.38980660552304
0 1.3898963556356
0 1.38963054132451
0 1.3896670000952
0 1.38976055904873
0 1.38995305505609
1 1.50165620090517
1 1.39012485050752
1 1.38870939658855
1 1.38559773144462
1 1.38740811343837
1 1.49838616201526
1 1.38711835460014
1 1.38800270134074
1 1.38816293992327
1 1.38936199736992
1 1.38789981766402
1 1.39634768685908
1 1.38978722997779
1 1.39174005151323
1 1.38943951411412
1 1.4072039289706
1 1.35331429053798
1 1.38843397117794
1 1.38998122755894
1 1.38896132487251
1 1.38994990104046
1 1.38994966876781
1 1.39135539595928
1 1.39001090920019
1 1.55288046265865
1 1.38999612448217
1 1.3882658421185
1 1.29017924488122
1 1.38863660061877
1 1.46847194448353
1 1.38886618073746
1 1.39765456196771
1 1.38807599950969
1 1.38995009093219
1 1.3899569102567
1 1.37917715487456
1 1.39013620703099
1 1.38375306338998
1 1.39000261703806
1 1.38997201301839
2 1.50942224056234
2 1.39055642587939
2 1.38911536139611
2 1.38749970566613
2 1.38552007514548
2 1.50505797891806
2 1.39701893859234
2 1.39065895897997
2 1.38831948556349
2 1.38616178381862
2 1.38948627804559
2 1.38988608798505
2 1.38778661924619
2 1.3898115076719
2 1.38927139254849
2 1.4209480629486
2 1.38994977943768
2 1.38733089623588
2 1.38970312325167
2 1.38759282797558
2 1.39058521983944
2 1.39013019991999
2 1.38951117522409
2 1.38960553820815
2 1.44317836309799
2 1.38769638856824
2 1.38757609082353
2 1.3206228559606
2 1.38838095741432
2 1.38929627207753
2 1.3897720851698
2 1.38791601821781
2 1.3898605991916
2 1.38809128494824
2 0.922685660108578
2 1.38822736149161
2 1.38907410053868
2 1.38987172035952
2 1.38826615371052
2 1.38963624132503
};
\end{axis}

\end{tikzpicture}

%% file: img/box_plots/box_plot_computation_times.tex
% This file was created with tikzplotlib v0.10.1.
\begin{tikzpicture}[scale=0.48]

\definecolor{darkgray176}{RGB}{176,176,176}
\definecolor{darkgoldenrod16612945}{RGB}{166,129,45}
\definecolor{lightblue}{RGB}{173,216,230}
\definecolor{yellowgreen13918159}{RGB}{139,181,59}

\begin{axis}[
log basis y={10},
tick align=outside,
tick pos=left,
x grid style={darkgray176},
xmin=-0.5, xmax=6.5,
xtick style={color=black},
xtick={0.5,3,5.5},
xticklabels={1,2,3},
ylabel={$t$ in s},
y grid style={darkgray176},
ymin=0, ymax=150,
ymode=log,
ytick style={color=black},
ytick={0.1,1,10,100},
yticklabels={0.1,1,10,100}
]
\path [draw=black, fill=lightblue]
(axis cs:2.1,0.125708333333333)
--(axis cs:2.9,0.125708333333333)
--(axis cs:2.9,0.459625)
--(axis cs:2.1,0.459625)
--(axis cs:2.1,0.125708333333333)
--cycle;
\addplot [black]
table {%
2.5 0.125708333333333
2.5 0.0395
};
\addplot [black]
table {%
2.5 0.459625
2.5 0.79025
};
\addplot [black]
table {%
2.3 0.0395
2.7 0.0395
};
\addplot [black]
table {%
2.3 0.79025
2.7 0.79025
};
\addplot [red, mark=o, mark size=1, mark options={solid,fill opacity=0,draw=black}, only marks]
table {%
2.5 0.98125
2.5 5.7305
2.5 1.27783333333333
};
\path [draw=black, fill=lightblue]
(axis cs:4.6,0.232520833333333)
--(axis cs:5.4,0.232520833333333)
--(axis cs:5.4,0.999158333333333)
--(axis cs:4.6,0.999158333333333)
--(axis cs:4.6,0.232520833333333)
--cycle;
\addplot [black]
table {%
5 0.232520833333333
5 0.0465
};
\addplot [black]
table {%
5 0.999158333333333
5 1.837
};
\addplot [black]
table {%
4.8 0.0465
5.2 0.0465
};
\addplot [black]
table {%
4.8 1.837
5.2 1.837
};
\addplot [red, mark=o, mark size=1, mark options={solid,fill opacity=0,draw=black}, only marks]
table {%
5 3.49675
5 11.64425
5 8.029
5 4.0345
5 17.0881666666667
};
\path [draw=black, fill=lightblue]
(axis cs:0.1,0.744041666666667)
--(axis cs:0.9,0.744041666666667)
--(axis cs:0.9,2.45515)
--(axis cs:0.1,2.45515)
--(axis cs:0.1,0.744041666666667)
--cycle;
\addplot [black]
table {%
0.5 0.744041666666667
0.5 0.083
};
\addplot [black]
table {%
0.5 2.45515
0.5 4.486875
};
\addplot [black]
table {%
0.3 0.083
0.7 0.083
};
\addplot [black]
table {%
0.3 4.486875
0.7 4.486875
};
\addplot [red, mark=o, mark size=1, mark options={solid,fill opacity=0,draw=black}, only marks]
table {%
0.5 11.97
0.5 12.5726666666667
0.5 6.88083333333333
0.5 11.7838571428571
0.5 8.43385714285714
0.5 23.3972857142857
};
\path [draw=black, fill=yellowgreen13918159]
(axis cs:3.1,0.740583333333333)
--(axis cs:3.9,0.740583333333333)
--(axis cs:3.9,6.41568303571429)
--(axis cs:3.1,6.41568303571429)
--(axis cs:3.1,0.740583333333333)
--cycle;
\addplot [black]
table {%
3.5 0.740583333333333
3.5 0.134
};
\addplot [black]
table {%
3.5 6.41568303571429
3.5 13.8692
};
\addplot [black]
table {%
3.3 0.134
3.7 0.134
};
\addplot [black]
table {%
3.3 13.8692
3.7 13.8692
};
\addplot [red, mark=o, mark size=1, mark options={solid,fill opacity=0,draw=black}, only marks]
table {%
3.5 21.7725
3.5 35.24225
3.5 86.5143333333333
3.5 21.6017142857143
};
\path [draw=black, fill=yellowgreen13918159]
(axis cs:5.6,0.986875)
--(axis cs:6.4,0.986875)
--(axis cs:6.4,23.7468125)
--(axis cs:5.6,23.7468125)
--(axis cs:5.6,0.986875)
--cycle;
\addplot [black]
table {%
6 0.986875
6 0.314
};
\addplot [black]
table {%
6 23.7468125
6 50.6585
};
\addplot [black]
table {%
5.8 0.314
6.2 0.314
};
\addplot [black]
table {%
5.8 50.6585
6.2 50.6585
};
\addplot [red, mark=o, mark size=1, mark options={solid,fill opacity=0,draw=black}, only marks]
table {%
6 63.52325
6 72.3965
6 662.294666666667
6 76.617
6 109.179857142857
6 98.35975
};
\addplot [darkgoldenrod16612945]
table {%
2.1 0.258416666666667
2.9 0.258416666666667
};
\addplot [darkgoldenrod16612945]
table {%
4.6 0.64855
5.4 0.64855
};
\addplot [darkgoldenrod16612945]
table {%
0.1 1.189875
0.9 1.189875
};
\addplot [darkgoldenrod16612945]
table {%
3.1 2.57991666666667
3.9 2.57991666666667
};
\addplot [darkgoldenrod16612945]
table {%
5.6 4.6705
6.4 4.6705
};
\end{axis}

\end{tikzpicture}

%% file: content/6_conclusion.tex
\section{Conclusion}
\label{sec:conclusion}

In this paper, we combine inverse optimal control based prediction as a unified prediction model with trajectory planning for collision avoidance with non-communicating mobile robots. Each robot uses inverse optimal control to determine current estimates of unknown goal states of all robots based on observed trajectories. Trajectory predictions are calculated by solving a joint prediction problem using the estimated goal states and considered for planning. We have successfully evaluated the combined algorithm in scenarios with 2-8 robots. Further, we have compared our outcomes against the results of the combined algorithm where the robots communicate their goals and where the goal states are estimated based on a constant acceleration model. The median of the total durations until all vehicles have reached their goals is \num{9.8}\% shorter compared to planning with constant acceleration based estimations and only \num{62.3}\% longer compared to planning with known goals. In future work, we want to examine the proposed approach in hardware experiments.

\begin{ack}

This work is co-funded within the European Regional Development Fund (ERDF) 2021-2027 by the European Union and the State of Baden-Württemberg under the ERDF Programme Baden-Württemberg [further information: https://efre-bw.de/]. It is conducted as part of the RegioWIN 2030 project "Last Mile City Lab – Testing Ground for Urban Logistics". The views and conclusions expressed in this publication are solely those of the authors and do not necessarily reflect the views of the European Union, the State of Baden-Württemberg, or the granting authorities. Neither the European Union, the State of Baden-Württemberg, nor the granting authorities can be held responsible for them.

\end{ack}